\documentclass[12pt,journal,draftcls,a4paper,twoside,onecolumn]{IEEEtran}

\usepackage{setspace}
\usepackage{dsfont}
\usepackage{amsmath,amsthm}
\usepackage{amssymb}
\usepackage{amsfonts}
\usepackage{graphicx}
\usepackage[all,poly]{xy}
\usepackage{multirow}
\usepackage{epstopdf}
\usepackage{color}
\usepackage[noadjust]{cite}

\usepackage{float}
\usepackage{newfloat}
\usepackage{subfig}

\usepackage{mathrsfs}
\usepackage{algorithmic}
\usepackage[linesnumbered,lined,ruled]{algorithm2e}
\usepackage{cancel}
\usepackage{epsfig}
\usepackage{stfloats}
\usepackage{placeins}
\usepackage{url}
\usepackage{gensymb}

\usepackage[maxfloats=40]{morefloats}

\newfloat{bottomtwocolumnequation}{b}{fff}
\floatname{bottomtwocolumnequation}{Equation}

\newtheorem{theorem}{Theorem}

\newtheorem*{remark*}{Remark}

\newcommand{\bs}{\boldsymbol}

\newcommand{\NoiCovMat}{\bs{\Lambda}}

\newcommand{\NumEnd}{p}
\newcommand{\vecD}{\bar{\bfD}}

\newcommand{\Best}[1]{\textbf{\textcolor[rgb]{0.00,0.00,0.80}{{#1}}}}

\def\bfy{{\mathbf{y}}}

\def\bfA{{\mathbf{A}}}
\def\bfB{{\mathbf{B}}}
\def\bfC{{\mathbf{C}}}
\def\bfD{{\mathbf{D}}}
\def\bfE{{\mathbf{E}}}
\def\bfF{{\mathbf{F}}}
\def\bfG{{\mathbf{G}}}
\def\bfH{{\mathbf{H}}}

\def\bfM{{\mathbf{M}}}
\def\bfN{{\mathbf{N}}}
\def\bfO{{\mathbf{O}}}

\def\bfQ{{\mathbf{Q}}}
\def\bfR{{\mathbf{R}}}
\def\bfS{{\mathbf{S}}}
\def\bfT{{\mathbf{T}}}

\def\bfV{{\mathbf{V}}}
\def\bfW{{\mathbf{W}}}
\def\bfX{{\mathbf{X}}}
\def\bfY{{\mathbf{Y}}}
\def\bfZ{{\mathbf{Z}}}

\def\calA{{\mathcal{A}}}

\def\calL{{\mathcal{L}}}
\def\calM{{\mathcal{M}}}

\def\calO{{\mathcal{O}}}

\def\calT{{\mathcal{T}}}

\def\calX{{\mathcal{X}}}

\def\bsa{{\boldsymbol{a}}}

\def\bsm{{\boldsymbol{m}}}

\def\wtm{\widetilde{m}}

\newcommand{\MATima}{\bfX}

\newcommand{\noisevar}[1]{{s^2_{#1}}}

\newcommand{\nbbandima}{m_{\lambda}}

\newcommand{\argmin}{\mathrm{arg}\min}

\newcommand{\Vzeros}[1]{\boldsymbol{0}_{#1}}
\newcommand{\Id}[1]{\textbf{I}_{#1}}

\newcounter{algo}
\renewcommand{\thealgo}{\arabic{algo}} 

\begin{document}
\title{Multi-Band Image Fusion \\ Based on Spectral Unmixing}
\author{
\IEEEauthorblockN{Qi Wei, \IEEEmembership{Member,~IEEE},
Jos\'e Bioucas-Dias, \IEEEmembership{Senior Member,~IEEE},
Nicolas Dobigeon, \IEEEmembership{Senior Member,~IEEE}, 
Jean-Yves Tourneret, \IEEEmembership{Senior Member,~IEEE},
Marcus Chen, \IEEEmembership{Member,~IEEE},
and Simon Godsill, \IEEEmembership{Member,~IEEE}}

\thanks{
This work was supported by the HYPANEMA ANR Project under 
Grant ANR-12-BS03-003, the Portuguese Science and Technology
Foundation under Project UID/EEA/50008/2013, and the Thematic Trimester on Image Processing of the CIMI Labex,
Toulouse, France, under Grant ANR-11-LABX-0040-CIMI within the Program ANR-11-IDEX-0002-02.
Part of this work was presented during the IEEE
FUSION 2016 \cite{Wei2016FUSION}.}
\thanks{
Qi Wei and Simon Godsill are with Department of Engineering, 
University of Cambridge, CB2 1PZ, Cambridge, UK (e-mail: \{qi.wei, sjg\}@eng.cam.ac.uk).
Jos\'e Bioucas-Dias is with Instituto de Telecomunica\c{c}\~oes
and Instituto Superior T\'ecnico, Universidade de Lisboa, 1049-001, Lisboa, Portugal (e-mail: bioucas@lx.it.pt).
Nicolas Dobigeon and Jean-Yves Tourneret are with IRIT/INP-ENSEEIHT, University of 
Toulouse, 31071, Toulouse, France (e-mail: \{nicolas.dobigeon, jean-yves.tourneret\}@enseeiht.fr). 
Marcus Chen is with School of Computer Engineering, Nanyang Technological University,
Singapore (e-mail: marcuschen@pmail.ntu.edu.sg).}
}

\maketitle
\begin{abstract}
This paper presents a multi-band image fusion algorithm based on unsupervised spectral unmixing
for combining a high-spatial low-spectral resolution image and a low-spatial high-spectral
resolution image. The widely used linear observation model (with additive Gaussian noise) is combined 
with the linear spectral mixture model to form the likelihoods of the observations. The non-negativity 
and sum-to-one constraints
resulting from the intrinsic physical properties of the abundances are introduced as prior information
to regularize this ill-posed problem. The joint fusion and unmixing problem is then formulated
as maximizing the joint posterior distribution with respect to the endmember signatures and abundance maps,
This optimization problem is attacked with an alternating optimization strategy. The two resulting
sub-problems are convex and are solved efficiently using the alternating direction method of multipliers.
Experiments are conducted for both synthetic and semi-real data.
Simulation results show that the proposed unmixing based fusion scheme improves both the abundance
and endmember estimation comparing with the state-of-the-art joint fusion and unmixing algorithms.
\end{abstract}


\section{Introduction}
\label{sec:intro_FUMI}
Fusing multiple multi-band images enables a synergetic exploitation of complementary information obtained by
sensors of different spectral ranges and different spatial resolutions. In general, a multi-band image can be represented as a
three-dimensional data cube indexed by three exploratory variables $(x,y,\lambda)$, where $x$ and $y$
are the two spatial dimensions of the scene, and $\lambda$ is the spectral dimension (covering a
range of wavelengths). Typical examples of multi-band images include hyperspectral
(HS) images \cite{Landgrebe2002}, multi-spectral (MS) images \cite{Navulur2006}, integral
field spectrographs \cite{Bacon2001}, magnetic resonance spectroscopy images \cite{Nelson2004}.
However, multi-band images with high spectral resolution generally suffers from the limited
spatial resolution of the data acquisition devices, mainly due to physical and technological reasons.
These limitations make it infeasible to acquire a high spectral resolution multi-band image with 
a spatial resolution comparable to those of MS and panchromatic (PAN) images (which are acquired
in much fewer bands) \cite{Chang2007}. For example, HS images benefit from excellent spectroscopic
properties with several hundreds or thousands of contiguous bands but are limited by their relatively
low spatial resolution \cite{Shaw2003}. As a consequence, reconstructing a high-spatial and high-spectral multi-band 
image from multiple and complementary observed images, although challenging, is a crucial inverse problem that has been
addressed in various scenarios. In particular, fusing a high-spatial low-spectral resolution image and a 
low-spatial high-spectral image is an archetypal instance of multi-band image reconstruction, 
such as pansharpening (MS+PAN) \cite{Aiazzi2012} or HS pansharpening (HS+PAN) \cite{Loncan2015}.
The interested reader is invited to consult the references \cite{Aiazzi2012} and \cite{Loncan2015}
for an overview of the HS pansharpening problems and corresponding fusion algorithms.

In general, the degradation mechanisms in HS, MS, and PAN imaging,
with respect to (w.r.t.) the target high-spatial and high-spectral image can be summarized
as spatial and spectral transformations. Thus, the multi-band image fusion
problem can be interpreted as restoring a three dimensional data-cube from two
degraded data-cubes, which is an inverse problem. As this inverse problem is 
generally ill-posed, introducing prior distributions (regularizers in the the
regularization framework) to regularize the target
image has been widely explored \cite{Wei2015JSTSP,Wei2015TGRS,Simoes2015}. 
Regarding regularization, the usual high spectral and spatial correlations
of the target images imply that they admit sparse or low rank representations,
which has in fact been exploited in, for example, \cite{Zhukov1999,Eismann2004,
Yokoya2012coupled,An2014,Huang2014,Wei2015JSTSP,Wei2015TGRS,Simoes2015}.


In \cite{Eismann2004}, a \emph{maximum a posterior} (MAP) estimator incorporating a stochastic mixing model
has been designed for the fusion of HS and MS images. In \cite{Wycoff2013}, 
a non-negative sparse promoting algorithm for fusing HS and RGB images has been 
developed by using an alternating optimization algorithm. However,
both approaches developed in \cite{Eismann2004} and \cite{Wycoff2013} require
a very basic assumption that a low spatial resolution pixel is obtained by averaging
the high resolution pixels belonging to the same area, whose size depends the
downsampling ratio. 
This nontrivial assumption implies that the fusion of two multi-band images can be 
divided into fusing small blocks, which greatly decreases the complexity of the overall problem.
Note that this assumption has also been used in \cite{Zhang2009,Kawakami2011,Huang2014}. 
However, this averaging assumption 
can be violated easily as the area in a high resolution image corresponding to a low resolution pixel
can be arbitrarily large (depending on the spatial blurring) and the downsampling ratio is generally
fixed (depending on the sensor physical characteristics). 

To overcome this limitation, a more general forward model, 
which formulates the blurring and downsampling as two separate operations, has been recently developed
and widely used \cite{Berne2010,Yokoya2012coupled,He2014,Wei2015JSTSP,Simoes2015,Loncan2015}.
Based on this model, a non-negative matrix factorization pansharpening of HS image has 
been proposed in \cite{Berne2010}. Similar works have been developed independently 
in \cite{Bieniarz2011,Zhang2013HS,An2014}. Later, Yokoya \emph{et al.} have proposed 
to use a coupled nonnegative matrix factorization (CNMF) unmixing for the fusion of 
low-spatial-resolution HS and high-spatial-resolution MS data, where both HS and MS data are
alternately unmixed into endmember and abundance matrices by the CNMF 
algorithm \cite{Yokoya2012coupled}. Though this algorithm is rooted in physical principles 
and easy to implement owing to its simple update rules, 
it does not use the abundances estimated from the HS image and the endmember signatures
estimated from the MS image, which makes the spectral and spatial information in both 
images not fully exploited. 
It is worthy to note that a similar fusion and unmixing framework was
recently introduced in \cite{Lanaras2015}, in which the alternating NMF
steps in CNMF were replaced by alternating proximal forward-backward steps.

In this work, we formulate the unmixing based multi-band image fusion problem 
as an inverse problem in which the regularization is implicitly imposed by a low
rank representation inherent to the linear spectral mixture model and by non-negativity
and sum-to-one constraints resulting from the intrinsic physical properties of the
abundances.
In the proposed approach, the endmember signatures and abundances are jointly estimated 
from the observed multi-band images. The optimization w.r.t. the endmember
signatures and the abundances are both constrained linear regression problems,
which can be solved efficiently by the alternating direction method of multipliers (ADMM).


The remaining of this paper is organized as follows.
Section \ref{sec:Prob_State_FUMI} gives a short introduction of the widely used linear 
mixture model and forward model for multi-band images. Section \ref{sec:Prob_Form_FUMI}
formulates the unmixing based fusion problem as an optimization problem, which is solved 
using the Bayesian framework by introducing the popular constraints associated with the
endmembers and abundances. The proposed fast alternating optimization algorithm 
is presented in Section \ref{sec:Proposed_FUMI}. 
Section \ref{sec:simu_FUMI} presents experimental results
assessing the accuracy and the numerical efficiency of the proposed method. Conclusions are 
finally reported in Section \ref{sec:concls}.

\section{Problem Statement}
\label{sec:Prob_State_FUMI}
To better distinguish spectral and spatial properties, the pixels of
the target multi-band image, which is of high-spatial and high-spectral
resolution, can be rearranged to build an $\nbbandima \times n$
matrix $\bfX$, where $\nbbandima$ is the number of spectral bands and
$n=n_r \times n_c$ is the number of pixels in each band ($n_r$ and $n_c$ represent the numbers of rows and columns respectively).
In other words, each column of the matrix $\bfX$ consists of a $\nbbandima$-valued pixel and each row gathers all the pixel values in a given spectral band.
\subsection{Linear Mixture Model}
\label{subsec:LMM}
This work exploits an intrinsic property of multi-band images, according to which each spectral vector of 
an image can be represented by a linear mixture of several spectral signatures, referred to as endmembers.
Mathematically, we have
\begin{equation}
\bf X = MA
\label{eq:mix_model}
\end{equation}
where $\bfM \in \mathbb{R}^{\nbbandima  \times \NumEnd}$ is
the endmember matrix whose columns are spectral signatures and $\bfA \in \mathbb{R}^{\NumEnd \times n}$ is the
corresponding abundance matrix whose columns are abundance fractions.
This linear mixture model has been widely used in HS unmixing (see \cite{Bioucas-Dias2012} for a detailed review). 

\subsection{Forward Model}
Based on the pixel ordering introduced at the beginning of Section \ref{sec:Prob_State_FUMI}, any linear operation 
applied to the left (resp. right) side of $\bfX$ describes a spectral (resp. spatial) degradation action. In this work, 
we assume that two complementary images of high-spectral or high-spatial resolutions, respectively,
are available to reconstruct the target high-spectral and high-spatial resolution target image. These images result
from linear spectral and spatial degradations of the full resolution image $\bfX$, according to the popular models
\begin{equation}
\begin{array}{ll}
\label{eq:obs_general}
\bfY_{\mathrm{M}} =  {\bfR} \MATima + \bfN_{\mathrm{M}}\\
\bfY_{\mathrm{H}} =  \MATima \bfB \bfS + \bfN_{\mathrm{H}}
\end{array}
\end{equation}
where
\begin{itemize}
\item $\MATima \in \mathbb{R}^{\nbbandima \times n}$ is
	the full resolution target image as described in Section \ref{subsec:LMM}.
\item ${\bfY}_{\mathrm{M}}\in\mathbb{R}^{n_{\lambda} \times n}$ and ${\bfY}_{\mathrm{H}} \in \mathbb{R}^{\nbbandima \times m}$
are the observed spectrally degraded and spatially degraded images.
\item ${\bfR} \in\mathbb{R}^{n_{\lambda} \times \nbbandima}$ is the spectral response of the MS sensor,
which can be \emph{a priori} known or estimated by cross-calibration \cite{Yokoya2013cross}.
\item $\bfB \in \mathbb{R}^{n \times n}$ is a cyclic convolution operator acting on the bands. 
\item ${\bf S}\in\mathbb{R}^{n\times m}$ is a $d$ uniform downsampling operator (it has $m=n/d$ ones and zeros elsewhere),
which satisfies ${\bfS}^T \bfS=\Id{m}$.
\item $\bfN_{\mathrm{M}}$ and $\bfN_{\mathrm{H}}$ are additive terms that include
both modeling errors and sensor noises.
\end{itemize}
The noise matrices are assumed to be distributed according to the following matrix normal 
distributions\footnote{The probability density function $p(\bfX | \bfM, \bs{\Sigma}_r, \bs{\Sigma}_c)$
of a matrix normal distribution $\mathcal{MN}_{r,c}(\mathbf{M}, \bs{\Sigma}_r, \bs{\Sigma}_c)$
is defined by
\begin{equation*}
\begin{split}
p(\bfX | \bfM, \bs{\Sigma}_r, \bs{\Sigma}_c)=
\frac{\exp\left( -\frac{1}{2}\mathrm{tr}\left[ \bs{\Sigma}_c^{-1} (\mathbf{X} - \mathbf{M})^{T} \bs{\Sigma}_r^{-1} (\mathbf{X} - \mathbf{M}) \right] \right)}{(2\pi)^{rc/2} |\bs{\Sigma}_c|^{r/2} |\bs{\Sigma}_r|^{c/2}}
\end{split}
\end{equation*}
where $\bfM \in \mathbb{R}^{r \times c}$ is the mean matrix, $\bs{\Sigma}_r \in \mathbb{R}^{r \times r}$ is
the row covariance matrix and $\bs{\Sigma}_c \in \mathbb{R}^{c \times c}$ is the column covariance matrix.}
\begin{equation*}
    \begin{array}{ll}
      \bfN_{\mathrm{M}}  \sim \mathcal{MN}_{\nbbandima,m}(\Vzeros{\nbbandima,m},\NoiCovMat_{\mathrm{M}} ,\Id{m} ) \\
      \bfN_{\mathrm{H}}  \sim \mathcal{MN}_{n_{\lambda},n}(\Vzeros{n_{\lambda},n}, \NoiCovMat_{\mathrm{H}}, \Id{n})
    \end{array}
\end{equation*}
where $\Vzeros{a,b}$ is an $a \times b$ matrix of zeros and $\Id{a}$ is the $a \times a$ identity matrix.
The column covariance matrices are assumed to be the identity matrix to reflect the fact that the
noise is pixel-independent. The row covariance matrices $\NoiCovMat_{\mathrm{M}}$ and
$\NoiCovMat_{\mathrm{H}}$ are assumed to be diagonal matrices, whose diagonal elements
can vary depending on the noise powers in the different bands. More specifically, 
$\NoiCovMat_{\mathrm{H}}= \mathrm{diag} \left[\noisevar{\mathrm{H},1}, \cdots, \noisevar{\mathrm{H},m_{\lambda}} \right]$ 
and $\NoiCovMat_{\mathrm{M}}= \mathrm{diag} \left[\noisevar{\mathrm{M},1}, \cdots,
\noisevar{\mathrm{M},n_{\lambda}} \right]$, where $\mathrm{diag}$ is an operator transforming a vector into 
a diagonal matrix, whose diagonal terms are the elements of this vector.


The matrix equation \eqref{eq:obs_general} has been widely advocated for
the pansharpening and HS pansharpening problems, which consist of
fusing a PAN image with an MS or an HS image \cite{Amro2011survey,Gonzalez2004fusion,Loncan2015}.
Similarly, most of the techniques developed to fuse MS and HS images also rely on
a similar linear model \cite{Hardie2004,Molina1999,Molina2008,Zhang2012,Yokoya2012coupled,
Wei2014Bayesian,Wei2015TGRS}. From an application point of view,
this model is also important as motivated by recent national programs,
e.g., the Japanese next-generation space-borne HS image suite (HISUI), which acquires 
and fuses the co-registered HS and MS images for the same scene under the same conditions,
following this linear model \cite{Yokoya2013}.


\subsection{Composite Fusion Model}
Combining the linear mixture model \eqref{eq:mix_model} and the forward model \eqref{eq:obs_general} leads to
\begin{equation}
\begin{array}{ll}
\bfY_{\mathrm{M}} =  {\bf RMA} + \bfN_{\mathrm{M}}\\
\bfY_{\mathrm{H}} =  {\bf MABS} + \bfN_{\mathrm{H}}
\end{array}
\label{eq:obs_specific}
\end{equation}
where all matrix dimensions and their respective relations are
summarized in Table \ref{tb:size}.

\begin{table}[h!]
\centering \caption{Matrix dimension summary}
\small
\setlength{\tabcolsep}{0.5mm}
\begin{tabular}{c|c|c}
\hline
\textbf{Notation} & \textbf{Definition} & \textbf{Relation} \\
\hline
$m $	   & no. of pixels in each row of $\bfY_{\mathrm{H}}$ & $m=n/d$\\
$n $	   & no. of pixels in each row of $\bfY_{\mathrm{M}}$ & $n=m \times d$\\
$d$        & decimation factor   & $d=n/m$\\
$\nbbandima$        & no. of bands in $\bfY_{\mathrm{H}}$   & $\nbbandima \gg n_{\lambda}$\\
$n_{\lambda}$        & no. of bands in $\bfY_{\mathrm{M}}$   & $n_{\lambda} \ll \nbbandima$\\
\hline
\end{tabular}
\label{tb:size}
\end{table}

Note that the matrix $\bfM$ can be selected from a known spectral library \cite{Iordache2011SparseUnmixing}
or estimated \emph{a priori} from the HS data \cite{Wei2015FastUnmixing}. Also, it can be estimated jointly
with the abundance matrix $\bfA$ \cite{Nascimento2005,Dobigeon2009,Li2008MVSA},
which will be the case in this work.

\subsection{Statistical Methods}
To summarize, the problem of fusing and unmixing high-spectral
and high-spatial resolution images can be formulated as estimating 
the unknown matrices $\bfM$ and $\bfA$ from \eqref{eq:obs_specific},
which can be regarded as a joint non-negative matrix factorization (NMF) problem.
As is well known, the NMF problem is non-convex and has no unique solution, 
leading to an ill-posed problem. Thus, it is necessary to incorporate some intrinsic
constraints or prior information to regularize this problem, improving the conditioning of the problem.

Various priors have been already advocated to regularize the multi-band image
fusion problem, such as Gaussian priors \cite{Wei2015JSTSP,Wei2015whispers}, sparse representations
\cite{Wei2015TGRS} or total variation (TV) priors \cite{Simoes2015}.
The choice of the prior usually depends on the information resulting from previous experiments
or from a subjective view of constraints affecting the unknown model parameters \cite{Robert2007,Gelman2013}.
The inference of $\bf M$ and $\bf A$ (whatever the form chosen for the prior) is a challenging
task, mainly due to the large size of $\bfX$ and to the presence of the downsampling operator $\bfS$,
which prevents any direct use of the Fourier transform to diagonalize the blurring operator $\bfB$.
To overcome this difficulty, several computational strategies, including Markov chain Monte Carlo
(MCMC) \cite{Wei2015JSTSP}, block coordinate descent method (BCD) \cite{Bertsekas1999}, and tailored 
variable splitting under the ADMM framework \cite{Simoes2015}, have been proposed,
both applied to different kinds of priors, e.g., the empirical Gaussian prior \cite{Wei2015JSTSP,Wei2015whispers},
the sparse presentation based prior \cite{Wei2015TGRS}, or the TV prior \cite{Simoes2015}. More recently, contrary
to the algorithms described above, a much more efficient method, named \emph{Robust Fast fUsion based
on Sylvester Equation (R-FUSE)} has been proposed to solve explicitly an underlying Sylvester equation associated 
with the fusion problem derived from \eqref{eq:obs_specific} \cite{Wei2016RFUSE}. 
This solution can be implemented \emph{per se} to compute the maximum likelihood estimator in a computationally 
efficient manner, which has also the great advantage of being easily generalizable within a Bayesian framework when considering various priors.

In our work, we propose to form priors by exploiting the intrinsic physical properties of abundances and endmembers,
which is widely used in conventional unmixing, to infer $\bfA$ and $\bfM$ from the observed data $\bfY_{\mathrm{M}}$
and $\bfY_{\mathrm{H}}$. More details will be give in following sections.




\section{Problem Formulation}
\label{sec:Prob_Form_FUMI}


Following the Bayes rule, the posterior distribution of the unknown parameters $\bfM$ and $\bfA$
can be obtained by the product of their likelihoods and prior distributions, which are detailed in what follows.

\subsection{Likelihoods (Data Fidelity Term)}
\label{subsec:likelihood}
Using the statistical properties of the noise matrices $\bfN_{\mathrm{M}}$ and $\bfN_{\mathrm{H}}$,
$\bfY_{\mathrm{M}}$ and $\bfY_{\mathrm{H}}$ have matrix Gaussian distributions, i.e.,
\begin{equation}
\begin{array}{ll}
\label{eq:Likelihood}
p\left(\bfY_{\mathrm{M}}|\bfM,\bfA\right) = \mathcal{MN}_{n_{\lambda},n}({\bf RMA},\NoiCovMat_{\mathrm{M}}, \Id{n})\\
p\left(\bfY_{\mathrm{H}}|\bfM,\bfA\right) = \mathcal{MN}_{\nbbandima,m}({\bf MABS}, \NoiCovMat_{\mathrm{H}}, \Id{m}).
\end{array}
\end{equation}
As the collected measurements $\bfY_{\mathrm{M}}$ and $\bfY_{\mathrm{H}}$ have been
acquired by different (possibly heterogeneous) sensors, the noise matrices $\bfN_{\mathrm{M}}$
and $\bfN_{\mathrm{H}}$ are sensor-dependent and can be generally assumed to be statistically independent.
Therefore, $\bfY_{\mathrm{M}}$ and $\bfY_{\mathrm{H}}$ are independent conditionally upon
the unobserved scene $\bfX=\bf MA$. As a consequence, the joint likelihood function of the observed
data is
\begin{equation}
\label{eq:Likelihood_joint}
p\left(\bfY_{\mathrm{M}},\bfY_{\mathrm{H}}|\bfM,\bfA\right)=p\left(\bfY_{\mathrm{M}}|\bfM,\bfA\right)p\left(\bfY_{\mathrm{H}}|\bfM,\bfA\right).
\end{equation}
The negative logarithm of the likelihood is
\begin{equation*}
\begin{array}{ll}
&-\log p\left(\bfY_{\mathrm{M}}, \bfY_{\mathrm{H}}|\bfM,\bfA\right)  \\
&= -\log p\left(\bfY_{\mathrm{M}}|\bfM,\bfA\right) -\textrm{log } p\left(\bfY_{\mathrm{H}}|\bfM,\bfA\right) +C\\
&=\frac{1}{2}\big\|\NoiCovMat_{\mathrm{H}}^{-\frac{1}{2}}\left(\bfY_{\mathrm{H}}- \bf{MABS}\right)\big\|_F^2
+\frac{1}{2} \big\|\NoiCovMat_{\mathrm{M}}^{-\frac{1}{2}}\left(\bfY_{\mathrm{M}}- \bf{RMA}\right)\big\|_F^2\\
&+C
\end{array}
\end{equation*}
where $\|{\bfX}\|_F = \sqrt{\text{trace}\left(\bfX^T\bfX\right)}$ is the Frobenius norm of $\bfX$ and $C$ is a constant.

\subsection{Priors (Regularization Term)}
\subsubsection{Abundances}
As the mixing coefficient $a_{i,j}$ (the element located in the $i$th row and $j$th column of $\bfA$) represents the
proportion (or probability of occurrence) of the the $i$th endmember in the $j$th measurement \cite{Keshava2002,Bioucas-Dias2012},
the abundance vectors satisfy the following \emph{abundance non-negativity constraint} (ANC) and \emph{abundance sum-to-one constraint} (ASC)
\begin{equation}
\label{eq:asc_anc_vector}
\bsa_j \geq 0 \quad \textrm{and} \quad \bs{1}_p^T \bsa_j=1, \forall j \in \left\{1,\cdots,n\right\}
\end{equation}
where $\bsa_j$ is the $j$th column of $\bfA$, $\geq$ means ``element-wise greater than'' and $\bs{1}_p^T$ is a $p \times 1$ vector with all
ones. Accounting for all the image pixels, the constraints \eqref{eq:asc_anc_vector}
can be rewritten in matrix form
\begin{equation}
\bfA\geq 0 \quad \textrm{and} \quad \bs{1}_p^T \bfA=\bs{1}_n^T.
\end{equation}
Moreover, the ANC and ASC constraints can be converted into a
uniform distribution for $\bfA$ on the feasible region $\mathcal{A}$, i.e.,
\begin{equation}
p(\bfA)=\left\{
\begin{array}{ll}
c_{\bfA} & \textrm{if } \bfA \in \calA \\
0 & \textrm{elsewhere}
\end{array} \right.
\end{equation}
where $\mathcal{A} = \left\{\bfA|\bfA \geq 0, \bs{1}^T_p \bfA=\bs{1}^T_n\right\}$,
$c_{\bfA}=1\slash\mathrm{vol}(\mathcal{A})$ and $\mathrm{vol}(\calA) = \int_{\bfA \in \calA} d {\bfA}$ is the volume of the set $\calA$.

\subsubsection{Endmembers}
As the endmember signatures represent the reflectances of different materials, each element of the matrix $\bfM$
should be between $0$ and $1$. Thus, the constraints for $\bfM$ can be written as
\begin{equation}
0 \leq \bfM \leq 1.
\label{eq:end_cons}
\end{equation}
Similarly, these constraints for the matrix $\bfM$ can be converted into a
uniform distribution on the feasible region $\calM$

\begin{equation*}
p(\bfM)=\left\{
\begin{array}{ll}
c_{\bfM} & \textrm{if } \bfM \in \calM \\
0 & \textrm{elsewhere}
\end{array} \right.
\end{equation*}
where $\mathcal{M} = \left\{\bfM|0 \leq \bfM \leq 1\right\}$ and $c_{\bfM}=1\slash\mathrm{vol}(\mathcal{M})$.

\subsection{Posteriors (Constrained Optimization)}
Combining the likelihoods \eqref{eq:Likelihood_joint} and the priors $p\left(\bfM\right)$ and $p\left(\bfA\right)$,
the Bayes theorem provides the posterior distribution of $\bfM$ and $\bfA$
\begin{equation*}
\label{eq:posterior_joint}
\begin{array}{cl}
  &p\left(\bfM,\bfA|\bfY_{\mathrm{H}},\bfY_{\mathrm{M}}\right)  \\
  &\propto p\left(\bfY_{\mathrm{H}}|\bfM,\bfA\right)p\left(\bfY_{\mathrm{M}}|\bfM,\bfA\right)p\left(\bfM\right)p\left(\bfA\right)
\end{array}
\end{equation*}
where $\propto$ means ``proportional to''.
Thus, the unmixing based fusion problem can be
interpreted as maximizing the joint posterior distribution of $\bfA$ and $\bfM$.
Moreover, by taking the negative logarithm of $p\left(\bfM,\bfA|\bfY_{\mathrm{H}},\bfY_{\mathrm{M}}\right)$,
the MAP estimator of $(\bfA,\bfM)$ can be obtained by solving
the minimization 
\begin{equation}
\label{eq:neglog_map_FUMI}
\begin{split}
\min\limits_{\bfM,\bfA} L(\bfM,\bfA)\quad
&\textrm{s.t.}  \quad\bfA \geq 0 \quad \textrm{and} \quad{\bf 1}^T_p{\bf A} = {\bf 1}_n^T\\
&\quad \quad 0 \leq \bfM \leq 1
\end{split}
\end{equation}
where
\begin{equation*}
\begin{split}
L(\bfM,\bfA)&=\frac{1}{2}\big\|\NoiCovMat_{\mathrm{H}}^{-\frac{1}{2}}\left(\bfY_{\mathrm{H}}-\bf{MABS}\right)\big\|_F^2\\ 
&+\frac{1}{2} \big\|\NoiCovMat_{\mathrm{M}}^{-\frac{1}{2}}\left(\bfY_{\mathrm{M}}- \bf{RMA}\right)\big\|_F^2.
\end{split}
\end{equation*}

In this formulation, the fusion problem can be regarded as a generalized unmixing problem, which includes
two data fidelity terms. Thus, both images contribute to the estimation of the
endmember signatures (endmember extraction step) and the high-resolution abundance maps
(inversion step). For the endmember estimation, a popular strategy is to use a
subspace transformation as a preprocessing step, such as in \cite{Bioucas2008,Dobigeon2009}. In general,
the subspace transformation is learned \emph{a priori} from the high-spectral resolution image empirically,
e.g., from the HS data. This empirical subspace transformation alleviates the computational burden greatly
and can be incorporated in our framework easily.

\section{Alternating Optimization Scheme}
\label{sec:Proposed_FUMI}

Even though problem \eqref{eq:neglog_map_FUMI} is convex w.r.t. $\bfA$ and $\bfM$ separately,
it is non-convex w.r.t. these two matrices jointly and 
has more than one solution. We propose an optimization technique that alternates optimizations
w.r.t. $\bfA$ and $\bfM$, which is also referred to as a BCD algorithm.
The optimization w.r.t. $\bfA$ (resp. $\bfM$) conditional on $\bfM$ (resp. $\bfA$)
can be achieved efficiently with the ADMM algorithm \cite{Boyd2011}, which converges to
a solution of the respective convex optimization under some mild conditions. 
The resulting alternating optimization algorithm, referred to as Fusion based on Unmixing for Multi-band Images (FUMI),
is detailed in Algorithm \ref{Algo:AlterOpti_FUMI}, where
EEA$({\bfY}_{\mathrm{H}})$ in line 1 represents an endmember extraction algorithm to estimate endmembers from HS data.
The optimization steps w.r.t. $\bfA$ and $\bfM$ are detailed below.


\IncMargin{1em}
\begin{algorithm}[h!]
\KwIn{$\bfY_{\mathrm{M}}$, $\bfY_{\mathrm{H}}$,  $\NoiCovMat_{\mathrm{M}}$, $\NoiCovMat_{\mathrm{H}}$, $\bfR$, $\bfB$, $\bfS$ }
\tcc{Initialize $\bfM$}
$\bfM^{(0)} \leftarrow \textrm{EEA}(\bfY_{\mathrm{H}})$\;
\For{ $t = 1,2, \ldots$ \KwTo stopping rule}{
    \tcc{Optimize w.r.t. $\bfA$ using ADMM (see Algorithm \ref{Algo:ADMM_A})}
    ${\bfA}^{(t)}  \in \argmin\limits_{\bfA \in \calA} L(\bfM^{(t-1)}, {\bfA})$\;
    \tcc{Optimize w.r.t. $\bfM$ using ADMM (see Algorithm \ref{Algo:ADMM_M})}
    ${\bfM}^{(t)}  \in \argmin\limits_{\bfM \in \calM} L(\bfM, {\bfA}^{(t)})$\;
}
Set $\hat{\bfA}={\bfA}^{(t)}$ and $\hat{\bfM}={\bfM}^{(t)}$\;
\KwOut{$\hat{\bfA}$ and $\hat{\bfM}$}
\caption{Multi-band Image Fusion based on Spectral Unmixing (FUMI)}
\label{Algo:AlterOpti_FUMI}
\end{algorithm}
\DecMargin{1em}

\subsection{Convergence Analysis}
To analyze the convergence of Algorithm \ref{Algo:AlterOpti_FUMI},
we recall a convergence criterion for the BCD algorithm stated in
\cite[p.~273]{Bertsekas1999}.

\begin{theorem}[Bertsekas, \cite{Bertsekas1999}; Proposition 2.7.1]
\label{theo:Conv_Rate_FUMI}
Suppose that $L$ is continuously differentiable w.r.t. $\bfA$ and $\bfM$ over 
the convex set $\calA \times \calM$. Suppose also that for 
each $\left\{\bfA,\bfM\right\}$, $L(\bfA,\bfM)$ viewed as a function
of $\bfA$, attains a unique minimum $\bar{\bfA}$. 
The similar uniqueness also holds for $\bfM$. Let $\left\{\bfA^{(t)},\bfM^{(t)}\right\}$ 
be the sequence generated by the BCD method as in Algorithm \ref{Algo:AlterOpti_FUMI}.
Then, every limit point of $\left\{\bfA^{(t)},\bfM^{(t)}\right\}$ is a stationary point.
\end{theorem}

The target function defined in \eqref{eq:neglog_map_FUMI} 
is continuously differentiable.
Note that it is not guaranteed that the minima w.r.t. $\bfA$ or $\bfM$ are unique. 
We may however argue that a simple modification of the objective function, consisting 
in adding the quadratic term $\alpha_1\|\bfA\|_F^2 + \alpha_2\|\bfM\|_F^2$, where $\alpha_1$ and $\alpha_2$
are very small thus obtaining a strictly convex objective function, ensures that 
the minima of $\eqref{eq:min_A}$ and $\eqref{eq:min_M}$ are uniquely attained and thus 
we may invoke the Theorem \eqref{theo:Conv_Rate_FUMI}. In practice, even without including
the quadratic terms, we have systematically observed convergence of Algorithm \ref{Algo:AlterOpti_FUMI}.



\subsection{Optimization w.r.t. the Abundance Matrix $\bfA$ ($\bfM$ fixed)}
The minimization of $L(\bfM,\bfA)$ w.r.t. the abundance matrix $\bfA$ conditional on $\bfM$
can be formulated as
\begin{equation}
\label{eq:min_A}
\begin{split}
&\min\limits_{\bfA} \frac{1}{2}\big\|\NoiCovMat_{\mathrm{H}}^{-\frac{1}{2}}\left(\bfY_{\mathrm{H}}- \bf{MABS}\right)\big\|_F^2
+\frac{1}{2} \big\|\NoiCovMat_{\mathrm{M}}^{-\frac{1}{2}}\left(\bfY_{\mathrm{M}}- \bf{RMA}\right)\big\|_F^2\\
&\textrm{s.t.}  \quad\bfA \geq 0 \quad \textrm{and} \quad{\bf 1}^T_p{\bf A} = {\bf 1}_n^T.\\
\end{split}
\end{equation}
This constrained minimization problem can be solved by introducing an auxiliary variable to
split the objective and the constraints, which is the spirit of the ADMM algorithm. More specifically,
by introducing the splitting $\bfV = \bfA$, the optimization problem \eqref{eq:min_A} w.r.t. $\bfA$ can be written as
\begin{equation*}
\min\limits_{\bfA,\bfV} L_1(\bfA)+\iota_{\calA}(\bfV) \textrm{  s.t. } \bfV = \bfA
\end{equation*}
where $L_1(\bfA)=$
\begin{equation*}
\frac{1}{2}\big\|\NoiCovMat_{\mathrm{H}}^{-\frac{1}{2}}\left(\bfY_{\mathrm{H}}- \bf{MABS}\right)\big\|_F^2
+\frac{1}{2} \big\|\NoiCovMat_{\mathrm{M}}^{-\frac{1}{2}}\left(\bfY_{\mathrm{M}}- \bf{RMA}\right)\big\|_F^2
\end{equation*}
and
\begin{equation*}
\iota_{\calA}(\bfV)=
\left\{
\begin{array}{ll}
0 & \textrm{if } \bfV \in \calA \\
+\infty & \textrm{otherwise.} \end{array} \right.
\end{equation*}
Recall that $\mathcal{A} = \left\{\bfA|\bfA \geq 0, \bs{1}_p^T \bfA=\bs{1}_n \right\}$.

The augmented Lagrangian associated with the optimization of $\bfA$ can be written as
\begin{equation}
\begin{split}
&\calL({\bfA}, {\bfV}, {\bf G})= \frac{1}{2}\big\|\NoiCovMat_{\mathrm{H}}^{-\frac{1}{2}}\left(\bfY_{\mathrm{H}}- \bf{MABS}\right)\big\|_F^2+\iota_{\calA}(\bfV)\\
&+\frac{1}{2} \big\|\NoiCovMat_{\mathrm{M}}^{-\frac{1}{2}}\left(\bfY_{\mathrm{M}}- \bf{RMA}\right)\big\|_F^2+\frac{\mu}{2}\big\|{\bfA}-{\bfV}-{\bfG}\big\|_F^2 \\
\label{eq:COST_LAG_FUMI}
\end{split}
\end{equation}
where $\bfG$ is the so-called scaled dual variable and $\mu >0$ is the augmented Lagrange multiplier, weighting the augmented Lagrangian term \cite{Boyd2011}.
The ADMM summarized in Algorithm \ref{Algo:ADMM_A}, consists of an $\bfA$-minimization step, a $\bfV$-minimization step and a dual variable $\bfG$ update step
(see \cite{Boyd2011} for further details about ADMM). Note that the operator $\Pi_{\calX}(\bfX)$ in Algorithm \ref{Algo:ADMM_A} represents
projecting the variable $\bfX$ onto a set $\calX$, which is defined as
\begin{equation*}
\Pi_{\calX}(\bfX)= \argmin\limits_{\bfZ \in \calX}{\big\|\bfZ-\bfX\big\|_F^2}.
\end{equation*}

\IncMargin{1em}
\begin{algorithm}[h!]
\KwIn{$\bfY_{\mathrm{M}}$, $\bfY_{\mathrm{H}}$, $\NoiCovMat_{\mathrm{M}}$, $\NoiCovMat_{\mathrm{H}}$, $\bfR$, $\bfB$, $\bfS$, $\mu>0$} 
{\bf Initialization}: $\bfV^{(0)}$,${\bf G}^{(0)}$\;

\For{ $k = 0$ \KwTo stopping rule}
{
\tcc{Optimize w.r.t $\bfA$ (Algorithm \ref{Algo:Sylvester_Algo})}
${\bfA}^{(t,k+1)} \in \argmin\limits_{\bfA} \calL({\bfA}, {\bfV}^{(k)}, {\bfG}^{(k)})$\;
\tcc{Optimize w.r.t $\bfV$ (Algorithm \ref{Algo:Proj_Simplex})}
${\bfV}^{(k+1)} \leftarrow \Pi_{\calA} ( {\bfA}^{(t,k+1)}-{\bfG}^{(k)})$\;
\tcc{Update Dual Variable $\bfG$}
${\bfG}^{(k+1)} \leftarrow  \bfG^{(k)} - \left({\bfA}^{(t,k+1)} -{\bf V}^{(k+1)}\right)$\;}
Set ${\bfA}^{(t+1)} = {\bfA}^{(t,k+1)}$\;
\KwOut{${\bfA}^{(t+1)}$}
\caption{ADMM sub-iterations to estimate $\bfA$}
\label{Algo:ADMM_A}
\end{algorithm}
\DecMargin{1em}

Given that the functions $L_1(\bfA)$ and $\iota_{\calA}(\bfV)$ are both closed, 
proper, and convex, thus, invoking the Eckstein and Bertsekas theorem \cite[Theorem 8]{Eckstein1992},
the convergence of Algorithm \ref{Algo:ADMM_A} to a solution of \eqref{eq:min_A} is guaranteed. 

\subsubsection{Updating $\bfA$}
In order to minimize \eqref{eq:COST_LAG_FUMI} w.r.t. $\bfA$,
we solve the equation ${\partial \calL({\bfA}, {\bfV}^{(k)}, {\bfG}^{(k)})}/{\partial \bfA}=\bs{0}$,
which is equivalent to the generalized Sylvester equation
\begin{equation}
\label{eq:sylvester}
\bfC_1\bfA+\bfA \bfC_2 = \bfC_3
\end{equation}
where 
\begin{equation*}
\begin{split}
\bfC_1&=\left(\bfM^T \NoiCovMat_{\mathrm{H}}^{-1} \bfM\right)^{-1}\left({\left({\bf RM}\right)}^T \NoiCovMat_{\mathrm{M}}^{-1} {{\bf RM}}+{\mu}\Id{p}\right)\\
\bfC_2&={\bf BS} \left(\bf BS \right)^T\\
\bfC_3&=\left(\bfM^T \NoiCovMat_{\mathrm{H}}^{-1} \bfM\right)^{-1} (\bfM^T \NoiCovMat_{\mathrm{H}}^{-1} {\bfY}_\mathrm{H} \left(\bf BS \right)^T \\
& + {\left({\bfR}\bfM\right)}^T \NoiCovMat_{\mathrm{M}}^{-1} {\bfY}_\mathrm{M}+\mu(\bfV^{(k)}+\bfG^{(k)})).
\end{split}
\end{equation*}
Eq. \eqref{eq:sylvester} can be solved analytically 
by exploiting the properties of the circulant and downsampling matrices $\bfB$ and $\bfS$, 
as summarized in Algorithm \ref{Algo:Sylvester_Algo} and demonstrated in \cite{Wei2016RFUSE}.
Note that the matrix $\bfF$ represents the FFT operation and its conjugate transpose (or Hermitian transpose) 
$\bfF^H$ represents the iFFT operation. The matrix $\bfD \in \mathbb{C}^{n \times n}$ is a diagonal matrix, which has 
eigenvalues of the matrix $\bfB$ in its diagonal line and can be rewritten as
\begin{equation*}
\bfD=\left[
\begin{array}{ccccc}
\bfD_1 &\bs{0} &\cdots &\bs{0}\\
\bs{0}& \bfD_2& \cdots &\bs{0}\\
\vdots&\vdots &\ddots &\vdots\\
\bs{0}& \bs{0}& \cdots &\bfD_d
\end{array}
\right]
\end{equation*}
where $\bfD_i \in \mathbb{C}^{m \times m}$.
Thus, we have $\vecD^H\vecD=\sum\limits_{t=1}^d\bfD_t^H\bfD_t=\sum\limits_{t=1}^d\bfD_t^2$, where
$\vecD=\bfD\left(\bs{1}_d \otimes \Id{m}\right)$.
Similarly, the diagonal matrix $\bf{\Lambda}_C$ has eigenvalues of the matrix $\bfC_1$ in its diagonal line
(denoted as $\lambda_1,\cdots,\lambda_{\wtm_{\lambda}}$ and $\lambda_i \geq 0$, $\forall i$).
The matrix $\bfQ$ contains eigenvectors of the matrix $\bfC_1$ in its columns.
The auxiliary matrix $\bar{\bfA} \in \mathbb{C}^{\nbbandima \times n}$ is decomposed
as $\bar{\bfA}=\left[\bar{\bsa}_1^T,\bar{\bsa}_2^T,\cdots,\bar{\bsa}_p^T\right]^T$.

\IncMargin{1em}
\begin{algorithm}[h!]
\KwIn{$\bfY_{\mathrm{M}}$, $\bfY_{\mathrm{H}}$, $\NoiCovMat_{\mathrm{M}}$,
 $\NoiCovMat_{\mathrm{H}}$, $\bfR$, $\bfB$, $\bfS$, $\bfV^{(k)}$, $\bfG^{(k)}$, $\mu>0$ }
 \tcc{Circulant matrix decomposition: $\bfB= {\bfF \bfD \bfF}^H$}
 $\bf{D} \leftarrow  \textrm{EigDec} \left(\bfB\right)$\;
 $\vecD \leftarrow \bfD\left(\bs{1}_d \otimes \Id{m}\right)$\;
 \tcc{Calculate $\bfC_1$}
 $\bfC_1 \leftarrow \left(\bfM^T \NoiCovMat_{\mathrm{H}}^{-1} \bfM\right)^{-1}\left({\left({\bf RM}\right)}^T \NoiCovMat_{\mathrm{M}}^{-1} {{\bf RM}}+{\mu}\Id{p}\right)$\;
 \tcc{Eigen-decomposition of $\bfC_1$: $\bfC_1=\bfQ{\bf{\Lambda}}_C\bfQ^{-1}$}
 $\left({\bfQ,\bf{\Lambda}}_C\right) \leftarrow \textrm{EigDec}\left(\bfC_1\right)$\;
 \tcc{Calculate ${\bfC}_3$}
 ${\bfC}_3 \leftarrow \left({\bfM}^T \NoiCovMat_{\mathrm{H}}^{-1} {\bfM}\right)^{-1} ({\bfM}^T \NoiCovMat_{\mathrm{H}}^{-1} {\bfY}_\mathrm{H} \left(\bf BS \right)^T $\
 $+ {\left({\bf RM}\right)}^T \NoiCovMat_{\mathrm{M}}^{-1} {\bfY}_\mathrm{M} +{\mu} (\bfV^{(k)}+\bfG^{(k)}) )$\;
  \tcc{Calculate $\bar{\bfC}_3$}
 $\bar{\bfC}_3 \leftarrow \bfQ^{-1} \bfC_3 {\bfF}$\;
 \tcc{Calculate $\bar{\bfA}$ band by band}
 \For{$l=1$ \KwTo $\NumEnd$}{
 \tcc{Calculate the $l$th band}
$\bar{\bsa}_{l}\leftarrow \lambda_l^{-1} (\bar{\bfC}_3)_{l}-\lambda_l^{-1} (\bar{\bfC}_3)_{l}\vecD\left(\lambda_l d \Id{m} +\sum\limits_{t=1}^d \bfD_t^2\right)\vecD^H$\;
}
Set $\bfA= {\bfQ}\bar{\bfA} {\bfF}^H$\;
\KwOut{${\bfA}$}
\caption{A closed-form solution of \eqref{eq:sylvester} w.r.t. $\bfA$}
\label{Algo:Sylvester_Algo}
\end{algorithm}
\DecMargin{1em}

\subsubsection{Updating $\bfV$}
The update of $\bfV$ can be made by simply computing the Euclidean projection of ${\bfA}^{(t,k+1)} -{\bfG}^{(k+1)}$ onto the
canonical simplex $\calA$, which can be expressed as follows
\begin{align*}
\label{eq:update_v}
\hat{\bfV}&=\argmin\limits_{\bfV}\frac{\mu}{2}\big\|{\bfV}-\left({\bfA}^{(t,k+1)} -{\bfG}^{(k+1)}\right)\big\|_F^2+\iota_{\calA}(\bfV) \\
&=\Pi_{\calA}\left({\bfA}^{(t,k+1)} -{\bfG}^{(k+1)}\right)
\end{align*}
where $\Pi_{\calA}$ denotes the projection (in the sense of the Euclidean norm) onto the simplex $\calA$.
This classical projection problem has been widely studied and can be achieved by numerous methods
\cite{Held1974, Michelot1986,Duchi2008,Condat2014Fast}. In this work, we adopt the popular strategy 
first proposed in \cite{Held1974} and summarized in Algorithm \ref{Algo:Proj_Simplex}. Note that 
the above optimization is decoupled w.r.t. the columns of $\bfV$, denoted by $(\bfV)_1,\cdots,(\bfV)_n$, 
which accelerates the projection dramatically. 

\IncMargin{1em}
\begin{algorithm}[h!]
\KwIn{${\bfA}^{(t,k+1)} -{\bfG}^{(k)}$}
\For{$i=1$ \KwTo $n$}{
$(\bfA-\bfG)_i \triangleq$ $i^\textrm{th}$ column of ${\bfA}^{(t,k+1)} -{\bfG}^{(k)}$\;
\tcc{Sorting the elements of $(\bfA-\bfG)_i$}
Sort $(\bfA-\bfG)_i$ into $\bfy$: $y_1 \geq \cdots \geq y_p$ \;
Set $K: = \max\limits_{1\leq k \leq p}\{k|\left(\sum_{r=1}^k y_r - 1\right)/k<y_k\}$\;
Set $\tau:=\left(\sum_{r=1}^K y_r - 1\right)/K$\;
\tcc{The $\max$ operation is component-wise}
Set $(\hat{\bfV})_i: = \max\{(\bfA-\bfG)_i-\tau,0\}$\;
}
\KwOut{$\bfV^{(k+1)}=\hat{\bfV}$}
\caption{Projection onto the Simplex $\calA$}
\label{Algo:Proj_Simplex}
\end{algorithm}
\DecMargin{1em}

In practice, the ASC constraint is sometimes criticized for not being 
able to account for every material in a pixel or due to endmember variability \cite{Bioucas-Dias2012}.
In this case, the sum-to-one constraint can be simply removed. Thus, the Algorithm \ref{Algo:Proj_Simplex} 
will degenerate to projecting $(\bfA-\bfG)_i$ onto the non-negative half-space, which simply consists of
setting the negative values of $(\bfA-\bfG)_i$ to zeros.

\subsection{Optimization w.r.t. the Endmember Matrix $\bfM$ ($\bfA$ fixed)}
The minimization of \eqref{eq:neglog_map_FUMI} w.r.t. the abundance matrix $\bfM$ conditional on $\bfA$
can be formulated as
\begin{equation}
\label{eq:min_M_org}
\begin{array}{cc}
\min\limits_{\bfM} &  L_1(\bfM) +  \iota_{\calM}(\bfM)\\ 
\end{array}
\end{equation}
where $L_1(\bfM)=$
\begin{equation*}
\frac{1}{2}\big\|\NoiCovMat_{\mathrm{H}}^{-\frac{1}{2}}\left(\bfY_{\mathrm{H}}- \bfM\bfA_{\mathrm{H}}\right)\big\|_F^2 
+\frac{1}{2} \big\|\NoiCovMat_{\mathrm{M}}^{-\frac{1}{2}}\left(\bfY_{\mathrm{M}}- \bf{RMA}\right)\big\|_F^2
\end{equation*}
and $\bfA_{\mathrm{H}}=\bf{ABS}$. By splitting the quadratic data fidelity term and the inequality constraints,
the augmented Lagrangian for \eqref{eq:min_M}
can be expressed as
\begin{equation}
\label{eq:min_M}
\calL(\bfM,\bfT,\bfG)=L_1(\bfM)+\iota_{\calM}(\NoiCovMat_{\mathrm{H}}^{\frac{1}{2}}\bfT) +\frac{\mu}{2}\big\|\NoiCovMat_{\mathrm{H}}^{-\frac{1}{2}}{\bfM}-{\bfT}-{\bfG}\big\|_F^2. \\
\end{equation}
The optimization of $\calL(\bfM,\bfT,\bfG)$ consists of updating $\bfM$, $\bfT$ and $\bfG$ iteratively as summarized
in Algorithm \ref{Algo:ADMM_M} and detailed below. As $L_1(\bfM)$ and $\iota_{\calM}(\NoiCovMat_{\mathrm{H}}^{\frac{1}{2}}\bfT)$ 
are closed, proper and convex functions and $\NoiCovMat_{\mathrm{H}}^{\frac{1}{2}}$ has full column rank, the ADMM 
is guaranteed to converge to a solution of problem \eqref{eq:min_M_org}.

\IncMargin{1em}
\begin{algorithm}[h!]
\KwIn{$\bfY_{\mathrm{M}}$, $\bfY_{\mathrm{H}}$, $\NoiCovMat_{\mathrm{M}}$, $\NoiCovMat_{\mathrm{H}}$, $\bfR$, $\bfB$, $\bfS$, $\bfA$, $\mu>0$}
{\bf Initialization}: $\bfT^{(0)}$,${\bf G}^{(0)}$\;

\For{ $k = 0$  \KwTo stopping rule}
{
\tcc{Optimize w.r.t $\bfM$}
${\bfM}^{(t,k+1)} \in \argmin\limits_{\bfM} \calL({\bfM}, {\bfT}^{(k)}, {\bfG}^{(k)})$\;
\tcc{Optimize w.r.t $\bfT$}
${\bfT}^{(k+1)} \leftarrow \Pi_{\calT} (\NoiCovMat_{\mathrm{H}}^{-\frac{1}{2}}{\bfM}^{(t,k+1)}-{\bfG}^{(k)})$\;
\tcc{Update Dual Variable $\bfG$}
${\bfG}^{(k+1)} \leftarrow  \bfG^{(k)} - \left(\NoiCovMat_{\mathrm{H}}^{-\frac{1}{2}}{\bfM}^{(k+1)} -{\bfT}^{(k+1)}\right)$\;}
Set ${\bfM}^{(t+1)} = {\bfM}^{(t,k+1)}$\;
\KwOut{${\bfM}^{(t+1)}$}
\caption{ADMM sub-iterations to estimate $\bfM$}
\label{Algo:ADMM_M}
\end{algorithm}
\DecMargin{1em}

\subsubsection{Updating $\bfM$}
Forcing the derivative of \eqref{eq:min_M} w.r.t. $\bfM$ to be zero leads to the following Sylvester equation
\begin{equation}
\begin{split}
\label{eq:Sylvester_M}
\bfH_1 \bfM +\bfM \bfH_2 = \bfH_3\\ 
\end{split}
\end{equation}
where
\begin{equation*}
\begin{split}
\bfH_1&=\NoiCovMat_{\mathrm{H}}\bfR^T\NoiCovMat_{\mathrm{M}}^{-1}\bfR\\
\bfH_2&=\left({\bfA_{\mathrm{H}} \bfA_{\mathrm{H}}}^T+\mu\Id{p}\right)\left(\bfA\bfA^T\right)^{-1}\\
\bfH_3&=\\
&\left[\bfY_{\mathrm{H}}\bfA_{\mathrm{H}}^T+\NoiCovMat_{\mathrm{H}}\bfR^T\NoiCovMat_{\mathrm{M}}^{-1}\bfY_{\mathrm{M}}\bfA^T+\mu \NoiCovMat_{\mathrm{H}}^{\frac{1}{2}}\left(\bfT+\bfG\right)\right]\left(\bfA\bfA^T\right)^{-1}.
\end{split}
\end{equation*}
Note that $\textrm{vec}{\left(\bf AXB\right)}= \left(\bfB^T \otimes \bfA\right) \textrm{vec}(\bfX)$, where
$\textrm{vec}\left(\bfX\right)$ denotes the vectorization of the matrix $\bfX$ formed by stacking the 
columns of $\bfX$ into a single column vector and $\otimes$ denotes the Kronecker product \cite{Horn2012}.
Thus, vectorizing both sides of \eqref{eq:Sylvester_M} leads to\footnote{The vectorization of the matrics 
$\bfM$,$\bfH_1$ and $\bfH_2$ is easy to do as the size of these matrices are small, which is not true for 
the matrices $\bfA$, $\bfC_1$ and $\bfC_2$ in \eqref{eq:sylvester}.}
\begin{equation}
\bfW \textrm{vec}(\bfM)=\textrm{vec}(\bfH_3)
\label{eq:Vec_M_eq}
\end{equation}
where $\bfW=\left(\Id{p} \otimes \bfH_1 + \bfH_2^T \otimes \Id{m_{\lambda}}\right)$. 
Thus, $\textrm{vec}\left(\hat{\bfM}\right)=\bfW^{-1}\textrm{vec}(\bfH_3)$.
Note that $\bfW^{-1}$ can be computed and stored in advance instead of being computed in each iteration.



Alternatively, there exists a more efficient way to calculate the solution $\bfM$ analytically
(avoiding to compute the inverse of the matrix $\bfW$). Note that the 
matrices $\bfH_1 \in \mathbb{R}^{m_{\lambda} \times m_{\lambda}}$ and 
$\bfH_2\in \mathbb{R}^{p \times p}$ are both the products of two symmetric positive definite
matrices. According to the Lemma 1 in \cite{Wei2015FastFusion}, $\bfH_1$ 
and $\bfH_2$ can be diagonalized by eigen-decomposition,
i.e., $\bfH_1=\bfV_1 \bfO_1 \bfV_1^{-1}$ and $\bfH_2=\bfV_2 \bfO_2 \bfV_2^{-1}$, where
$\bfO_1$ and $\bfO_2$ are diagonal matrices denoted as
\begin{equation}
\begin{split}
\label{eq:D1_D2}
&\bfO_1=\textrm{diag}\{s_1,\cdots,s_{m_{\lambda}}\}\\
&\bfO_2=\textrm{diag}\{t_1,\cdots,t_p\}.
\end{split}
\end{equation}
Thus, \eqref{eq:Sylvester_M} can be transformed to
\begin{equation}
\begin{split}
\label{eq:Sylvester_special}
\bfO_1 \tilde{\bfM} + \tilde{\bfM} \bfO_2 = \bfV_1^{-1} \bfH_3 \bfV_2\\
\end{split}.
\end{equation}
where $\tilde{\bfM}=\bfV_1^{-1} \bfM \bfV_2$.
Straightforward computations lead to
\begin{equation}
\tilde{\bfH}  \circ \tilde{\bfM}= \bfV_1^{-1} \bfH_3 \bfV_2
\end{equation}
where
\begin{equation}
\tilde{\bfH}=\left[
\begin{array}{ccccc}
s_1+t_1& s_1+t_2 &\cdots &s_1+t_p\\
s_2+t_1& s_2+t_2 &\cdots &s_2+t_p\\
\vdots    & \vdots     &\ddots &\vdots\\
s_{m_{\lambda}}+t_1& s_{m_{\lambda}}+t_2 &\cdots &s_{m_{\lambda}+t_p}
\end{array}
\right]
\end{equation}
and $\circ$ represents the Hadamard product, defined as the component-wise product of two matrices (having the same size).
Then, $\tilde{\bfM}$ can be calculated by component-wise division of $\bfV_1^{-1} \bfH_3 \bfV_2$ and $\tilde{\bfH}$.
Finally, $\bfM$ can be estimated as $\hat{\bfM}=\bfV_1 \tilde{\bfM} \bfV_2^{-1}$.
Note that the computational complexity of the latter strategy is of order $\calO(\mathrm{max}(m_{\lambda}^3,p^3))$,
which is lower than the complexity order $\calO((m_{\lambda}p)^3))$ of solving \eqref{eq:Vec_M_eq}.
\subsubsection{Updating $\bfT$}
The optimization w.r.t. $\bfT$ can be transformed as
\begin{equation}
\label{eq:Update_M_T}
\argmin_{\bfT} \frac{1}{2} \big\|{\bfT}-\NoiCovMat_{\mathrm{H}}^{-\frac{1}{2}}{\bfM}+{\bfG}\big\| + \iota_{\calT}(\bfT)
\end{equation}
where $\iota_{\calT}(\bfT)=\iota_{\calM}(\NoiCovMat_{\mathrm{H}}^{\frac{1}{2}}\bfT)$.
As $\NoiCovMat_{\mathrm{H}}^{-\frac{1}{2}}$ is a diagonal matrix, the solution of
\eqref{eq:Update_M_T} can be obtained easily by setting
\begin{equation}
\hat{\bfT}= \NoiCovMat_{\mathrm{H}}^{-\frac{1}{2}} \min \left(\max\left(\bfM - \NoiCovMat_{\mathrm{H}}^{\frac{1}{2}}\bfG,0\right),1\right)
\end{equation}
where $\min$ and $\max$ are to be understood component-wise.

\begin{remark*}
If the endmember signatures are fixed a priori, i.e., $\bfM$ is known, the unsupervised unmixing and fusion will degenerate 
to a supervised unmixing and fusion by simply not updating $\bfM$. In this case, the alternating scheme is not
necessary, since Algorithm \ref{Algo:AlterOpti_FUMI} reduces to Algorithm \ref{Algo:ADMM_A}. Note that fixing $\bfM$ a priori
transforms the non-convex problem \eqref{eq:neglog_map_FUMI} into a convex one, which can be solved much more efficiently. 
The solution produced by the resulting algorithm is also guaranteed to be the global optimal point instead of a stationary point.
\end{remark*}

\section{Experimental results}
\label{sec:simu_FUMI}
This section applies the proposed unmixing based fusion method
to multi-band images associated with both synthetic and semi-real data.
All the algorithms have been implemented using MATLAB R2014A on a 
computer with Intel(R) Core(TM) i7-2600 CPU@3.40GHz and 8GB RAM.

\subsection{Quality metrics}
\label{subsec:performance}
\subsubsection{Fusion quality}
To evaluate the quality of the proposed fusion strategy, five
image quality measures have been investigated. Referring to \cite{Zhang2009,Wei2015TGRS},
we propose to use the restored signal to noise ratio (RSNR), the averaged spectral angle
mapper (SAM), the universal image quality index (UIQI), the relative dimensionless global
error in synthesis (ERGAS) and the degree of distortion (DD) as quantitative
measures. The larger RSNR and UIQI, the better the fusion. 
The smaller SAM, ERGAS and DD, the better the fusion.

\subsubsection{Unmixing quality}
In order to analyze the quality of the unmixing results, we consider the
normalized mean square error (NMSE) for both endmember and abundance matrices
\begin{align*}
    \textrm{NMSE}_{\bfM} = \frac{\|\widehat{\bf M}- {\bf M}\|^2_F}{\| {\bf M}\|^2_F}\\
    \textrm{NMSE}_{\bfA} = \frac{\|\widehat{\bf A}- {\bf A}\|^2_F}{\| {\bf A}\|^2_F}.
\end{align*}
The smaller NMSE, the better the quality of the unmixing.
The SAM between the actual and estimated endmembers (different
from SAM defined previously for pixel vectors) is a measure of
spectral distortion defined as
\begin{equation*}
\textrm{SAM}_{\bfM}(\bsm_n,\hat{\bsm}_n)=\textrm{arccos} \left(\frac{\langle\bsm_n,\hat{\bsm}_n\rangle}{ \|\bsm_n\|_2\|\hat{\bsm}_n\|_2}\right).
\label{eq:SAM}
\end{equation*}
The overall SAM is finally obtained by averaging the SAMs computed from all endmembers.
Note that the value of SAM is expressed in degrees and thus belongs to $(-90,90]$.
The smaller the absolute value of SAM, the less important the spectral distortion.

\subsection{Synthetic data}
\label{subsec:simu_syn_data}
This section applies the proposed FUMI method to synthetic data
and compares it with the joint unmixing and fusion methods investigated 
in \cite{Berne2010} and \cite{Yokoya2012coupled}. 
Note that the method studied in \cite{Berne2010} can be regarded as a 
one-step version of \cite{Yokoya2012coupled}. 

The reference endmembers are $m$ reflectance spectra selected  
randomly from the United States Geological Survey (USGS) digital spectral
library\footnote{http://speclab.cr.usgs.gov/spectral.lib06/}.
Each reflectance spectrum consists of $L = 224$ spectral bands
from $383$ nm to $2508$ nm. 
In this simulation, the number of endmembers is fixed to $p=5$.
The abundances $\bfA$ are generated according to a Dirichlet distribution 
over the simplex defined by the ANC and ASC constraints.
There is one vector of abundance per pixel, i.e., $\bfA \in \mathbb{R}^{5 \times 100^2}$,
for the considered image of size $100 \times 100$ pixels.
The synthetic image is then generated by the product of endmembers and
abundances, i.e., $\bf X= MA$.


\begin{itemize}

\item \textbf{Initialization}: As shown in Algorithm \ref{Algo:AlterOpti_FUMI}, the proposed algorithm only requires the
initialization of the endmember matrix $\bfM$. Theoretically, any endmember extraction algorithm (EEA) can be used 
to initialize $\bfM$. In this work, we have used the \emph{simplex
identification via split augmented Lagrangian} (SISAL) method \cite{Bioucas2009SISAL}, 
which is a state-of-the-art method that does not require the presence
of pure pixels in the image.

\item \textbf{Subspace Identification}: 
For the endmember estimation, a popular strategy is to use a subspace transformation as a preprocessing step,
such as in \cite{Bioucas2008,Dobigeon2009}. In general, the subspace transformation is estimated \emph{a priori}
from the high-spectral resolution image, e.g., from the HS data. 
In this work, the projection matrix has been learned by 
computing the singular value decomposition (SVD) of 
$\bfY_{\mathrm{H}}$ and retaining the left-singular vectors associated 
with the largest eigenvalues.
Then the input HS data $\bfY_{\mathrm{H}}$, the HS noise covariance matrix $\NoiCovMat_{\mathrm{H}}$
and the spectral response $\bfR$ in Algorithm \ref{Algo:AlterOpti_FUMI} are replaced with their projections 
onto the learned subspace as $\bfY_{\mathrm{H}} \leftarrow \bfE^T \bfY_{\mathrm{H}}$,
$\NoiCovMat_{\mathrm{H}} \leftarrow \bfE^T \NoiCovMat_{\mathrm{H}} \bfE$
and $\bfR \leftarrow \bf RE$, where $\bfE \in \bfR^{\nbbandima \times \wtm_{\lambda}}$ is the 
estimated orthogonal basis using SVD and $\wtm_{\lambda} \ll \nbbandima$.
Given that the formulation using the transformed entities is equivalent to the original one but 
the matrix dimension is now much smaller, the subspace transformation brings huge numerical advantage.

\item \textbf{Parameters in ADMM}: The value of $\mu$ adopted in all the experiments is fixed to the average 
of the noise power of HS and MS images, which is motivated by balancing the data term and regularization term.
As ADMM is used to solve sub-problems, it is not necessary to use complicated stopping rule to run ADMM
exhaustively. Thus, the number of ADMM iterations has been fixed to $30$. Experiments have demonstrated that varying 
these parameters do not affect much the convergence of the whole algorithm.

%
\item \textbf{Stopping rule}: 
The stopping rule for Algorithm \ref{Algo:AlterOpti_FUMI} is that the relative difference for
the successive updates of the objective $L(\bfM,\bfA)$ is less than $10^{-4}$, i.e.,
\begin{equation*}
\frac{|L(\bfM^{(t+1)},\bfA^{(t+1)})-L(\bfM^{(t)},\bfA^{(t)})|}{|L(\bfM^{(t)},\bfA^{(t)})|}\leq 10^{-4}.
\end{equation*}
\end{itemize}

\subsubsection{HS and MS image fusion}
\label{subsubsec:HS+MS+Pure}
In this section, we consider the fusion of HS and MS images. 
The HS image $\bfY_{\mathrm{H}}$ has been generated by applying a $7 \times 7$ 
Gaussian filter (with zero mean and standard deviation $\sigma_{\bfB} =1.7$) and
then by down-sampling every $4$ pixels in both vertical and horizontal directions 
for each band of the reference image. A $7$-band MS image $\bfY_{\mathrm{M}}$ has 
been obtained by filtering $\bfX$ with the LANDSAT-like reflectance spectral 
responses. 
The HS and MS images are both contaminated by zero-mean additive Gaussian noises.
Our simulations have been conducted with $\textrm{SNR}_{\mathrm{H},i}=50$dB for all
the HS bands with $\textrm{SNR}_{\mathrm{H},i}=10\log \left(\frac{\|\left(\MATima \bf BS\right)_i\|_F^2}{\noisevar{\mathrm{H},i}}\right)$.
For the MS image $\textrm{SNR}_{\mathrm{M},j}=10\log \left( \frac{\|\left({\bfR} \MATima\right)_j\|_F^2}{\noisevar{\mathrm{M},j}}\right) = 50\textrm{dB}$
for all spectral bands.

As the endmembers are selected randomly from the USGS library, 30 Monte Carlo simulations have been 
implemented and all the results have been obtained by averaging these 30 Monte Carlo runs. The fusion and
unmixing results using different methods are reported in Tables \ref{tb:MS_HS_pure_Q_fusion} and 
\ref{tb:MS_HS_pure_Q_unmixing}, respectively. For fusion performance, the proposed FUMI method 
outperforms the other two methods, with a competitive time complexity. Regarding unmixing,
Berne's method and FUMI perform similarly for endmember estimation, both much better than Yokoya's.
In terms of abundance estimation, FUMI outperforms the other methods.

\begin{table}[h!]
\renewcommand{\arraystretch}{1.1}
\centering \caption{Fusion Performance for Synthetic {HS+MS} dataset: RSNR (in dB), UIQI, SAM (in degree), ERGAS, DD (in $10^{-3}$) and time (in second).}
\vspace{0.1cm}
\begin{tabular}{|c|cccccc|}
\hline
 Methods & RSNR & UIQI & SAM  & ERGAS & DD & Time \\
\hline
\hline
Berne2010 &  48.871 & 0.9995 & 0.169 & 0.1011 &1.404 & \Best{8.13}\\
\hline
Yokoya2012 &  48.278 &  0.9995 & 0.188 & 0.1077 &1.513& 29.95\\
\hline
Proposed FUMI &  \Best{50.100} &  \Best{0.9996} & \Best{0.146} & \Best{0.0877} &\Best{1.235} & 8.50\\
\hline
\end{tabular}
\label{tb:MS_HS_pure_Q_fusion}
\end{table}

\begin{table}[h!]
\renewcommand{\arraystretch}{1.1}
\centering \caption{Unmixing Performance for Synthetic {HS+MS} dataset: SAM$_\bfM$ (in degree), NMSE$_\bfM$ (in dB) and NMSE$_\bfA$ (in dB).}
\vspace{0.1cm}
\begin{tabular}{|c|ccc|}
\hline
 Methods & SAM$_\bfM$ & NMSE$_\bfM$ & NMSE$_\bfA$ \\
\hline
\hline
Berne2010 &  \Best{0.549}  & -39.44  & -18.22\\
\hline
Yokoya2012 &  1.443 &  -31.91 & {-13.97}\\
\hline
Proposed FUMI &  0.690 &  \Best{-39.71} & \Best{-22.44}\\
\hline
\end{tabular}
\label{tb:MS_HS_pure_Q_unmixing}
\end{table}

\subsubsection{HS and PAN image fusion}
When the number of MS bands degrade to one, 
the fusion of HS and MS degenerates to HS pansharpening, which
is a more challenging problem. In this experiment, the PAN image is obtained by averaging the 
first 50 bands of the reference image. The quantitative results obtained after averaging 
30 Monte Carlo runs for fusion and unmixing are summarized in Tables \ref{tb:HS_PAN_pure_Q_fusion} 
and \ref{tb:HS_PAN_pure_Q_unmixing}, respectively. In terms of fusion performance,
the proposed FUMI method performs the best for all the quality measures,
with the least CPU time. Regarding the unmixing performance, Berne's method
gives the best estimation for endmembers whereas FUMI gives best abundance estimations.


\begin{table}[h!]
\renewcommand{\arraystretch}{1.1}
\centering \caption{Fusion Performance for Synthetic {HS+PAN} dataset: RSNR (in dB), UIQI, SAM (in degree), ERGAS, DD (in $10^{-3}$) and time (in second).}
\vspace{0.1cm}
\begin{tabular}{|c|cccccc|}
\hline
 Methods & RSNR & UIQI & SAM  & ERGAS & DD & Time \\
\hline
\hline
Berne2010 & 32.34  & 0.9887 & 0.669 & 0.682 &6.776 & {6.74}\\
\hline
Yokoya2012 & 33.00 &  0.9901 & 0.592 & 0.633 &6.072 & 11.65\\
\hline
Proposed FUMI &  \Best{36.16} &  \Best{0.9960} & \Best{0.399} & \Best{0.458} &\Best{3.899} & \Best{6.36}\\
\hline
\end{tabular}
\label{tb:HS_PAN_pure_Q_fusion}
\end{table}

\begin{table}[h!]
\renewcommand{\arraystretch}{1.1}
\centering \caption{Unmixing Performance for Synthetic {HS+PAN} dataset: SAM$_\bfM$ (in degree), NMSE$_\bfM$ (in dB) and NMSE$_\bfA$ (in dB).}
\vspace{0.1cm}
\begin{tabular}{|c|ccc|}
\hline
 Methods & SAM$_\bfM$ & NMSE$_\bfM$ & NMSE$_\bfA$ \\
\hline
\hline
Berne2010 &  \Best{0.566}  & \Best{-39.03}  & -16.38 \\
\hline
Yokoya2012 &  1.543 &  -29.31 & -14.09\\
\hline
Proposed FUMI &  0.716 &  {-38.07} & \Best{-18.49}\\
\hline
\end{tabular}
\label{tb:HS_PAN_pure_Q_unmixing}
\end{table}

\subsection{Semi-real data}
In this section, we test the proposed FUMI algorithm on semi-real 
datasets, for which we have the real HS image as the reference image and 
have simulated the degraded images from the reference image. 
 
\subsubsection{Moffett dataset}
\begin{figure*}[h!]
\centering
    \subfloat{
    \includegraphics[width=0.25\textwidth]{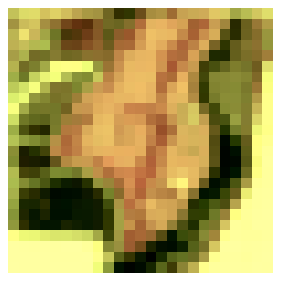}}
    \subfloat{
    \includegraphics[width=0.25\textwidth]{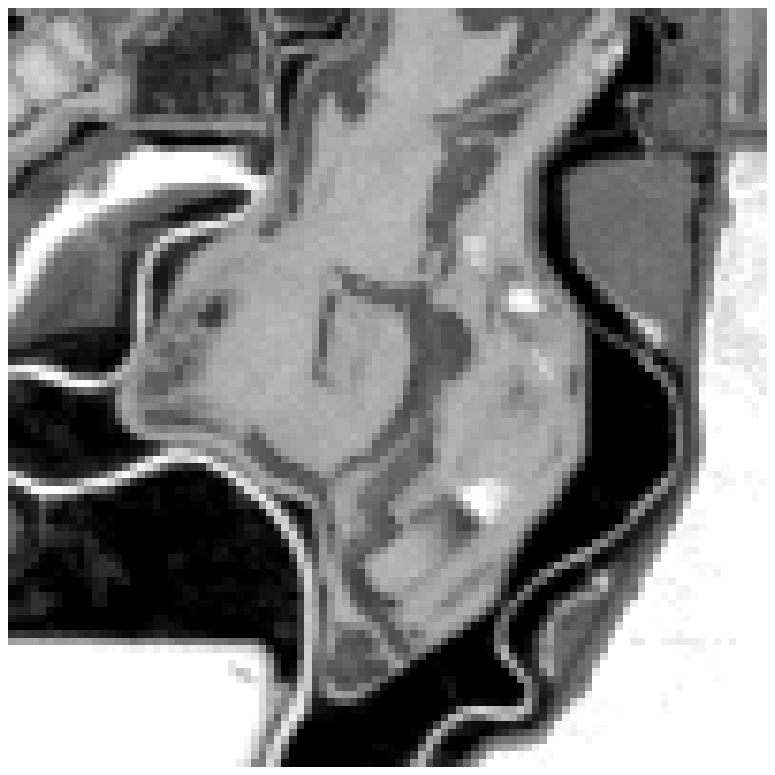}}
    \subfloat{
    \includegraphics[width=0.25\textwidth]{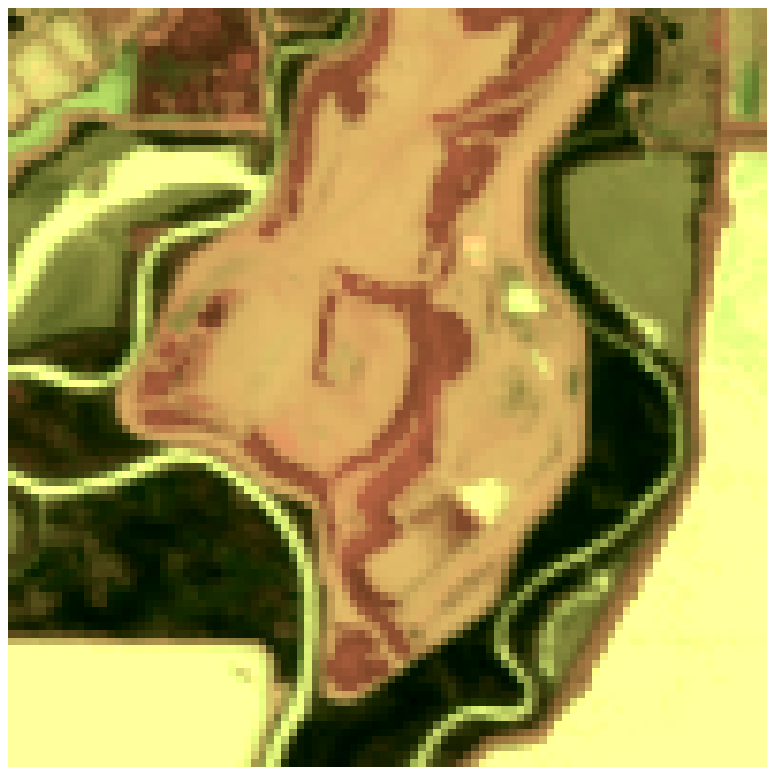}}\\
    \subfloat{
    \includegraphics[width=0.25\textwidth]{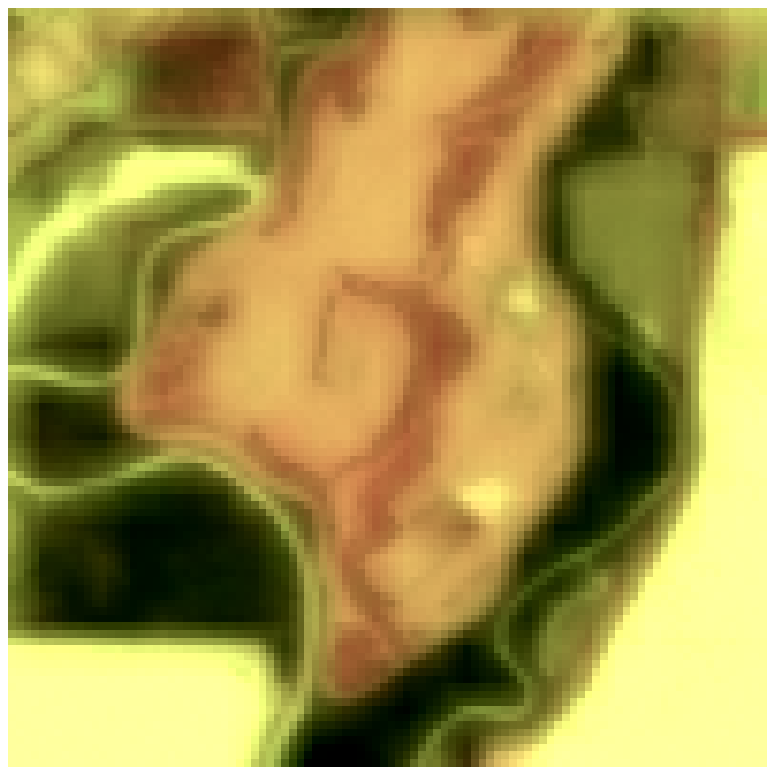}}
    \subfloat{
    \includegraphics[width=0.25\textwidth]{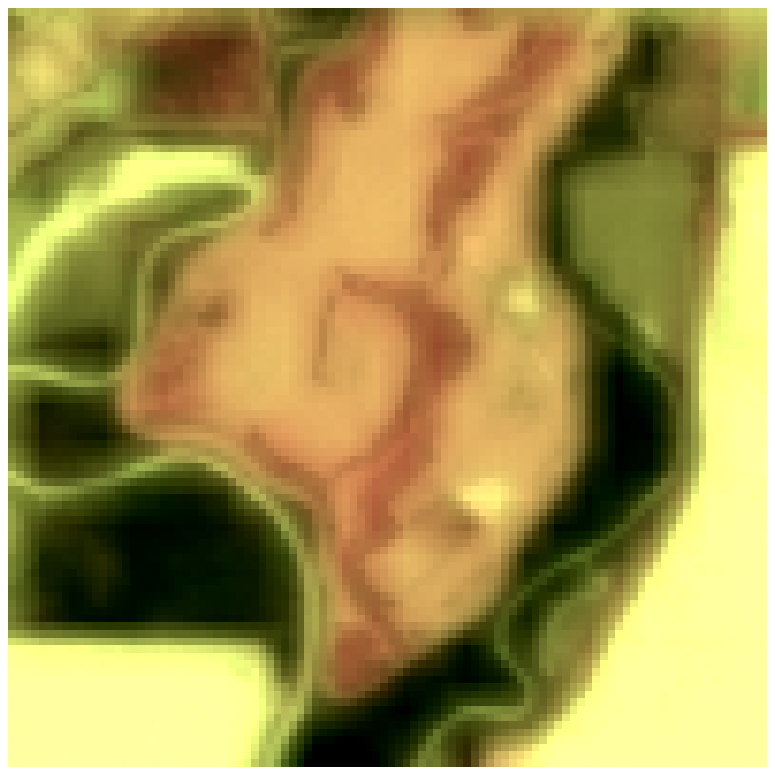}}
    \subfloat{
    \includegraphics[width=0.25\textwidth]{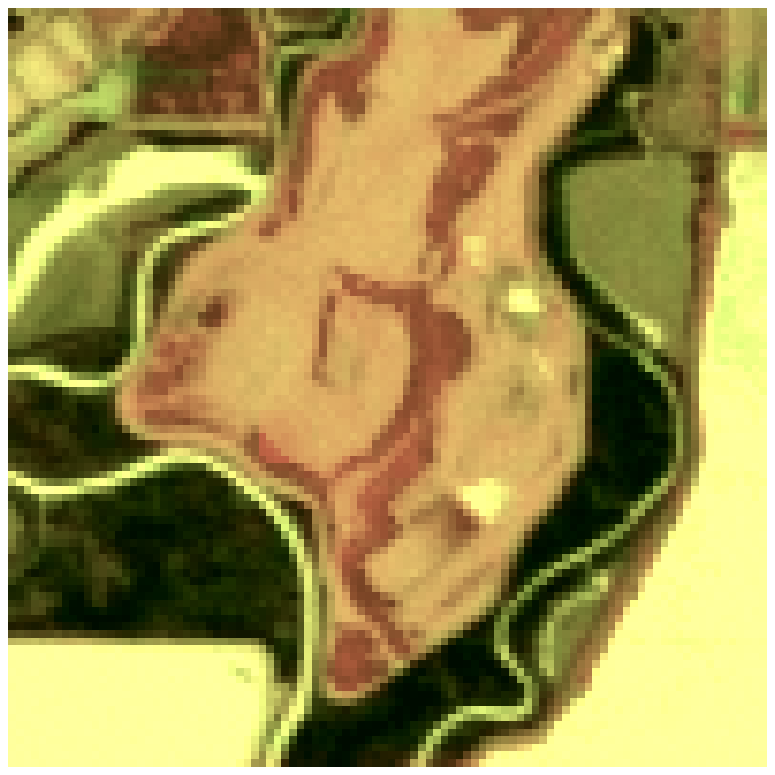}}
    \subfloat{
    \includegraphics[width=0.25\textwidth]{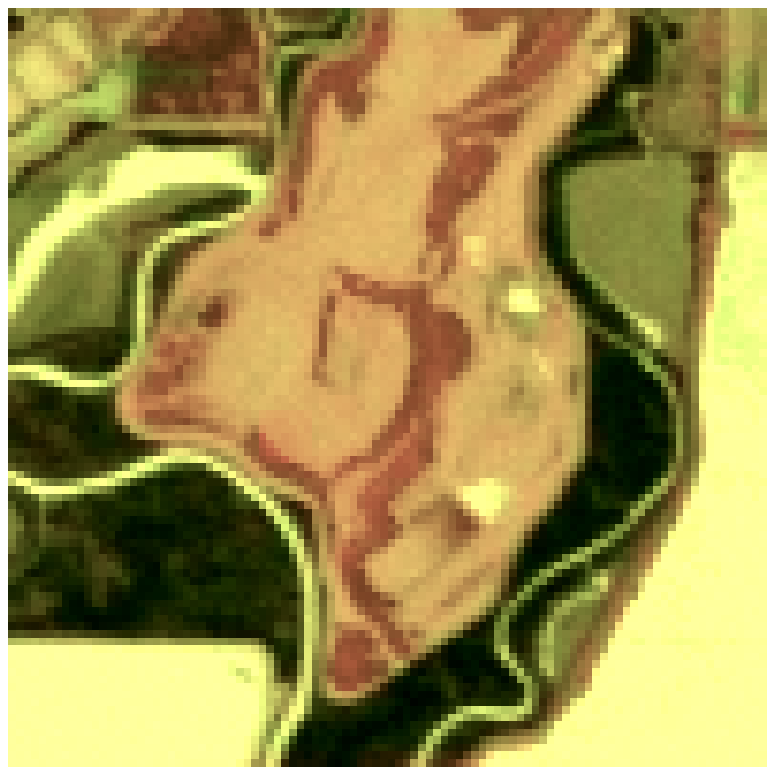}}\\
    \subfloat{
    \includegraphics[width=0.25\textwidth]{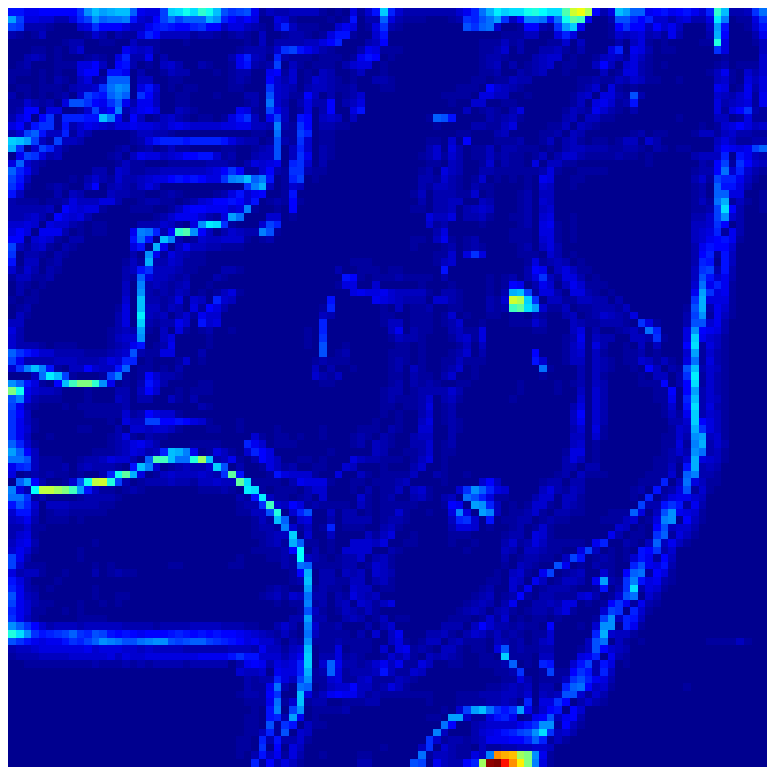}}
    \subfloat{
    \includegraphics[width=0.25\textwidth]{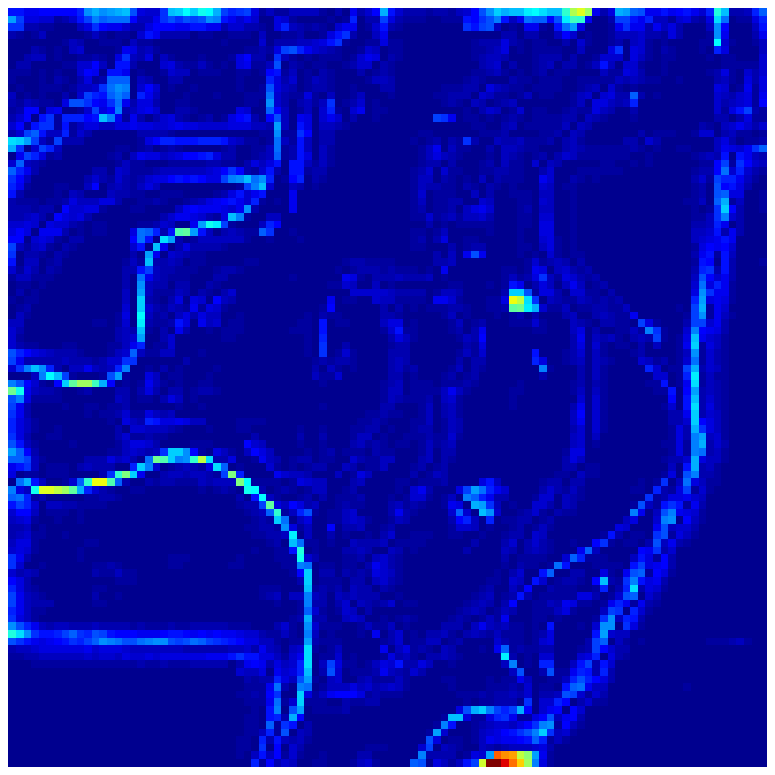}}
    \subfloat{
    \includegraphics[width=0.25\textwidth]{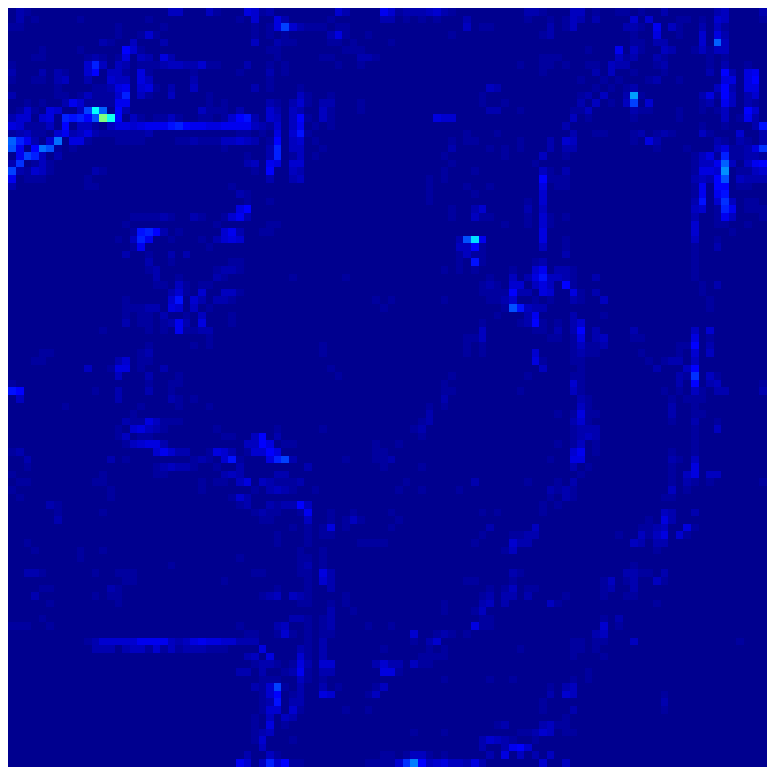}}
    \subfloat{
    \includegraphics[width=0.25\textwidth]{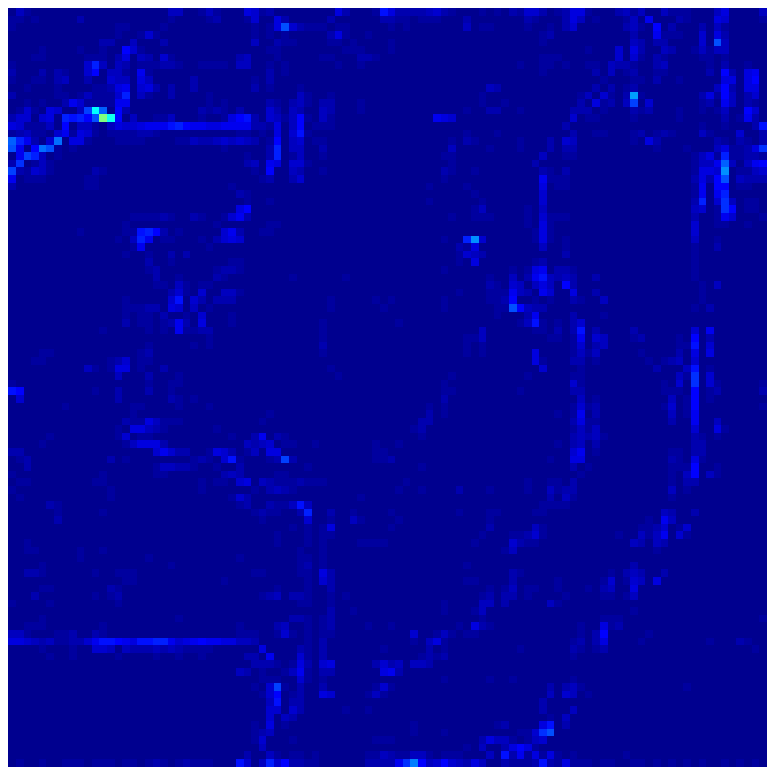}}
    \caption{Hyperspectral pansharpening results (Moffett dataset): (Top 1) HS image. (Top 2) MS image. (Top 3) Reference image. (Middle 1) Berne's method. (Middle 2) Yokoya's method. (Middle 3) S-FUMI (Middle 4) UnS-FUMI. (Bottom 1-4) The corresponding RMSE maps.}
\label{fig:HS_PAN_Moffet}
\end{figure*}

\begin{figure}[h!]
\centering
\includegraphics[width=0.5\textwidth]{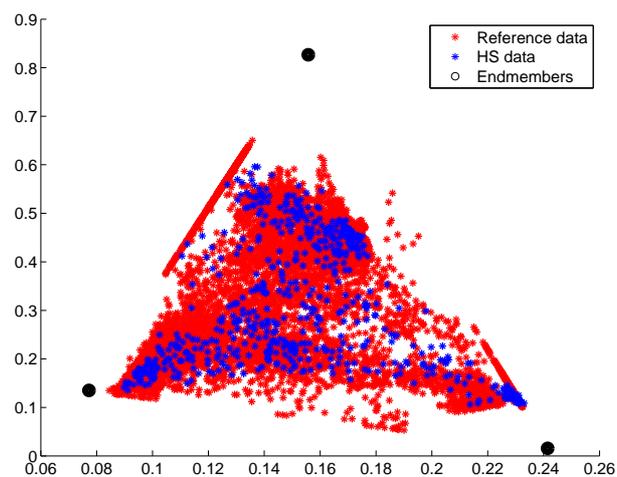}
\caption{Scattered Moffett data: The $1$st and the $100$th bands are selected as the coordinates.}
\label{fig:scatter_data_Moffet}
\end{figure}

In this experiment, 
the reference image is an HS image of size $100 \times 100 \times 176$
acquired over Moffett field, CA, in 1994 by the JPL/NASA airborne visible/infrared
imaging spectrometer (AVIRIS) \cite{Green1998imaging}. This image
was initially composed of $224$ bands that have been
reduced to $176$ bands after removing the water vapor absorption bands.
A composite color image of the scene of interest is shown in the top right of Fig. \ref{fig:HS_PAN_Moffet}
and its scattered data have been displayed as the red points in Fig. \ref{fig:scatter_data_Moffet}.
As there is no ground truth for endmembers and abundances for the reference image, we have first
unmixed this image (with any unsupervised unmixing method) and then reconstructed the reference 
image $\MATima$ with the estimated endmembers and abundances (after appropriate normalization).
The number of endmembers has been fixed to $p=3$ empirically as in \cite{Dobigeon2009}.

The reference image $\MATima$ is reconstructed from one HS and one coregistered 
PAN images. The observed HS image has been generated by
applying a $7 \times 7$ Gaussian filter with zero mean and standard deviation 
$\sigma_{\bfB}=1.7$ and by down-sampling every $4$ pixels in both vertical and horizontal directions 
for each band of $\MATima$, as done in Section \ref{subsubsec:HS+MS+Pure}. 
In a second step, the PAN image has been obtained by averaging the first
50 HS bands. The HS and PAN images are both contaminated by additive Gaussian noises,
whose SNRs are $50$dB for all the bands. The scattered data are displayed in 
Fig. \ref{fig:scatter_data_Moffet}, showing that there is no pure pixel in the degraded HS image.

To analyze the impact of endmember estimation, the proposed FUMI method has been implemented 
in two scenarios: estimating $\bfA$ with fixed $\bfM$, referred to as \emph{supervised FUMI} (S-FUMI) and 
estimating $\bfA$ and $\bfM$ jointly, referred to as \emph{unsupervised FUMI} (UnS-FUMI). In this work,
the S-FUMI algorithm has been run with a matrix $\bfM$ obtained using SISAL. 

The proposed FUMI algorithm (including both S-FUMI and UnS-FUMI) and other state-of-the-art methods 
have been implemented to fuse the two observed images and to unmix the HS image. The fusion results and
RMSE maps (averaged over all the bands) are shown in Figs. \ref{fig:HS_PAN_Moffet}. Visually, S-FUMI and UnS-FUMI give better fused images than
the other methods. This result is confirmed by the RMSE maps, where the two FUMI methods offer much smaller
errors than the other two methods. Furthermore, the quantitative fusion results reported in Table \ref{tb:Moffet_Q_fusion} 
are consistent with this conclusion as S-FUMI and UnS-FUMI outperform the other methods for all the fusion metrics. 
Regarding the computation time, S-FUMI and UnS-FUMI cost more than the other two methods, mainly due to the alternating 
update of the endmembers and abundances and also the ADMM updates within the alternating updates.

\begin{figure*}[h!]
\centering
	\subfloat{
	\includegraphics[width=0.4\textwidth]{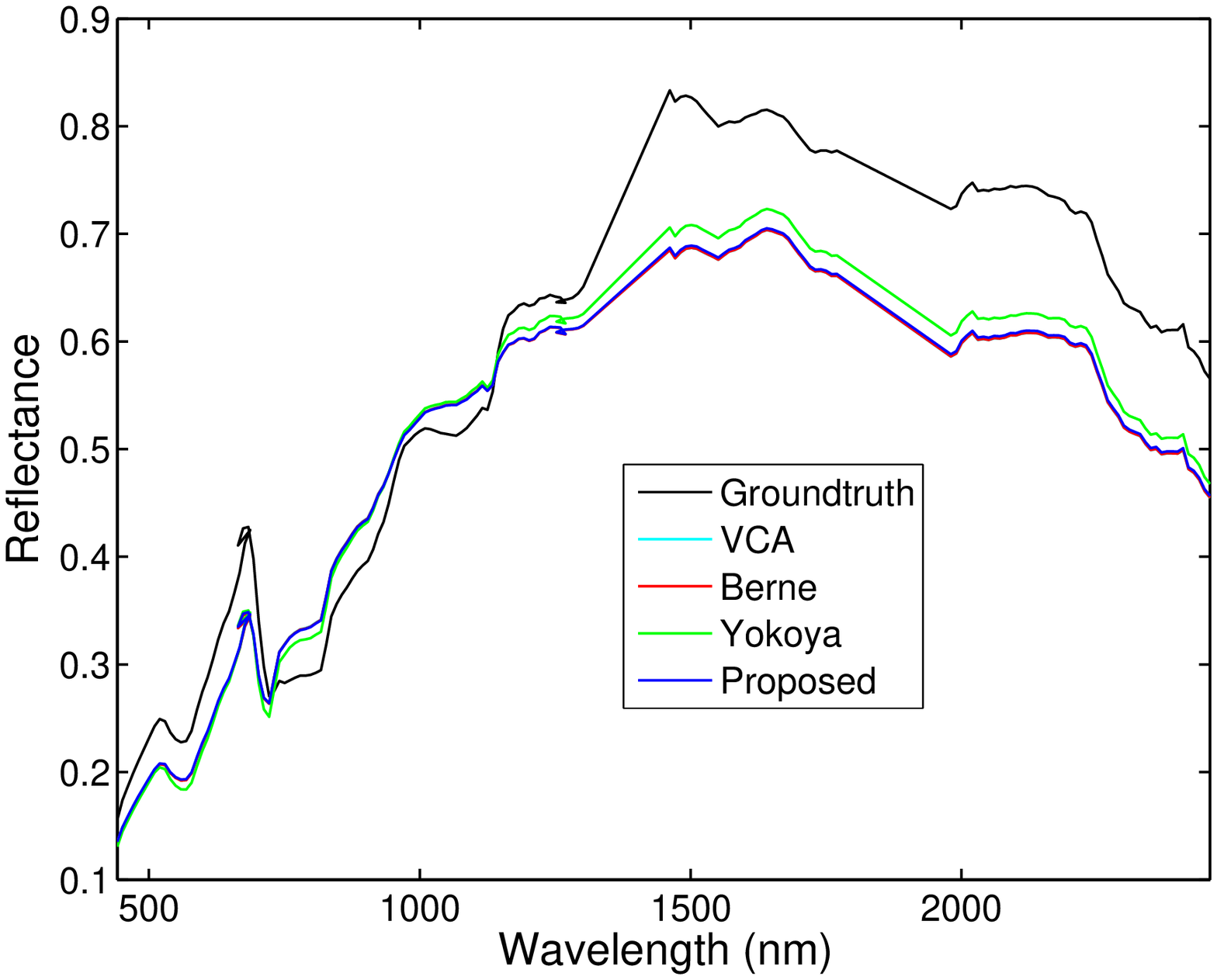}}
	\subfloat{
	\includegraphics[width=0.4\textwidth]{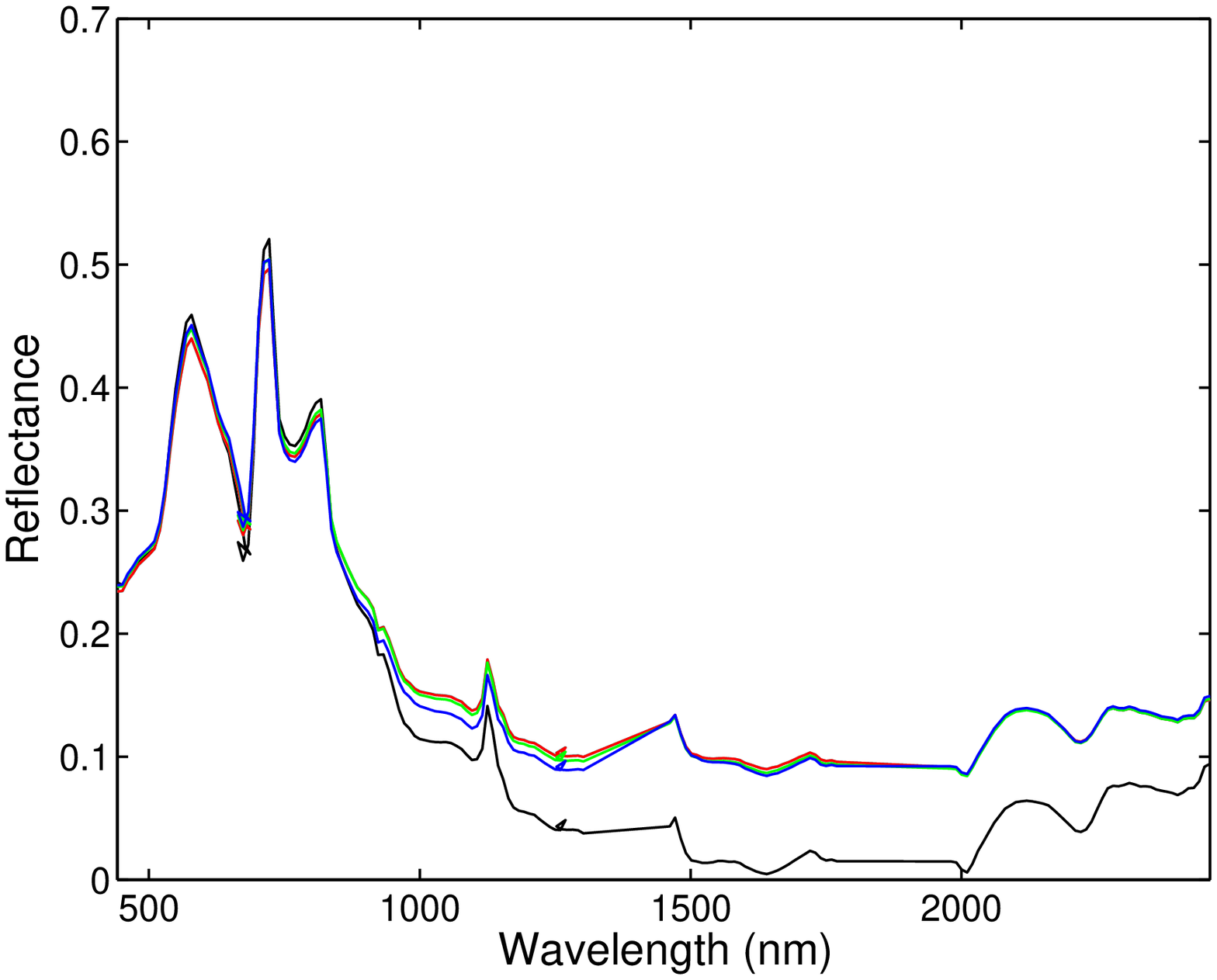}}\\
	\subfloat{
	\includegraphics[width=0.4\textwidth]{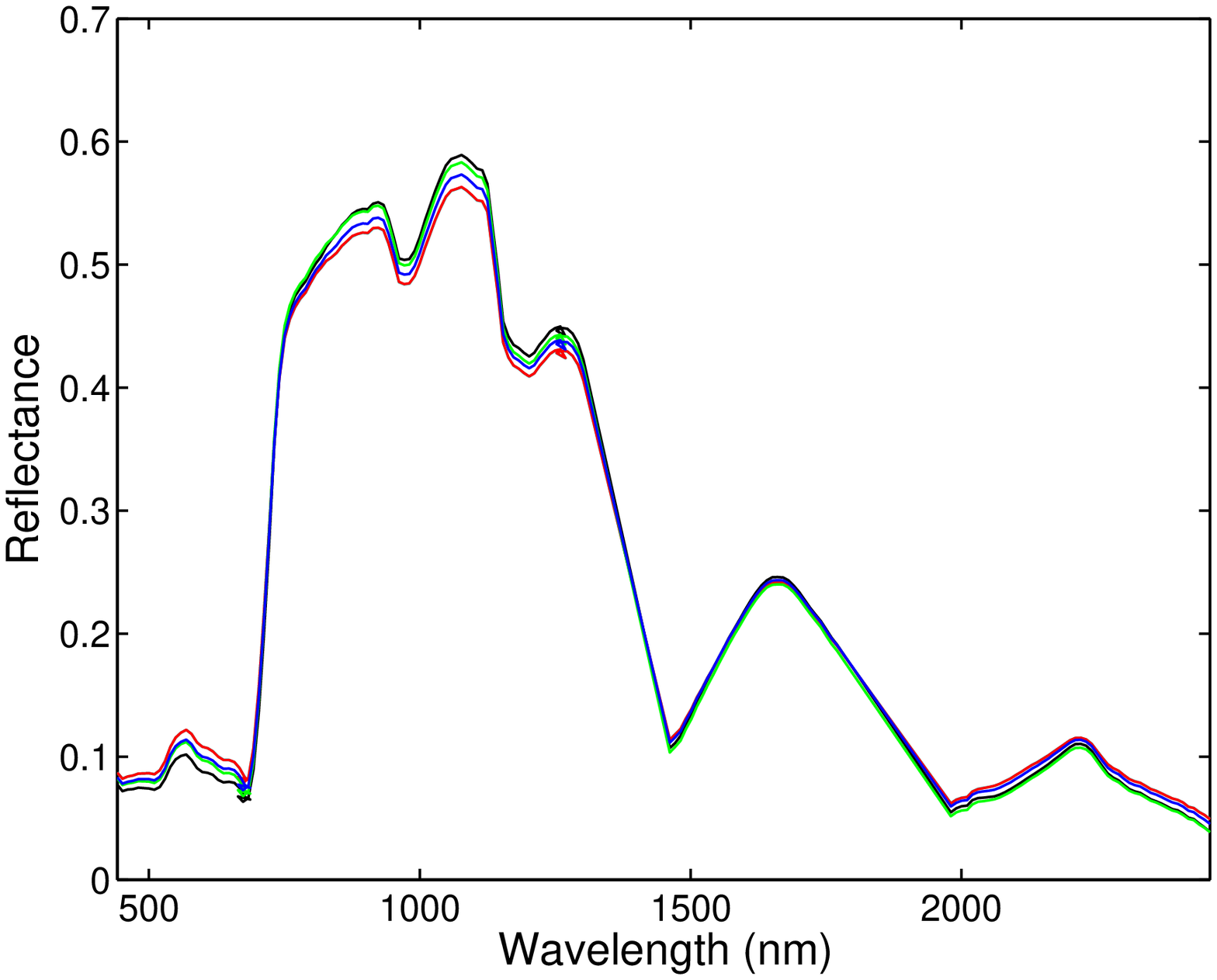}}
	\subfloat{
	\includegraphics[width=0.4\textwidth]{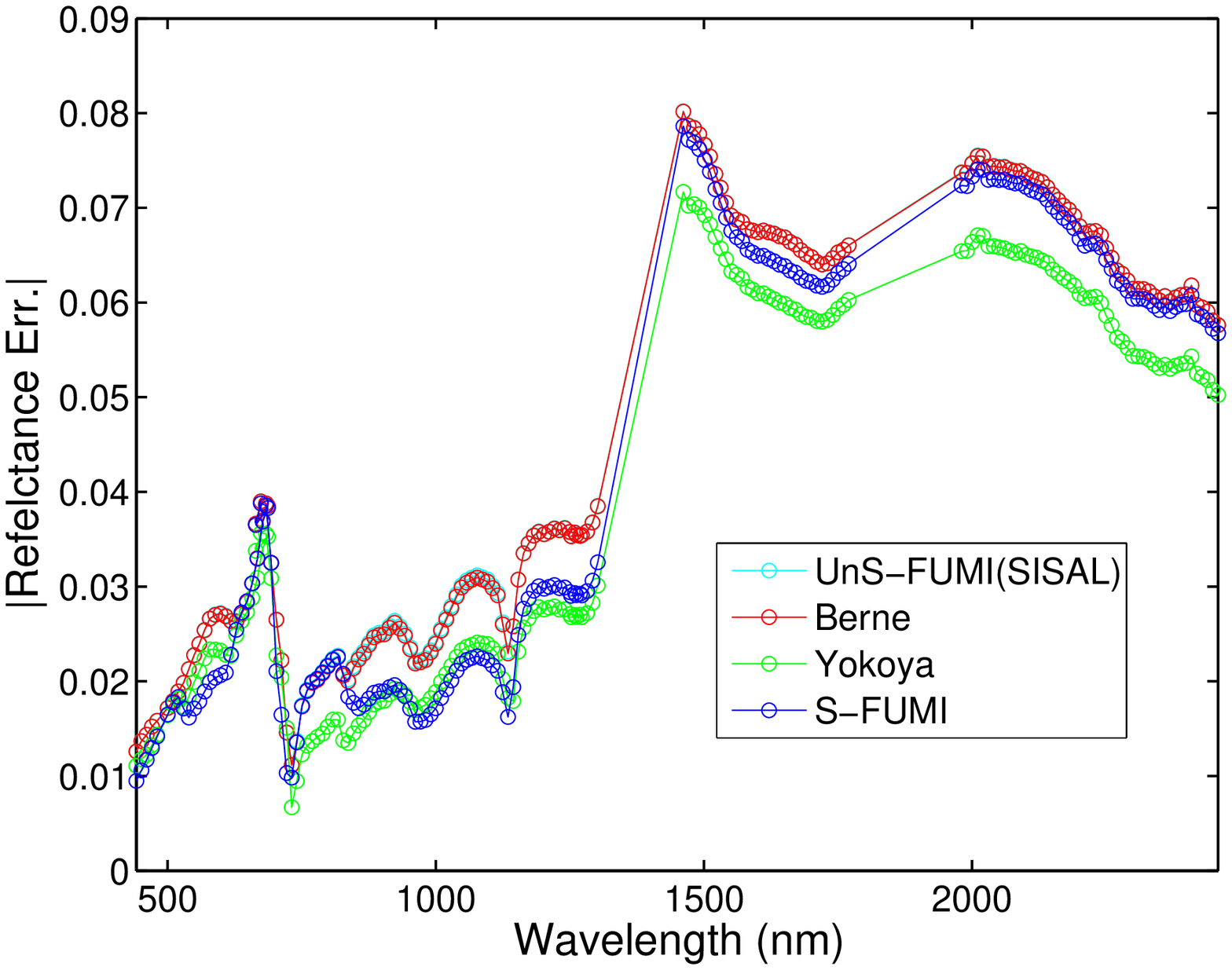}}
	\caption{Unmixed endmembers for Moffett HS+PAN dataset: (Top and bottom left) Estimated three endmembers and ground truth. (Bottom right) Sum of absolute value of all endmember errors as a function of wavelength. }
\label{fig:Moffet_End}
\end{figure*}

\begin{figure*}[h!]
\centering
 	\subfloat{
    \includegraphics[width=0.25\textwidth]{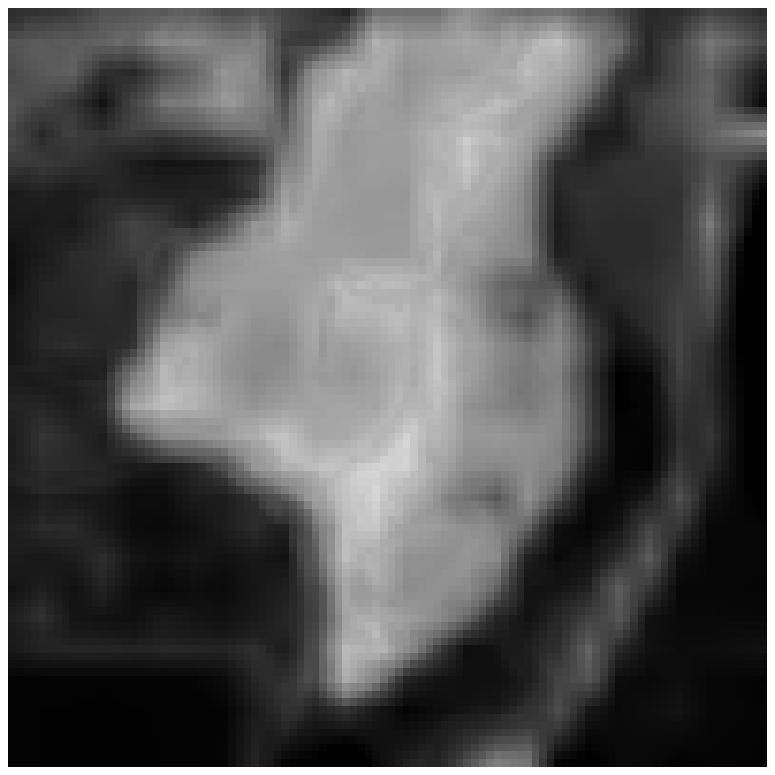}}
    \subfloat{
    \includegraphics[width=0.25\textwidth]{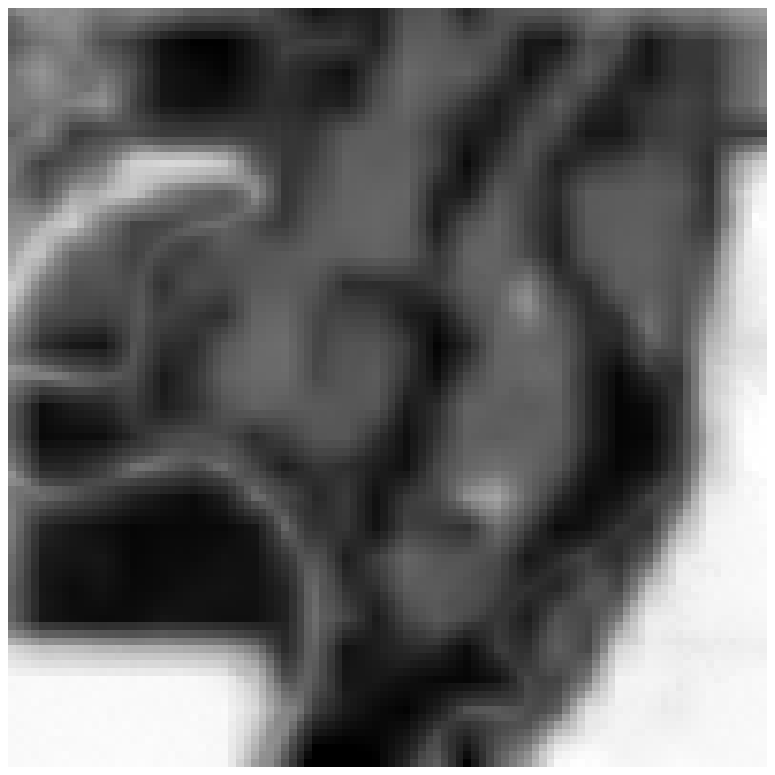}}
    \subfloat{
    \includegraphics[width=0.25\textwidth]{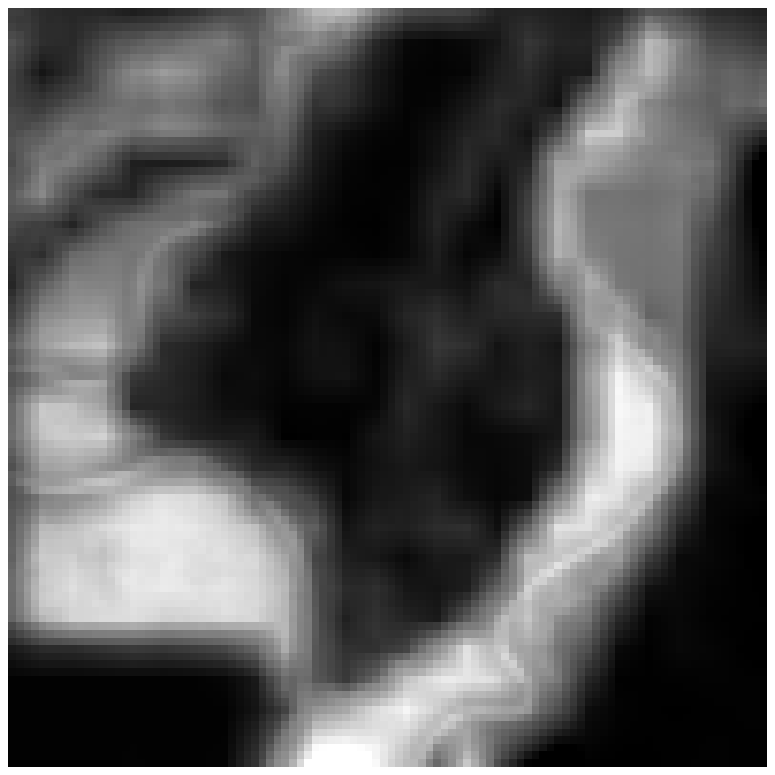}}\\
    \subfloat{
    \includegraphics[width=0.25\textwidth]{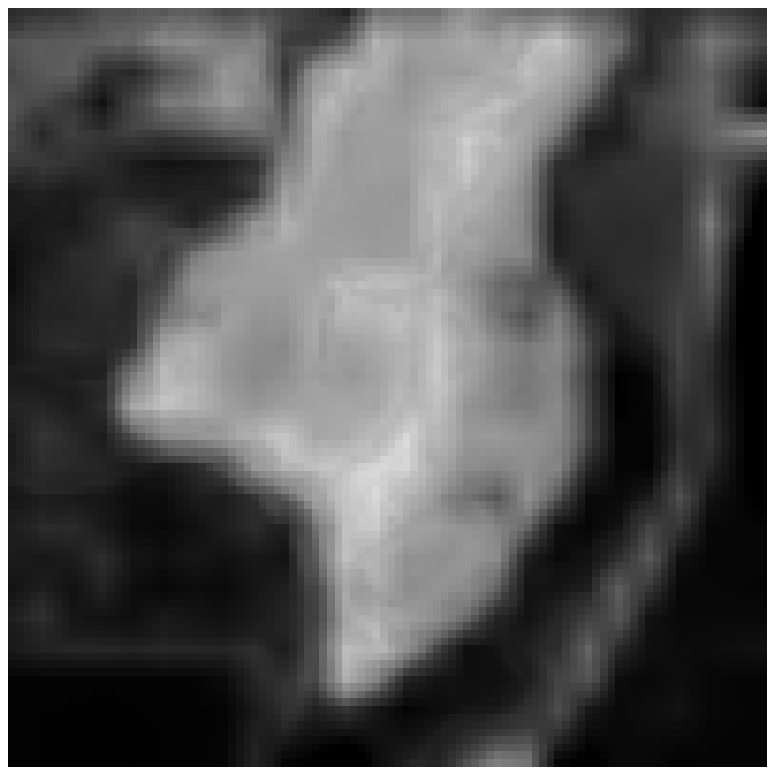}}
    \subfloat{
    \includegraphics[width=0.25\textwidth]{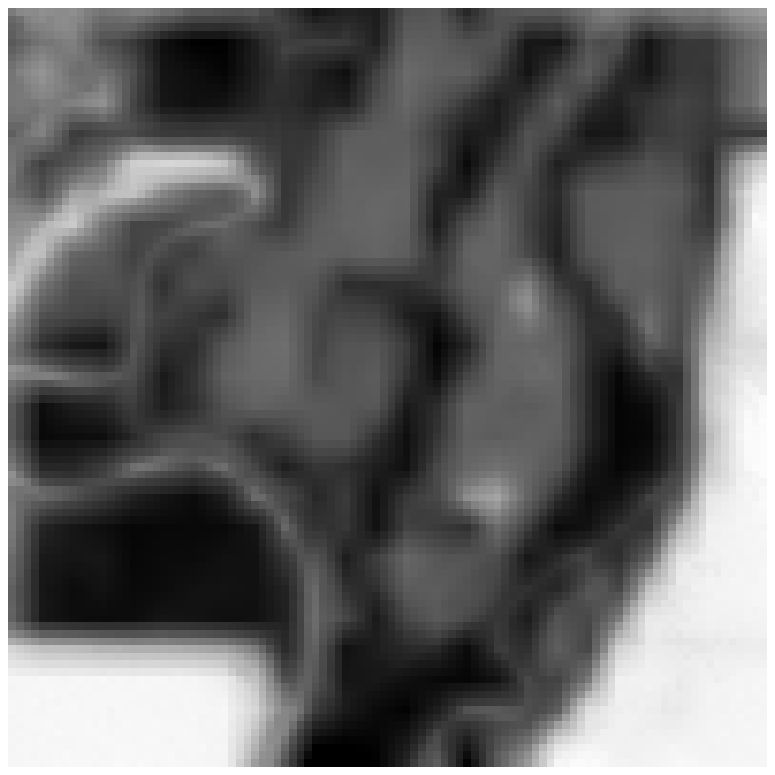}}
    \subfloat{
    \includegraphics[width=0.25\textwidth]{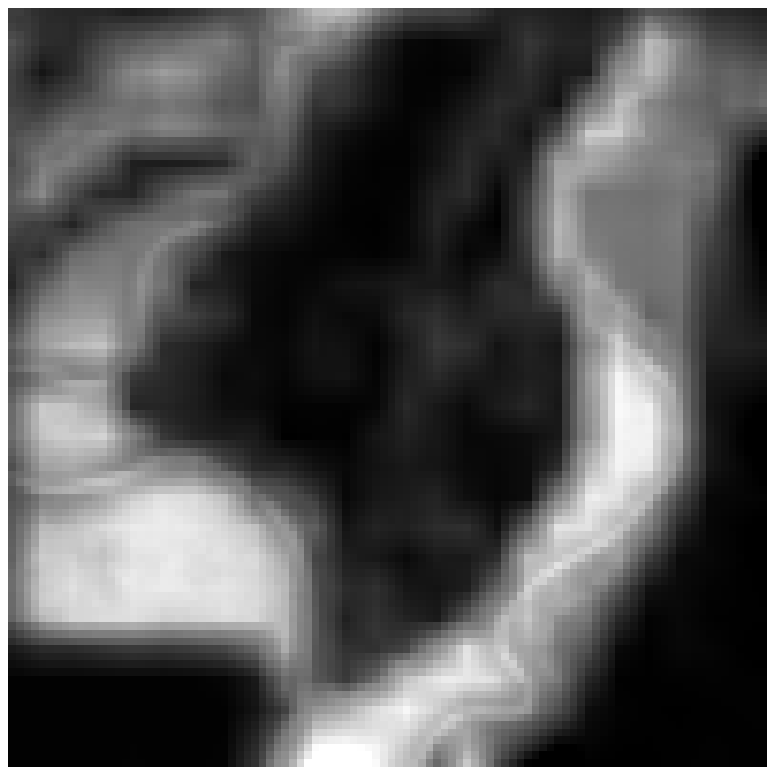}}\\
    \subfloat{
    \includegraphics[width=0.25\textwidth]{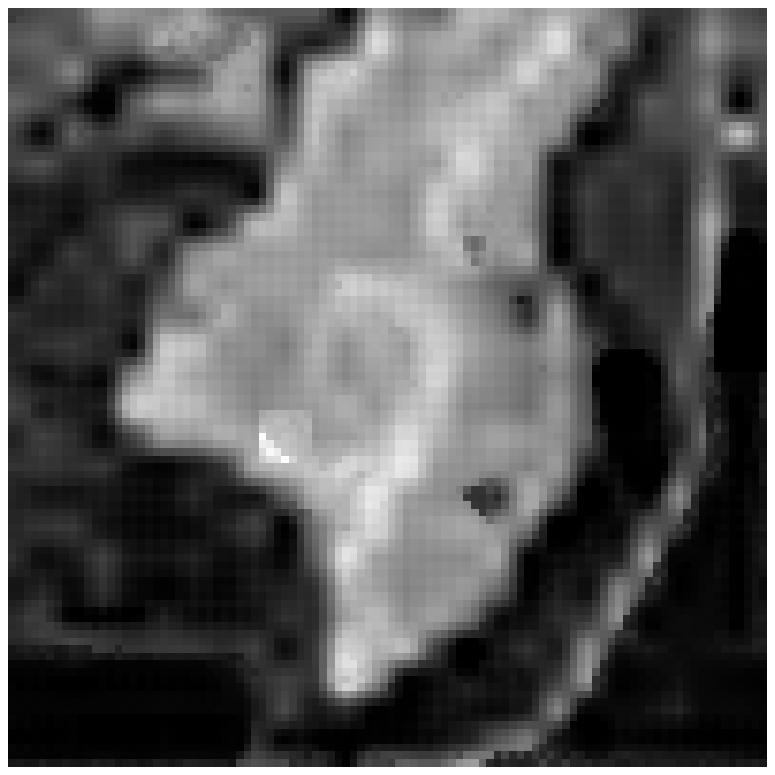}}
    \subfloat{
    \includegraphics[width=0.25\textwidth]{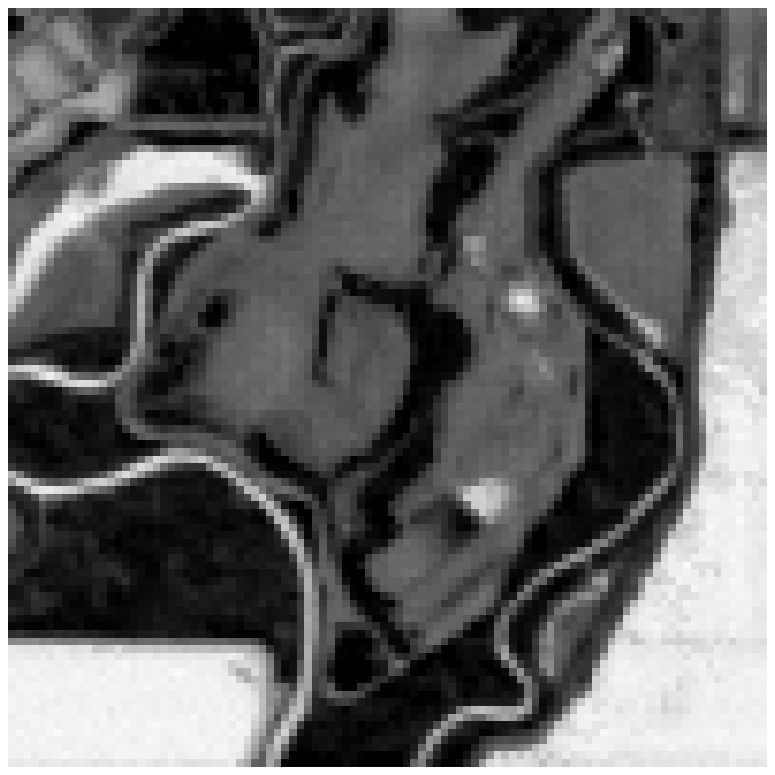}}
    \subfloat{
    \includegraphics[width=0.25\textwidth]{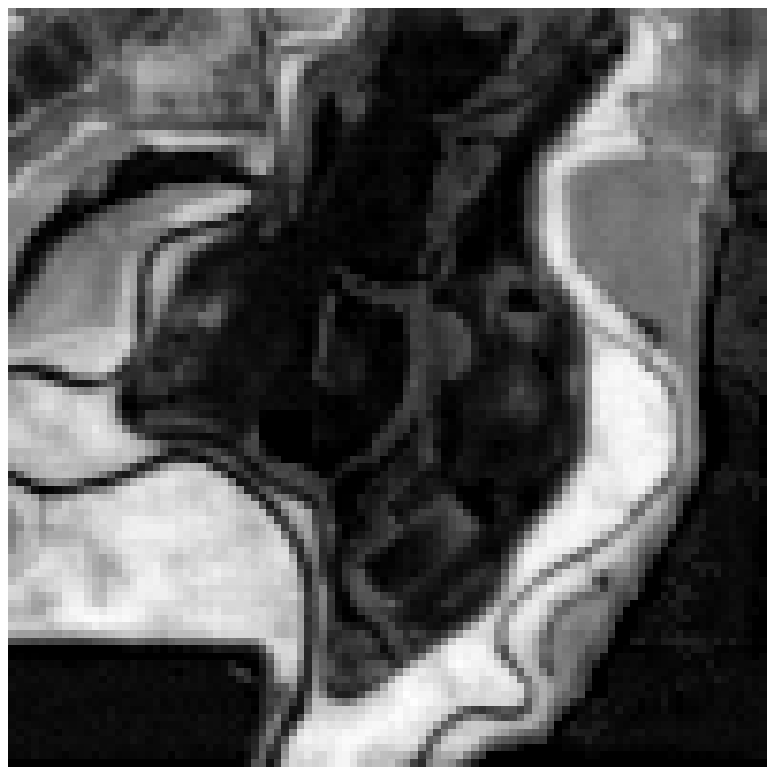}}\\
    \subfloat{
    \includegraphics[width=0.25\textwidth]{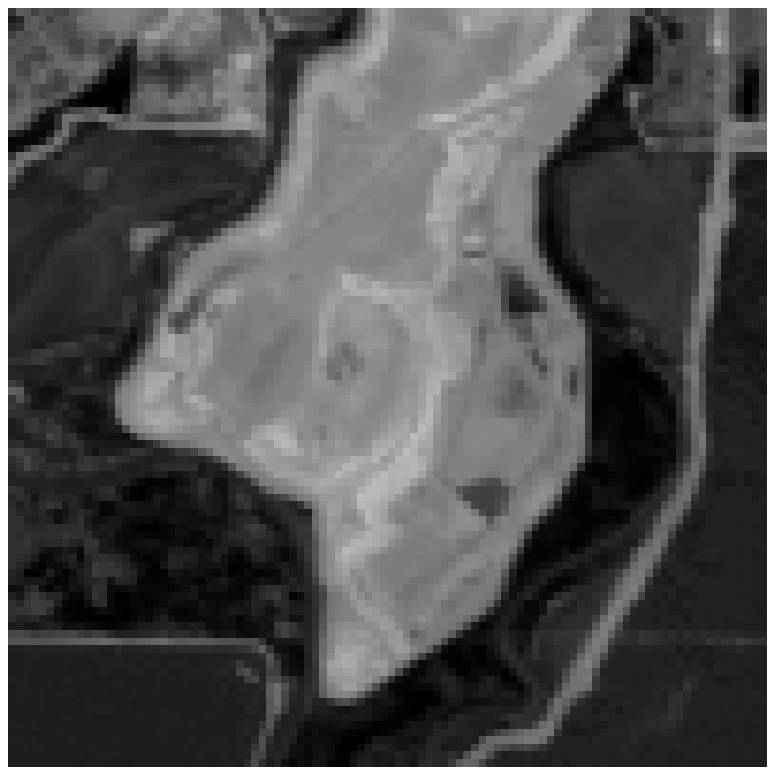}}
    \subfloat{
    \includegraphics[width=0.25\textwidth]{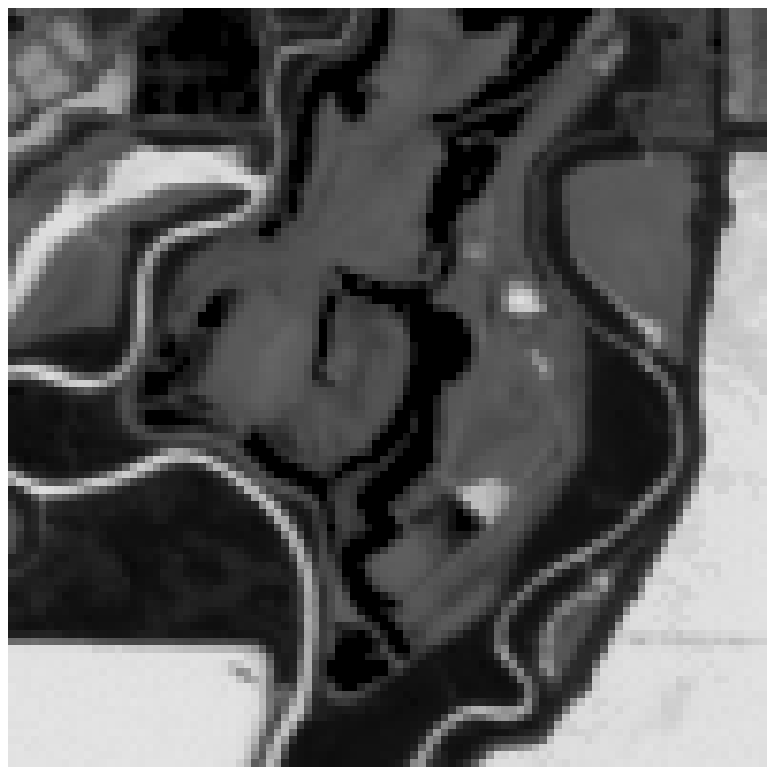}}
    \subfloat{
    \includegraphics[width=0.25\textwidth]{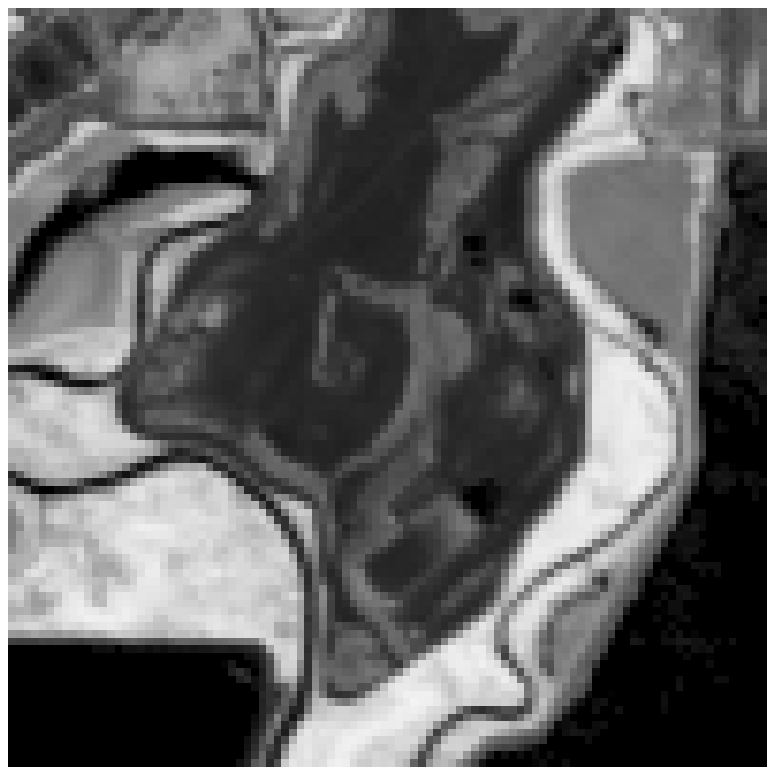}}
    \caption{Unmixed abundance maps for Moffett HS+PAN dataset: Estimated abundance maps using (Row 1) Berne's method, (Row 2) Yokoya's method, and (Row 3) UnS-FUMI. (Row 4) Reference abundance maps.}
    \label{fig:Moffet_Abu}
\end{figure*}

The unmixed endmembers and abundance maps are displayed in Figs. \ref{fig:Moffet_End} and \ref{fig:Moffet_Abu}
whereas quantitative unmixing results are reported in Table \ref{tb:Moffet_Q_unmixing}.
FUMI offers competitive endmember estimation and much better abundance estimation compared with Berne's and 
Yokoya's methods. It is interesting to note that S-FUMI and UnS-FUMI share very similar fusion results.
However, the endmember estimation of UnS-FUMI is much better compared with S-FUMI, which only exploits
the HS image to estimate the endmembers. This demonstrates that the estimation of endmembers benefits from being updated jointly with abundances, 
thanks to the complementary spectral and spatial information contained in the HS and high resolution PAN images.

\begin{table}[h!]
\renewcommand{\arraystretch}{1.1}
\centering \caption{Fusion Performance for Moffett HS+PAN dataset: RSNR (in dB), UIQI, SAM (in degree), ERGAS, DD (in $10^{-2}$) and time (in second).}
\vspace{0.1cm}
\begin{tabular}{|c|cccccc|}
\hline
 Methods & RSNR & UIQI & SAM  & ERGAS & DD & Time \\
\hline
\hline
Berne2010 &  16.95  & 0.8923 & 4.446 & 3.777 & 3.158 & \Best{0.3}\\
\hline
Yokoya2012 &  17.04 &  0.9002 & 4.391 & 3.734 &3.132 & 1.1\\
\hline
S-FUMI &  \Best{22.57} & \Best{0.9799} & \Best{2.184} & \Best{2.184} &\Best{1.488} & 21.1\\
\hline
UnS-FUMI &  {22.15} & {0.9778} & {2.346} & {2.292} &{1.577} & 32.2\\
\hline
\end{tabular}
\label{tb:Moffet_Q_fusion}
\end{table}

\begin{table}[h!]
\renewcommand{\arraystretch}{1.1}
\centering \caption{Unmixing Performance for Moffett HS+PAN dataset: SAM$_\bfM$ (in degree), NMSE$_\bfM$ (in dB) and NMSE$_\bfA$ (in dB).}
\vspace{0.1cm}
\begin{tabular}{|c|ccc|}
\hline
 Methods & SAM$_\bfM$ & NMSE$_\bfM$ & NMSE$_\bfA$ \\
\hline
\hline
Berne2010 &  7.568  & -16.425  & -11.167 \\
\hline
Yokoya2012 & \Best{6.772} & \Best{-17.405} &  -11.167\\
\hline
S-FUMI   &  {7.579} & {-16.419} & {-14.172}\\
\hline
UnS-FUMI & {7.028} & {-16.685} & \Best{-14.695}\\
\hline
\end{tabular}
\label{tb:Moffet_Q_unmixing}
\end{table}

\subsubsection{Pavia dataset}
\begin{figure*}[h!]
\centering
    \subfloat{
    \includegraphics[width=0.25\textwidth]{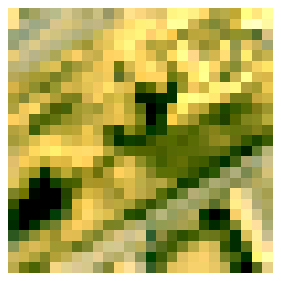}}
    \subfloat{
    \includegraphics[width=0.25\textwidth]{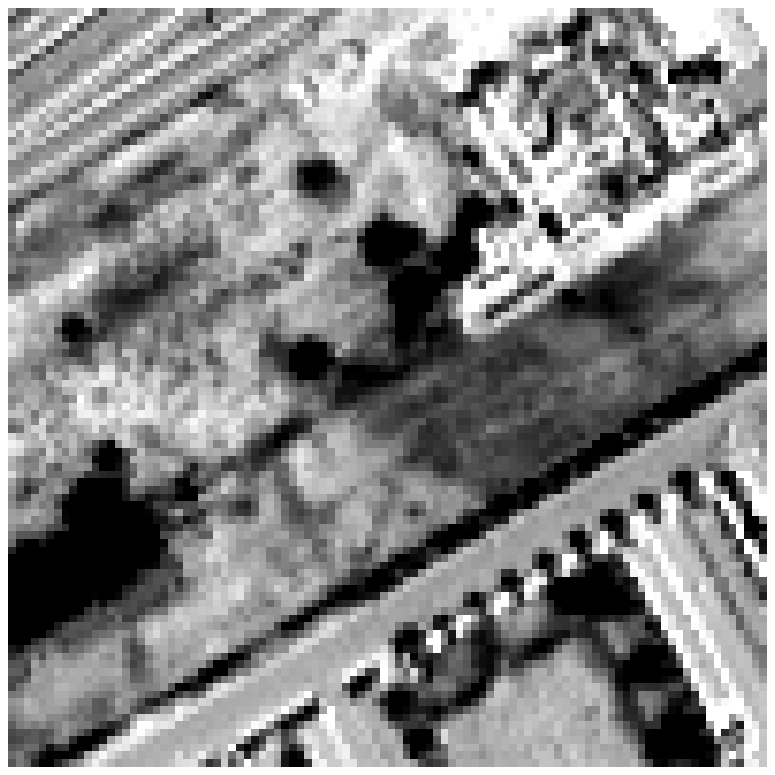}}
    \subfloat{
    \includegraphics[width=0.25\textwidth]{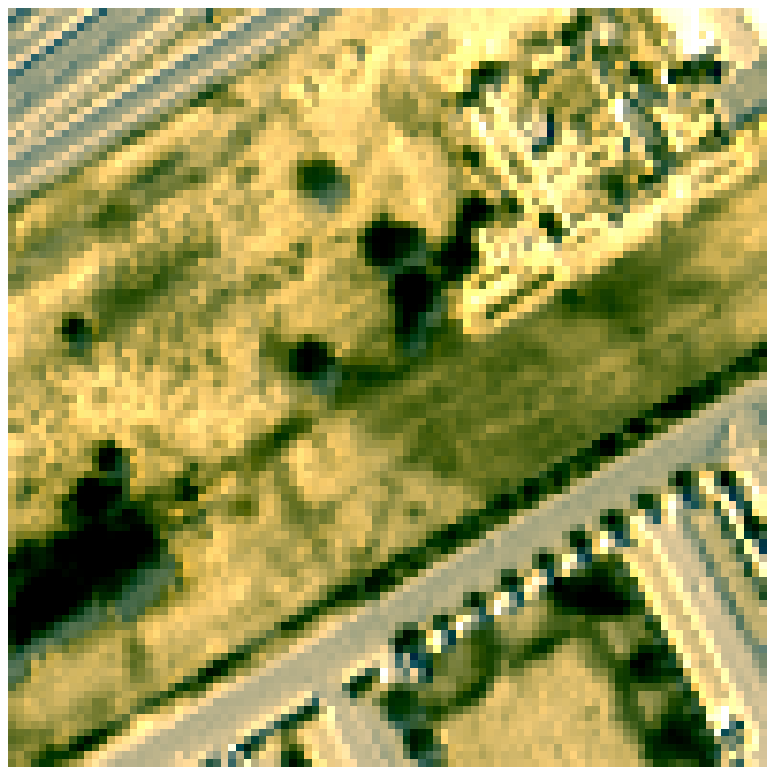}}\\
    \subfloat{
    \includegraphics[width=0.25\textwidth]{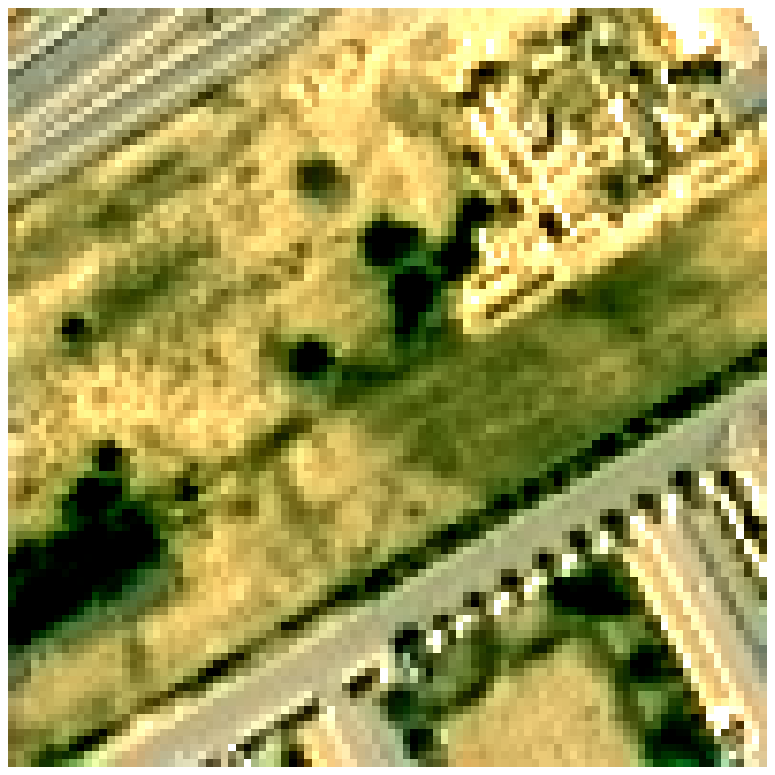}}
    \subfloat{
    \includegraphics[width=0.25\textwidth]{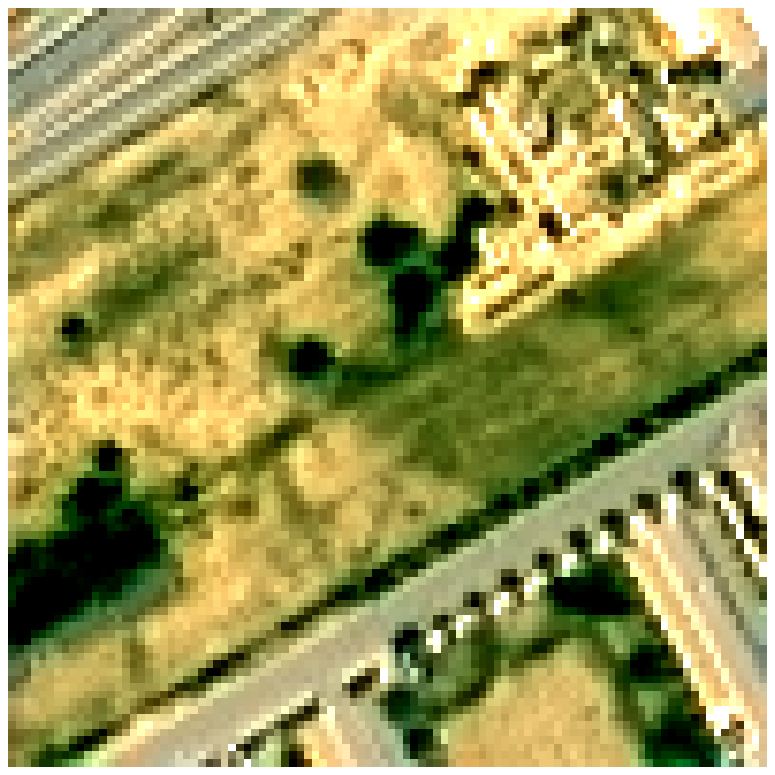}}
    \subfloat{
    \includegraphics[width=0.25\textwidth]{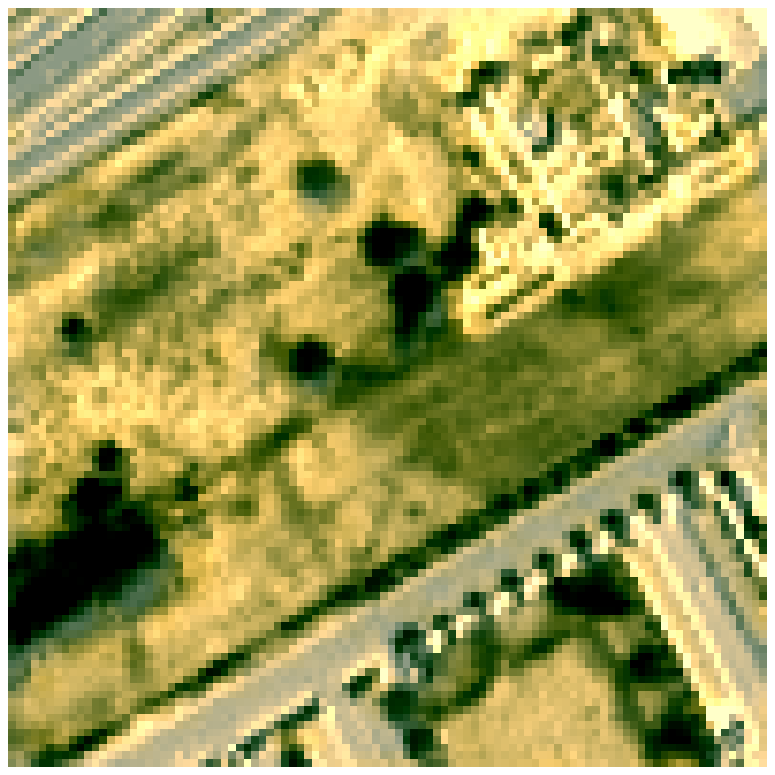}}
    \subfloat{
    \includegraphics[width=0.25\textwidth]{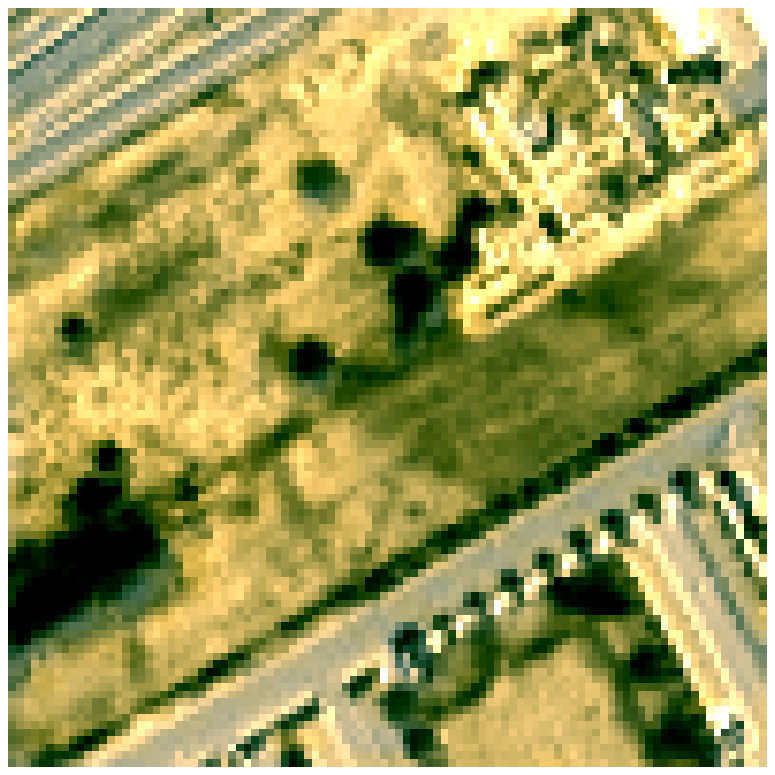}}\\
    \subfloat{
    \includegraphics[width=0.25\textwidth]{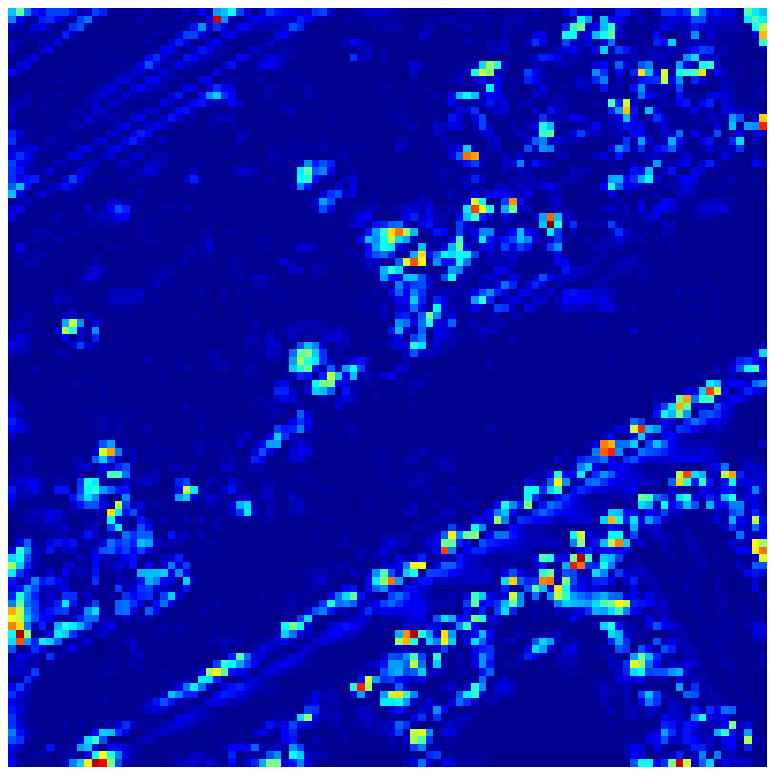}}
    \subfloat{
    \includegraphics[width=0.25\textwidth]{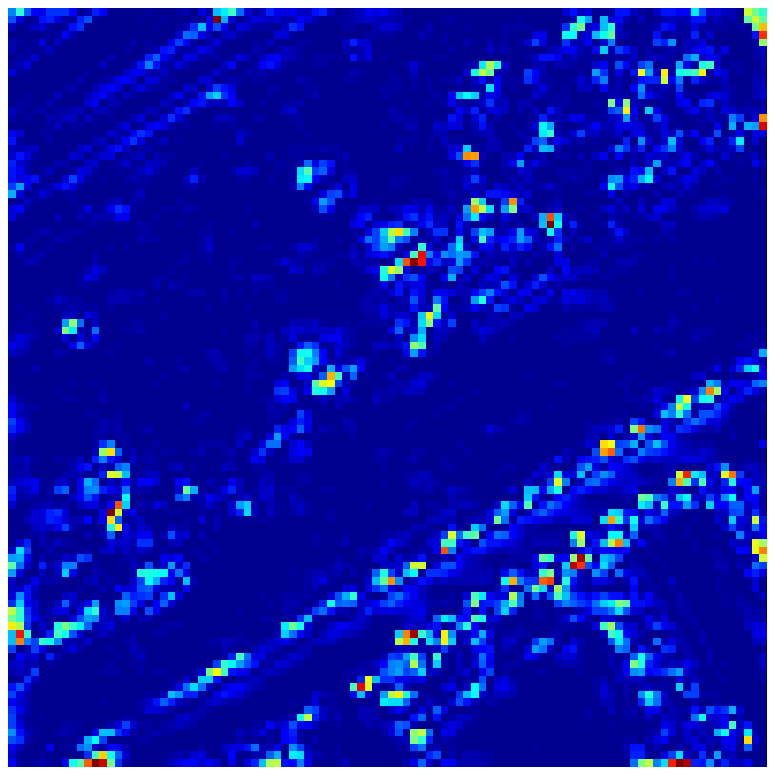}}
    \subfloat{
    \includegraphics[width=0.25\textwidth]{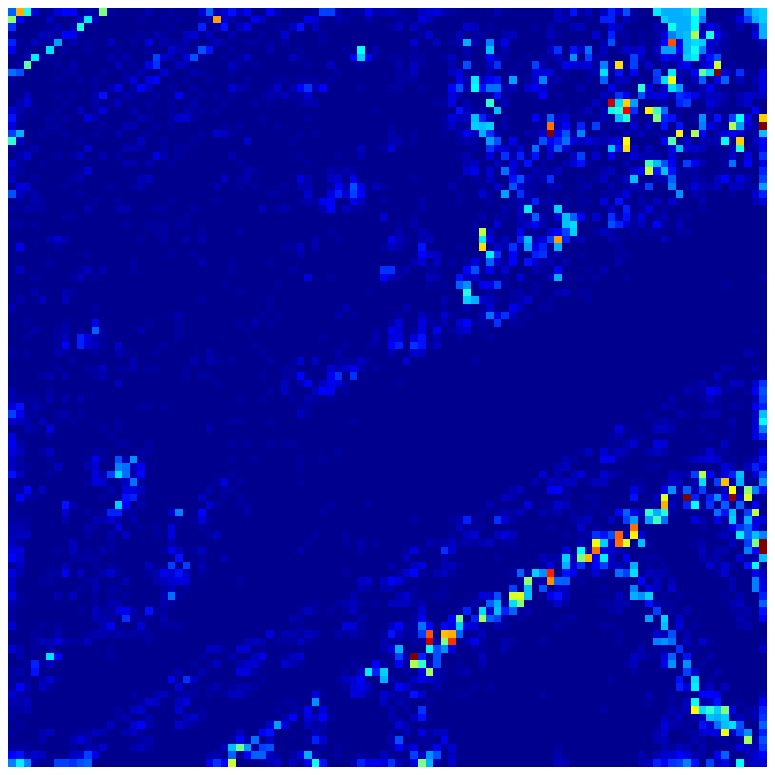}}
    \subfloat{
    \includegraphics[width=0.25\textwidth]{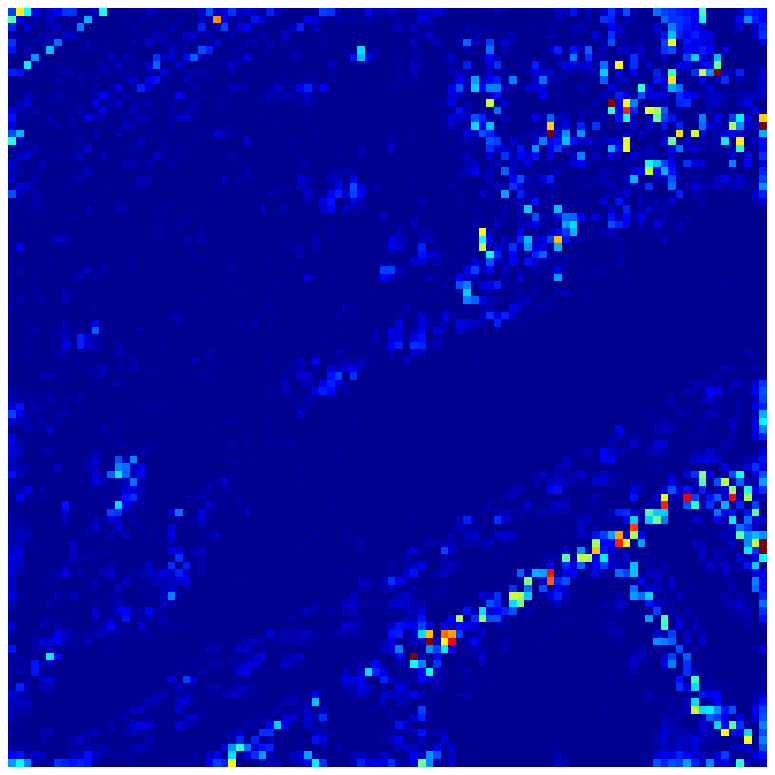}}
    \caption{Hyperspectral pansharpening results (Pavia dataset): (Top 1) HS image. (Top 2) PAN image. (Top 3) Reference image. (Middle 1) Berne's method. (Middle 2) Yokoya's method. (Middle 3) S-FUMI method. (Middle 4) UnS-FUMI method. (Bottom 1-4) The corresponding RMSE maps.} 
	\label{fig:HS_PAN_PAVIA}
\end{figure*}

\begin{figure}[h!]
\centering
\includegraphics[width=0.5\textwidth]{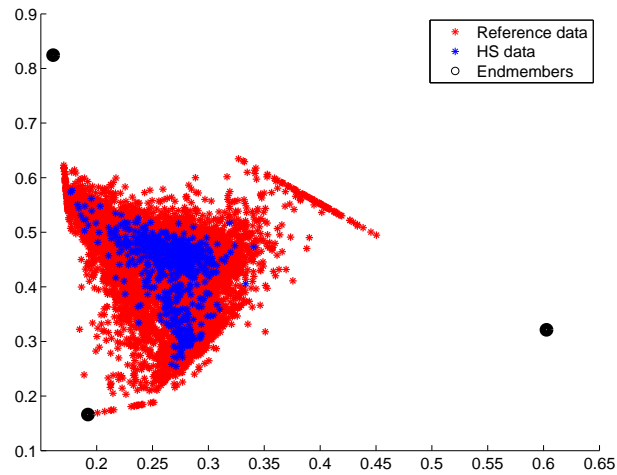}
\caption{Scattered Pavia data: The 30th and the 80th bands are selected as the coordinates.}
\label{fig:scatter_data_PAVIA}
\end{figure}

\begin{figure*}[h!]
\centering
	\subfloat{
	\includegraphics[width=0.4\textwidth]{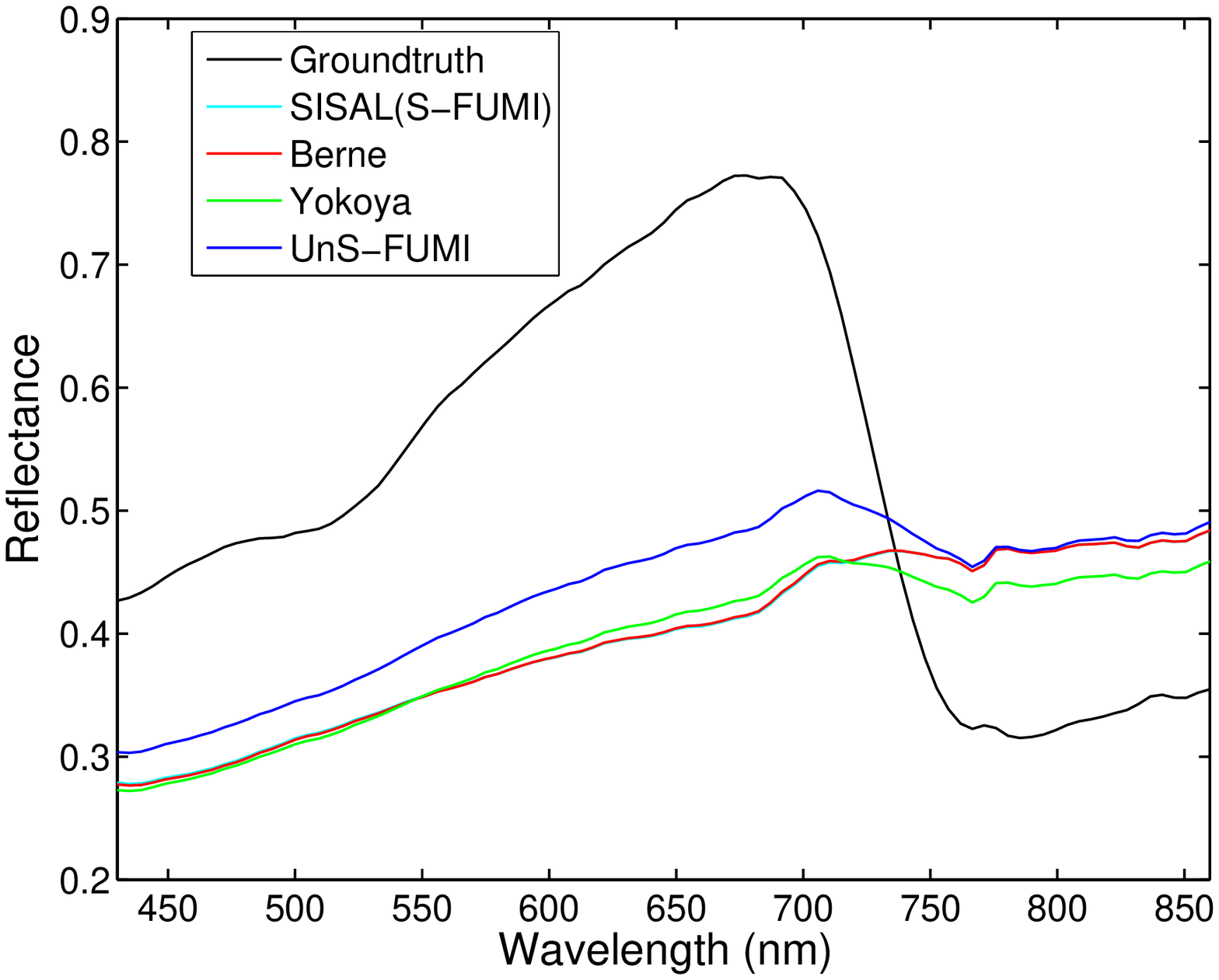}}
	\subfloat{
	\includegraphics[width=0.4\textwidth]{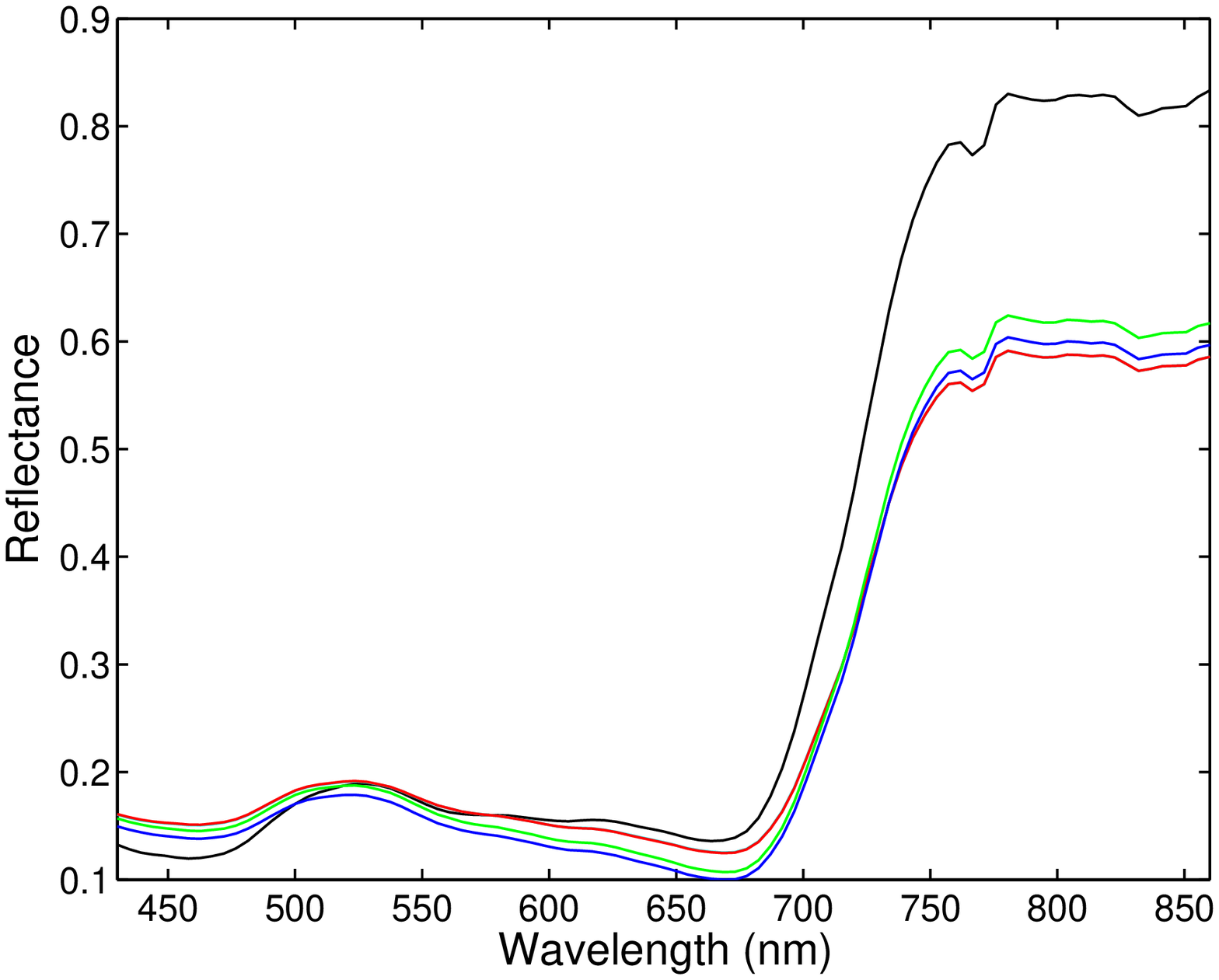}}\\
	\subfloat{
	\includegraphics[width=0.4\textwidth]{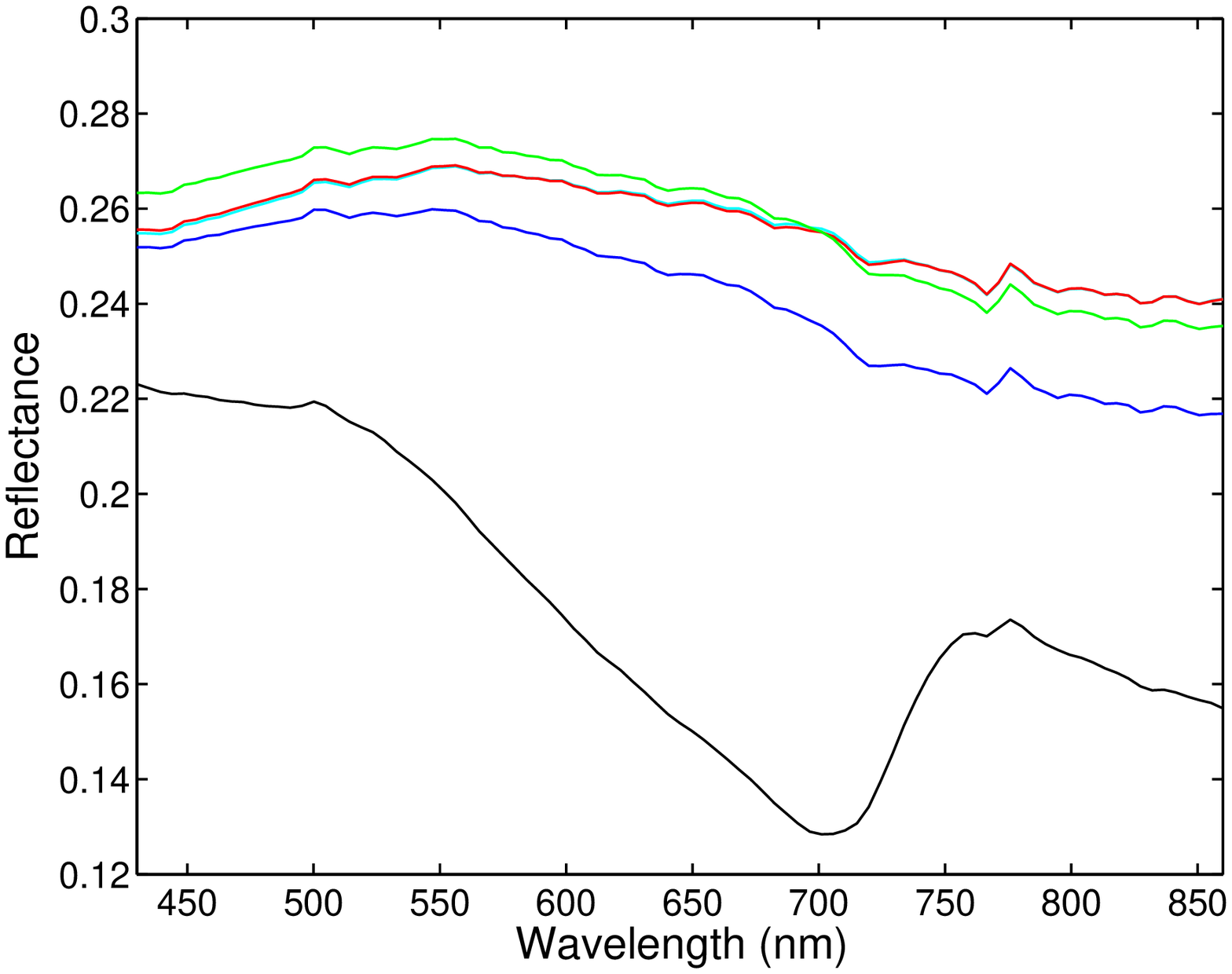}}
	\subfloat{
	\includegraphics[width=0.4\textwidth]{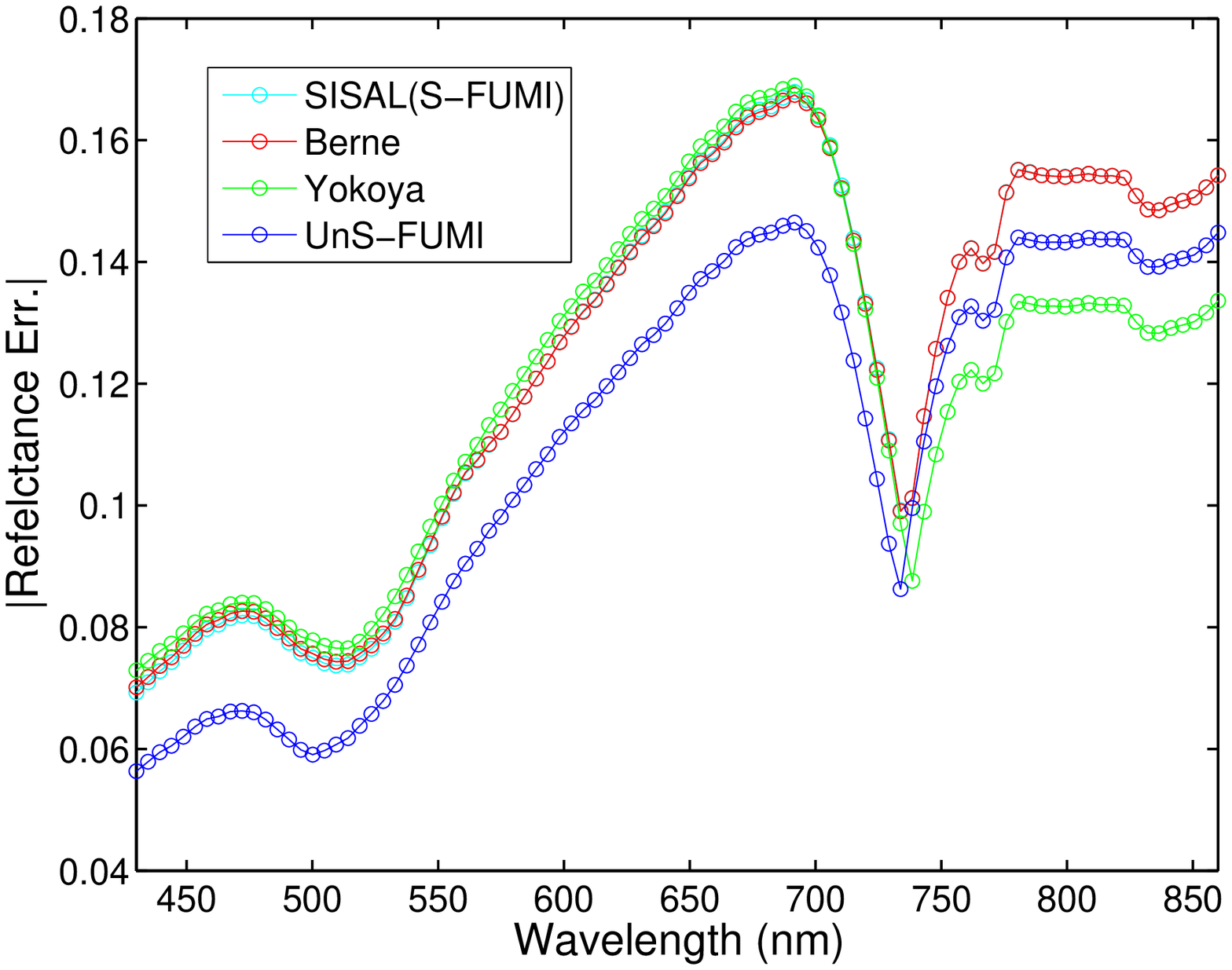}}
	\caption{Unmixed endmembers for Pavia HS+PAN dataset: (Top and bottom left) Estimated three endmembers and ground truth. (Bottom right) Sum of absolute value of all endmember errors as a function of wavelength. }
\label{fig:PAVIA_End}
\end{figure*}

\begin{figure*}[h!]
\centering
\subfloat{
    \includegraphics[width=0.25\textwidth]{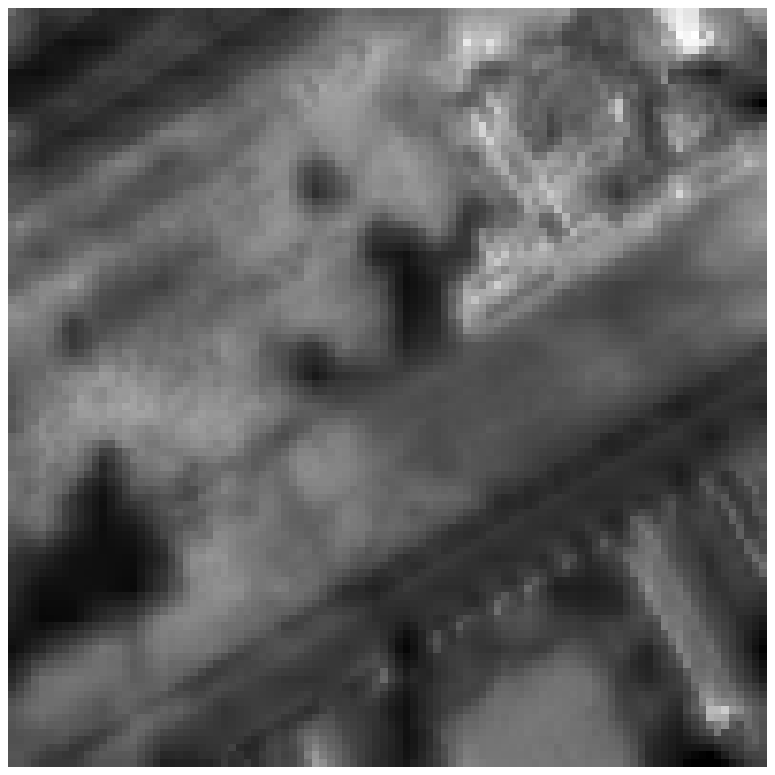}}
    \subfloat{
    \includegraphics[width=0.25\textwidth]{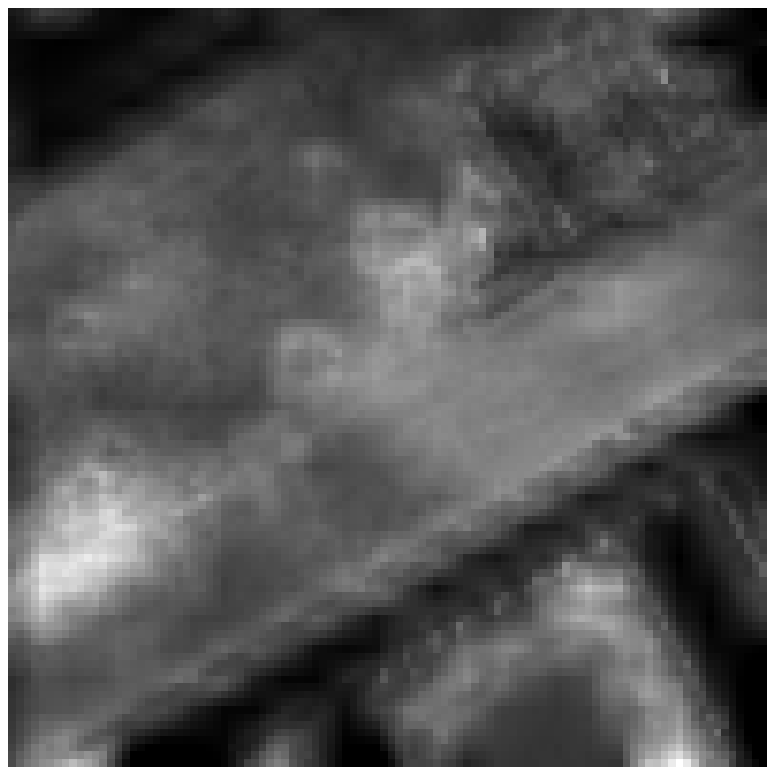}}
    \subfloat{
    \includegraphics[width=0.25\textwidth]{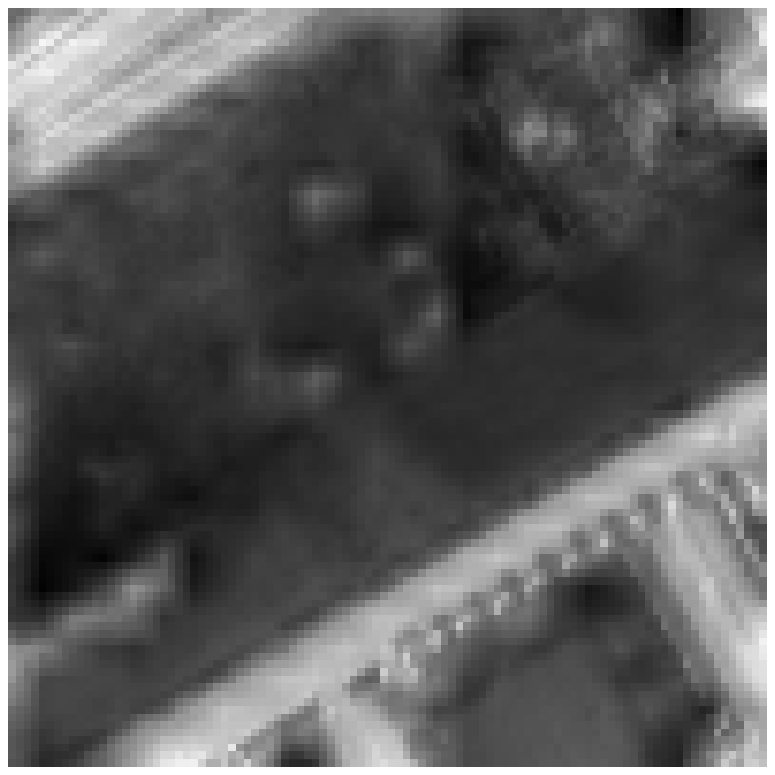}}\\
    \subfloat{
    \includegraphics[width=0.25\textwidth]{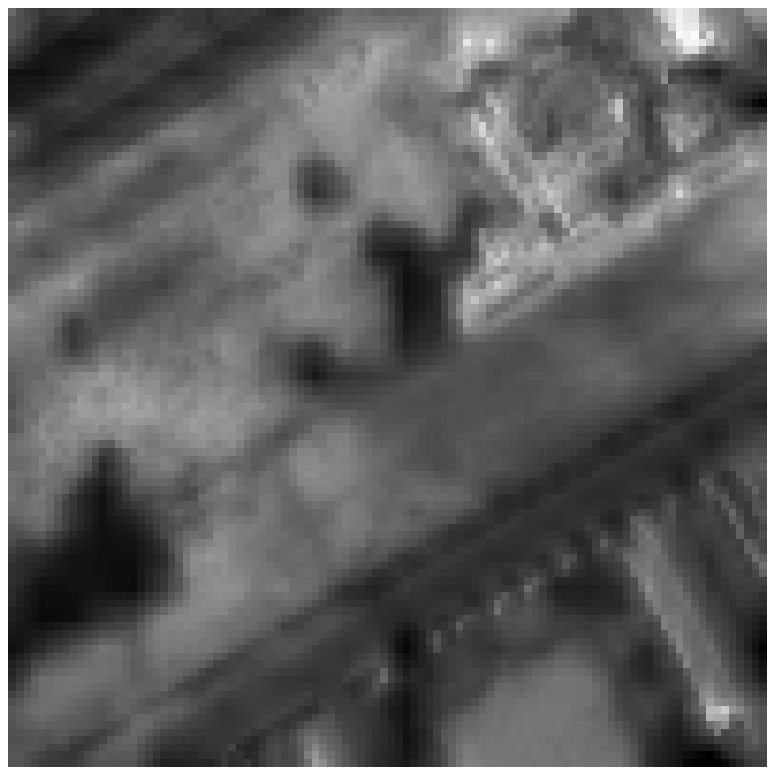}}
    \subfloat{
    \includegraphics[width=0.25\textwidth]{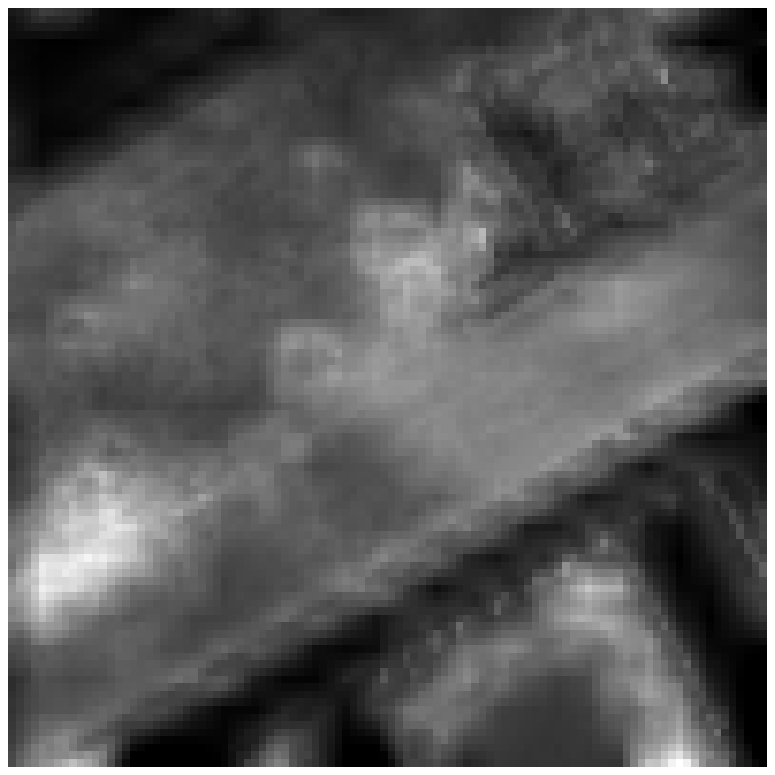}}
    \subfloat{
    \includegraphics[width=0.25\textwidth]{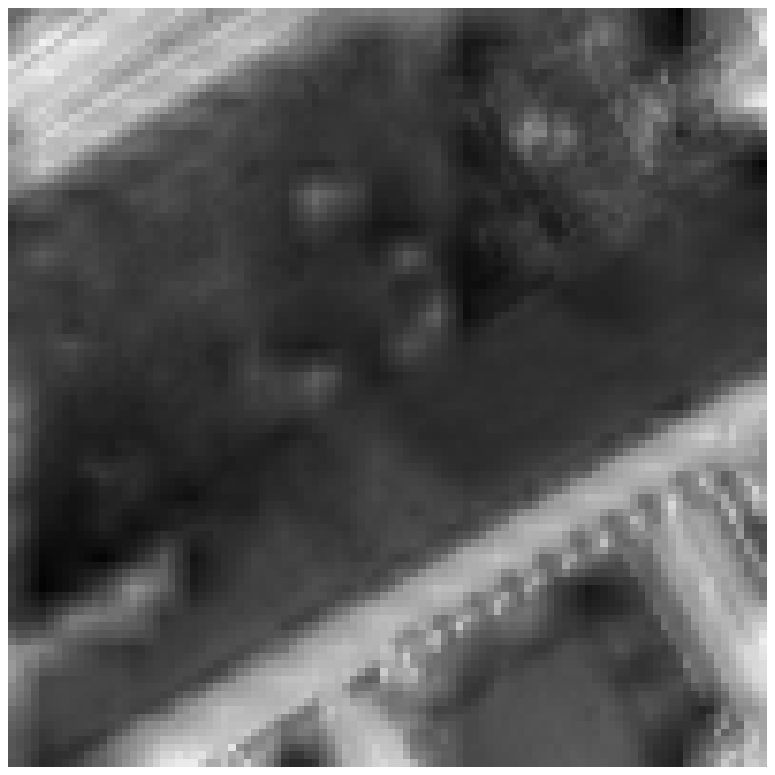}}\\
 	\subfloat{
    \includegraphics[width=0.25\textwidth]{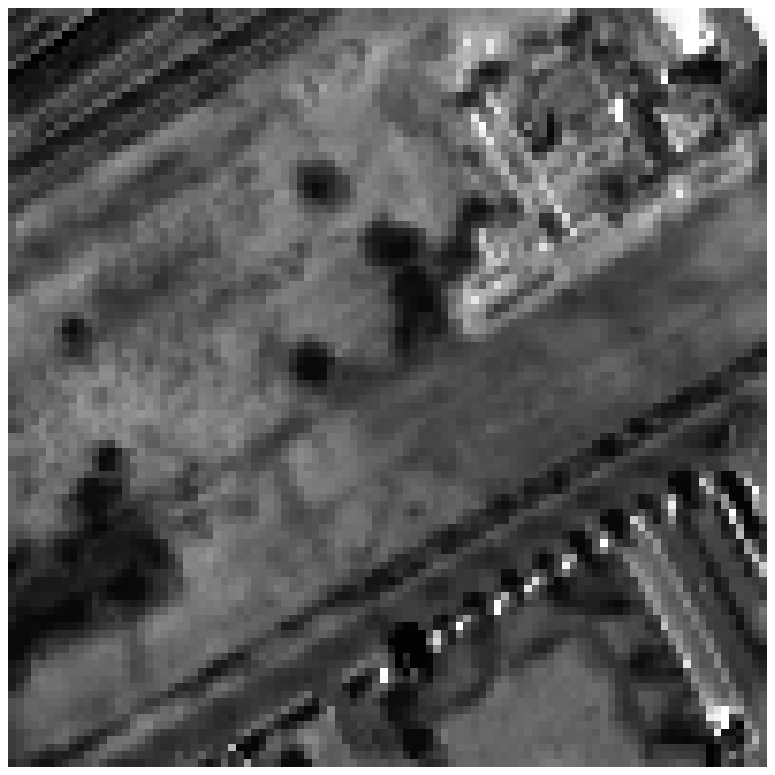}}
    \subfloat{
    \includegraphics[width=0.25\textwidth]{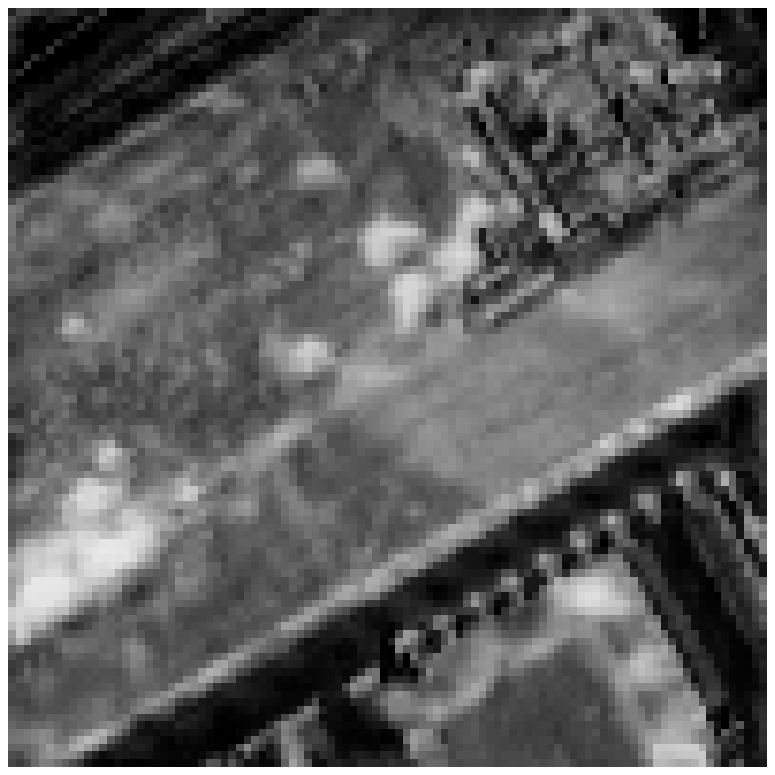}}
    \subfloat{
    \includegraphics[width=0.25\textwidth]{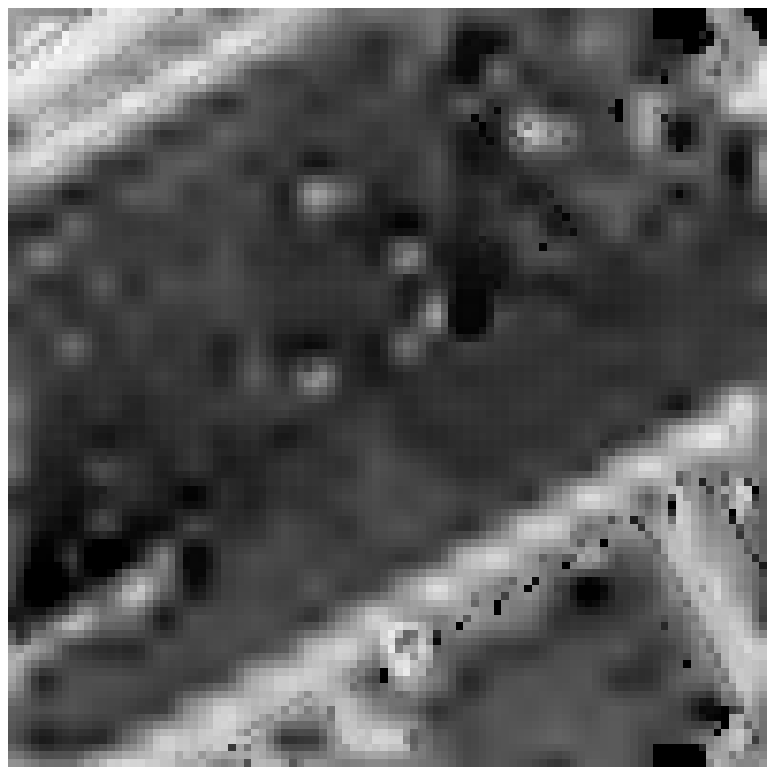}}\\
    \subfloat{
    \includegraphics[width=0.25\textwidth]{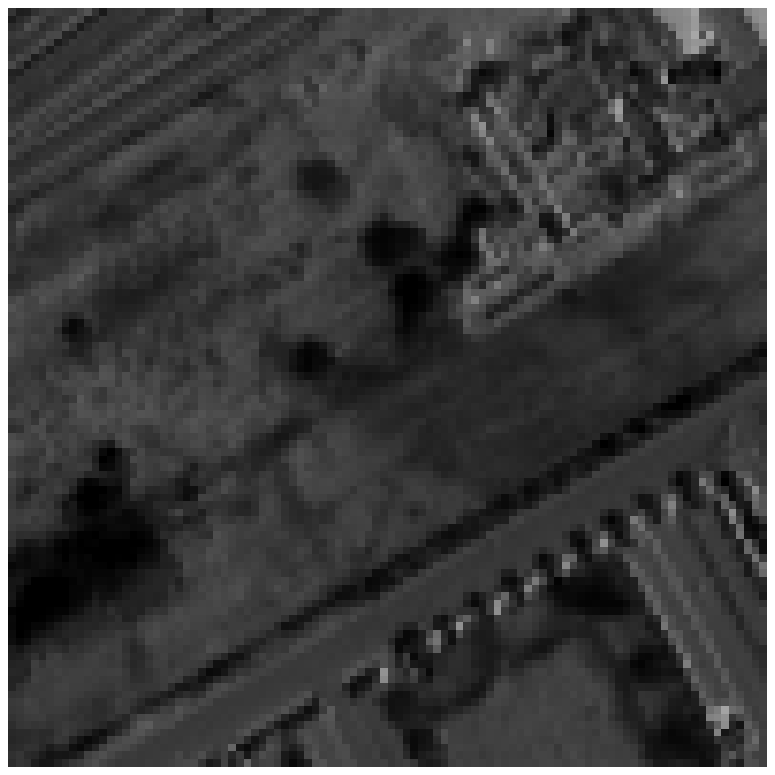}}
    \subfloat{
    \includegraphics[width=0.25\textwidth]{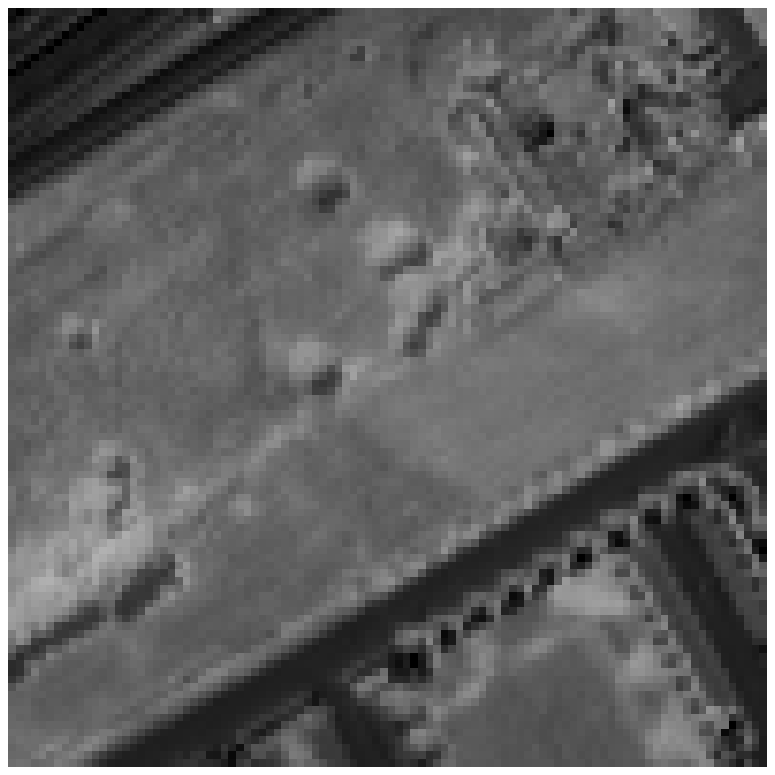}}
    \subfloat{
    \includegraphics[width=0.25\textwidth]{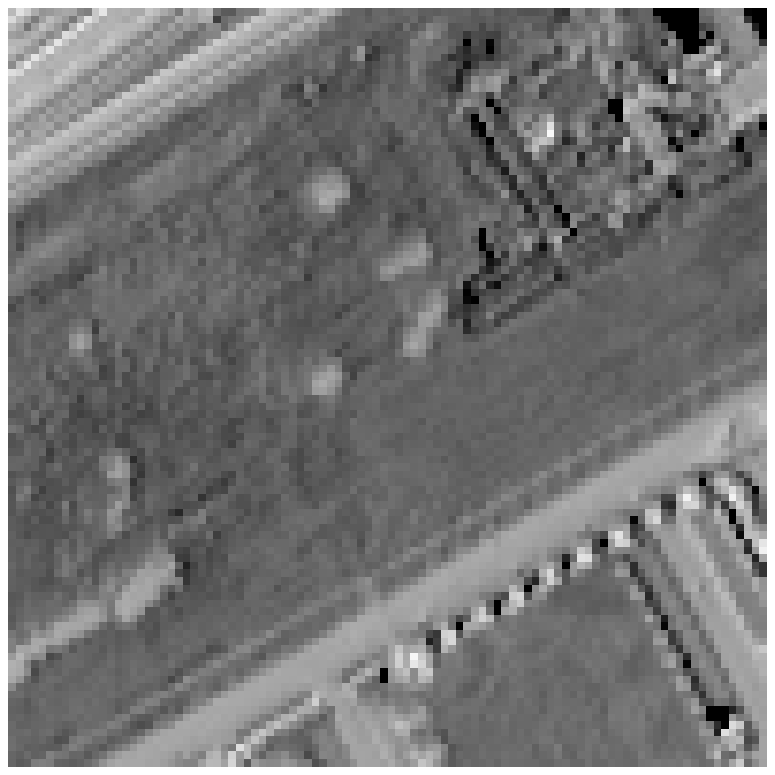}}
    \caption{Unmixed abundance maps for Pavia HS+PAN dataset: Estimated abundance maps using (Row 1) Berne's method, (Row 2) Yokoya's method, and (Row 3) UnS-FUMI. (Row 4) Reference abundance maps.}
    \label{fig:PAVIA_Abu}
\end{figure*}

In this section, we test the proposed algorithm on another dataset, in which 
the reference image is a $100 \times 100 \times 93$ HS image acquired 
over Pavia, Italy, by the reflective optics system imaging spectrometer (ROSIS). 
This image was initially composed of $115$ bands that have been reduced to $93$ bands
after removing the water vapor absorption bands. A composite color image of the scene
of interest is shown in the top right of Figs. \ref{fig:HS_PAN_PAVIA}.
The observed HS and co-registered PAN images are simulated similarly to the Moffet 
dataset and are shown in the top left and middle of Figs. \ref{fig:HS_PAN_PAVIA}.
The scattered reference and HS data are displayed in Fig. \ref{fig:scatter_data_PAVIA},
showing the high mixture of endmembers in the HS image.
The fusion results are displayed in Figs. \ref{fig:HS_PAN_PAVIA}
whereas the unmixed endmembers and abundance maps are shown in 
Figs. \ref{fig:PAVIA_End} and \ref{fig:PAVIA_Abu}.
The corresponding quantitative fusion and unmixing results are reported in Tables \ref{tb:PAVIA_Q_fusion} 
and \ref{tb:PAVIA_Q_unmixing}. These results are consistent with the analysis associated 
with the Moffet dataset. Both visually and quantitatively, S-FUMI and UnS-FUMI give competitive results,
which are much better than the other methods. In terms of unmixing, UnS-FUMI outperforms S-FUMI for 
both endmember and abundance estimations, due to the alternating update of endmembers and abundances.


%

\begin{table}[h!]
\renewcommand{\arraystretch}{1.1}
\centering \caption{Fusion Performance for Pavia HS+PAN dataset: RSNR (in dB), UIQI, SAM (in degree), ERGAS, DD (in $10^{-2}$) and time (in second).}
\vspace{0.1cm}
\begin{tabular}{|c|cccccc|}
\hline
 Methods & RSNR & UIQI & SAM  & ERGAS & DD & Time \\
\hline
\hline
Berne2010 &  21.53  & 0.9023 & 2.499 & 1.692 & 1.425 & \Best{0.6}\\
\hline
Yokoya2012 &  21.73 & 0.9119 & 2.416 & 1.655 &1.388 & 3.3\\
\hline
S-FUMI &  {24.13} &  {0.9456} & \Best{1.504} & {1.261} &{0.948} & 4.9\\
\hline
UnS-FUMI & \Best{24.26} & \Best{0.9504} & {1.541} & \Best{1.215} &\Best{0.925} & 34.3\\
\hline
\end{tabular}
\label{tb:PAVIA_Q_fusion}
\end{table}

\begin{table}[h!]
\renewcommand{\arraystretch}{1.1}
\centering \caption{Unmixing Performance for Pavia HS+PAN dataset: SAM$_\bfM$ (in degree), NMSE$_\bfM$ (in dB) and NMSE$_\bfA$ (in dB).}
\vspace{0.1cm}
\begin{tabular}{|c|ccc|}
\hline
 Methods & SAM$_\bfM$ & NMSE$_\bfM$ & NMSE$_\bfA$ \\
\hline
\hline
Berne2010 &  11.77  & -8.78  & -7.21 \\
\hline
Yokoya2012 & 10.43 & -9.21 &  -7.26\\
\hline
S-FUMI  &  {11.80} & {-8.78} & {-6.19}\\
\hline
UnS-FUMI  &  \Best{9.71} & \Best{-10.04} & \Best{-8.06}\\
\hline
\end{tabular}
\label{tb:PAVIA_Q_unmixing}
\end{table}


\section{Conclusion}
\label{sec:concls}
This paper proposed a new algorithm based on spectral unmixing for fusing multi-band images.
In this algorithm, the endmembers and abundances were updated alternatively, both using an 
alternating direction method of multipliers. The updates for abundances consisted of solving
a Sylvester matrix equation and projecting onto a simplex. Thanks to the recently developed
R-FUSE algorithm, this Sylvester equation was solved analytically thus efficiently, 
requiring no iterative update. The endmember updating was divided into two steps: a least 
square regression and a thresholding, that are both not computationally intensive. 
Numerical experiments showed that the proposed joint fusion and unmixing algorithm compared
competitively with two state-of-the-art methods, with the advantage of improving the performance 
for both fusion and unmixing. Future work will consist of incorporating the spatial and
spectral degradation into the estimation framework. 

\bibliographystyle{ieeetran}
\bibliography{strings_all_ref,biblio_all}

\begin{thebibliography}{10}
\providecommand{\url}[1]{#1}
\csname url@samestyle\endcsname
\providecommand{\newblock}{\relax}
\providecommand{\bibinfo}[2]{#2}
\providecommand{\BIBentrySTDinterwordspacing}{\spaceskip=0pt\relax}
\providecommand{\BIBentryALTinterwordstretchfactor}{4}
\providecommand{\BIBentryALTinterwordspacing}{\spaceskip=\fontdimen2\font plus
\BIBentryALTinterwordstretchfactor\fontdimen3\font minus
  \fontdimen4\font\relax}
\providecommand{\BIBforeignlanguage}[2]{{%
\expandafter\ifx\csname l@#1\endcsname\relax
\typeout{** WARNING: IEEEtran.bst: No hyphenation pattern has been}%
\typeout{** loaded for the language `#1'. Using the pattern for}%
\typeout{** the default language instead.}%
\else
\language=\csname l@#1\endcsname
\fi
#2}}
\providecommand{\BIBdecl}{\relax}
\BIBdecl

\bibitem{Wei2016FUSION}
Q.~Wei, J.~M. Bioucas-Dias, N.~Dobigeon, J.-Y. Tourneret, and S.~Godsill,
  ``High-resolution hyperspectral image fusion based on spectral unmixing,'' in
  \emph{Proc. IEEE Int. Conf. Inf. Fusion (FUSION)}, Heidelberg, Germany, July
  2016, submitted.

\bibitem{Landgrebe2002}
D.~Landgrebe, ``Hyperspectral image data analysis,'' \emph{IEEE Signal Process.
  Mag.}, vol.~19, no.~1, pp. 17--28, Jan. 2002.

\bibitem{Navulur2006}
K.~Navulur, \emph{Multispectral Image Analysis Using the Object-Oriented
  Paradigm}, ser. Remote Sensing Applications Series.\hskip 1em plus 0.5em
  minus 0.4em\relax Boca Raton, FL: CRC Press, 2006.

\bibitem{Bacon2001}
R.~Bacon, Y.~Copin, G.~Monnet, B.~W. Miller, J.~Allington-Smith, M.~Bureau,
  C.~M. Carollo, R.~L. Davies, E.~Emsellem, H.~Kuntschner \emph{et~al.}, ``The
  sauron project--i. the panoramic integral-field spectrograph,'' \emph{Monthly
  Notices of the Royal Astronomical Society}, vol. 326, no.~1, pp. 23--35,
  2001.

\bibitem{Nelson2004}
S.~J. Nelson, ``Magnetic resonance spectroscopic imaging,'' \emph{IEEE
  Engineering Med. Biology Mag.}, vol.~23, no.~5, pp. 30--39, 2004.

\bibitem{Chang2007}
C.-I. Chang, \emph{Hyperspectral data exploitation: theory and
  applications}.\hskip 1em plus 0.5em minus 0.4em\relax New York: John Wiley \&
  Sons, 2007.

\bibitem{Shaw2003}
G.~A. Shaw and H.-h.~K. Burke, ``Spectral imaging for remote sensing,''
  \emph{Lincoln Laboratory Journal}, vol.~14, no.~1, pp. 3--28, 2003.

\bibitem{Aiazzi2012}
B.~Aiazzi, L.~Alparone, S.~Baronti, A.~Garzelli, and M.~Selva, ``25 years of
  pansharpening: a critical review and new developments,'' in \emph{Signal and
  Image Processing for Remote Sensing}, 2nd~ed., C.~H. Chen, Ed.\hskip 1em plus
  0.5em minus 0.4em\relax Boca Raton, FL: CRC Press, 2011, ch.~28, pp.
  533--548.

\bibitem{Loncan2015}
L.~Loncan, L.~B. Almeida, {J. M. Bioucas-Dias}, X.~Briottet, J.~Chanussot,
  N.~Dobigeon, S.~Fabre, W.~Liao, G.~Licciardi, M.~Simoes, J.-Y. Tourneret,
  M.~Veganzones, G.~Vivone, Q.~Wei, and N.~Yokoya, ``Hyperspectral
  pansharpening: a review,'' \emph{IEEE Geosci. Remote Sens. Mag.}, vol.~3,
  no.~3, pp. 27--46, Sept. 2015.

\bibitem{Wei2015JSTSP}
Q.~Wei, N.~Dobigeon, and J.-Y. Tourneret, ``Bayesian fusion of multi-band
  images,'' \emph{IEEE J. Sel. Topics Signal Process.}, vol.~9, no.~6, pp.
  1117--1127, Sept. 2015.

\bibitem{Wei2015TGRS}
Q.~Wei, J.~Bioucas-Dias, N.~Dobigeon, and J.~Tourneret, ``Hyperspectral and
  multispectral image fusion based on a sparse representation,'' \emph{IEEE
  Trans. Geosci. Remote Sens.}, vol.~53, no.~7, pp. 3658--3668, Jul. 2015.

\bibitem{Simoes2015}
M.~Simoes, J.~Bioucas-Dias, L.~Almeida, and J.~Chanussot, ``A convex
  formulation for hyperspectral image superresolution via subspace-based
  regularization,'' \emph{IEEE Trans. Geosci. Remote Sens.}, vol.~53, no.~6,
  pp. 3373--3388, Jun. 2015.

\bibitem{Zhukov1999}
B.~Zhukov, D.~Oertel, F.~Lanzl, and G.~Reinhackel, ``Unmixing-based multisensor
  multiresolution image fusion,'' \emph{IEEE Trans. Geosci. Remote Sens.},
  vol.~37, no.~3, pp. 1212--1226, May 1999.

\bibitem{Eismann2004}
M.~T. Eismann and R.~C. Hardie, ``Application of the stochastic mixing model to
  hyperspectral resolution enhancement,'' \emph{IEEE Trans. Geosci. Remote
  Sens.}, vol.~42, no.~9, pp. 1924--1933, Sep. 2004.

\bibitem{Yokoya2012coupled}
N.~Yokoya, T.~Yairi, and A.~Iwasaki, ``Coupled nonnegative matrix factorization
  unmixing for hyperspectral and multispectral data fusion,'' \emph{IEEE Trans.
  Geosci. Remote Sens.}, vol.~50, no.~2, pp. 528--537, 2012.

\bibitem{An2014}
Z.~An and Z.~Shi, ``Hyperspectral image fusion by multiplication of spectral
  constraint and {NMF},'' \emph{Optik-International Journal for Light and
  Electron Optics}, vol. 125, no.~13, pp. 3150--3158, 2014.

\bibitem{Huang2014}
B.~Huang, H.~Song, H.~Cui, J.~Peng, and Z.~Xu, ``Spatial and spectral image
  fusion using sparse matrix factorization,'' \emph{IEEE Trans. Geosci. Remote
  Sens.}, vol.~52, no.~3, pp. 1693--1704, 2014.

\bibitem{Wycoff2013}
E.~Wycoff, T.-H. Chan, K.~Jia, W.-K. Ma, and Y.~Ma, ``A non-negative sparse
  promoting algorithm for high resolution hyperspectral imaging,'' in
  \emph{Proc. IEEE Int. Conf. Acoust., Speech, and Signal Processing
  (ICASSP)}.\hskip 1em plus 0.5em minus 0.4em\relax Vancouver, Canada: IEEE,
  2013, pp. 1409--1413.

\bibitem{Zhang2009}
Y.~Zhang, S.~De~Backer, and P.~Scheunders, ``Noise-resistant wavelet-based
  {B}ayesian fusion of multispectral and hyperspectral images,'' \emph{IEEE
  Trans. Geosci. Remote Sens.}, vol.~47, no.~11, pp. 3834 --3843, Nov. 2009.

\bibitem{Kawakami2011}
R.~Kawakami, J.~Wright, Y.-W. Tai, Y.~Matsushita, M.~Ben-Ezra, and K.~Ikeuchi,
  ``High-resolution hyperspectral imaging via matrix factorization,'' in
  \emph{Proc. IEEE Int. Conf. Comp. Vision and Pattern Recognition
  (CVPR)}.\hskip 1em plus 0.5em minus 0.4em\relax Providence, USA: IEEE, Jun.
  2011, pp. 2329--2336.

\bibitem{Berne2010}
O.~Berne, A.~Helens, P.~Pilleri, and C.~Joblin, ``Non-negative matrix
  factorization pansharpening of hyperspectral data: An application to
  mid-infrared astronomy,'' in \emph{Proc. IEEE GRSS Workshop Hyperspectral
  Image SIgnal Process.: Evolution in Remote Sens. (WHISPERS)}, Reykjavik,
  Iceland, Jun. 2010, pp. 1--4.

\bibitem{He2014}
X.~He, L.~Condat, J.~Bioucas-Dias, J.~Chanussot, and J.~Xia, ``A new
  pansharpening method based on spatial and spectral sparsity priors,''
  \emph{IEEE Trans. Image Process.}, vol.~23, no.~9, pp. 4160--4174, Sep. 2014.

\bibitem{Bieniarz2011}
J.~Bieniarz, D.~Cerra, J.~Avbelj, P.~Reinartz, and R.~M{\"u}ller,
  ``Hyperspectral image resolution enhancement based on spectral unmixing and
  information fusion,'' in \emph{Proc. ISPRS Hannover Workshop 2011:
  High-Resolution Earth Imaging for Geospatial Information}, Hannover, Germany,
  2011.

\bibitem{Zhang2013HS}
Z.~Zhang, Z.~Shi, and Z.~An, ``Hyperspectral and panchromatic image fusion
  using unmixing-based constrained nonnegative matrix factorization,''
  \emph{Optik-International Journal for Light and Electron Optics}, vol. 124,
  no.~13, pp. 1601--1608, 2013.

\bibitem{Lanaras2015}
C.~Lanaras, E.~Baltsaias, and K.~Schindler, ``Advances in hyperspectral and
  multispectral image fusion and spectral unmixing,'' \emph{Int. Soc.
  Photogrammetry Remote Sens.}, vol. XL-3/W3, no.~3, pp. 451--458, 2015.

\bibitem{Bioucas-Dias2012}
J.~Bioucas-Dias, A.~Plaza, N.~Dobigeon, M.~Parente, Q.~Du, P.~Gader, and
  J.~Chanussot, ``Hyperspectral unmixing overview: Geometrical, statistical,
  and sparse regression-based approaches,'' \emph{IEEE J. Sel. Topics Appl.
  Earth Observ. Remote Sens.}, vol.~5, no.~2, pp. 354--379, Apr. 2012.

\bibitem{Yokoya2013cross}
N.~Yokoya, N.~Mayumi, and A.~Iwasaki, ``Cross-calibration for data fusion of
  {EO}-1/{H}yperion and {T}erra/{ASTER},'' \emph{IEEE J. Sel. Topics Appl.
  Earth Observ. Remote Sens.}, vol.~6, no.~2, pp. 419--426, 2013.

\bibitem{Amro2011survey}
I.~Amro, J.~Mateos, M.~Vega, R.~Molina, and A.~K. Katsaggelos, ``A survey of
  classical methods and new trends in pansharpening of multispectral images,''
  \emph{EURASIP J. Adv. Signal Process.}, vol. 2011, no.~79, pp. 1--22, 2011.

\bibitem{Gonzalez2004fusion}
M.~Gonz{\'a}lez-Aud{\'\i}cana, J.~L. Saleta, R.~G. Catal{\'a}n, and
  R.~Garc{\'\i}a, ``Fusion of multispectral and panchromatic images using
  improved {IHS} and {PCA} mergers based on wavelet decomposition,'' \emph{IEEE
  Trans. Geosci. Remote Sens.}, vol.~42, no.~6, pp. 1291--1299, 2004.

\bibitem{Hardie2004}
R.~C. Hardie, M.~T. Eismann, and G.~L. Wilson, ``{MAP} estimation for
  hyperspectral image resolution enhancement using an auxiliary sensor,''
  \emph{IEEE Trans. Image Process.}, vol.~13, no.~9, pp. 1174--1184, Sep. 2004.

\bibitem{Molina1999}
R.~Molina, A.~K. Katsaggelos, and J.~Mateos, ``Bayesian and regularization
  methods for hyperparameter estimation in image restoration,'' \emph{IEEE
  Trans. Image Process.}, vol.~8, no.~2, pp. 231--246, 1999.

\bibitem{Molina2008}
R.~Molina, M.~Vega, J.~Mateos, and A.~K. Katsaggelos, ``Variational posterior
  distribution approximation in {B}ayesian super resolution reconstruction of
  multispectral images,'' \emph{Applied and Computational Harmonic Analysis},
  vol.~24, no.~2, pp. 251 -- 267, 2008.

\bibitem{Zhang2012}
Y.~Zhang, A.~Duijster, and P.~Scheunders, ``A {B}ayesian restoration approach
  for hyperspectral images,'' \emph{IEEE Trans. Geosci. Remote Sens.}, vol.~50,
  no.~9, pp. 3453 --3462, Sep. 2012.

\bibitem{Wei2014Bayesian}
Q.~Wei, N.~Dobigeon, and J.-Y. Tourneret, ``Bayesian fusion of hyperspectral
  and multispectral images,'' in \emph{Proc. IEEE Int. Conf. Acoust., Speech,
  and Signal Processing (ICASSP)}, Florence, Italy, May 2014.

\bibitem{Yokoya2013}
N.~Yokoya and A.~Iwasaki, ``Hyperspectral and multispectral data fusion mission
  on hyperspectral imager suite ({HISUI}),'' in \emph{Proc. IEEE Int. Conf.
  Geosci. Remote Sens. (IGARSS)}, Melbourne, Australia, Jul. 2013, pp.
  4086--4089.

\bibitem{Iordache2011SparseUnmixing}
M.-D. Iordache, J.~Bioucas-Dias, and A.~Plaza, ``Sparse unmixing of
  hyperspectral data,'' \emph{IEEE Trans. Geosci. Remote Sens.}, vol.~49,
  no.~6, pp. 2014--2039, Jun. 2011.

\bibitem{Wei2015FastUnmixing}
Q.~Wei, J.~Bioucas-Dias, N.~Dobigeon, and J.-Y. Tourneret, ``Fast spectral
  unmixing based on {D}ykstra's alternating projection,'' \emph{IEEE Trans.
  Signal Process.}, submitted.

\bibitem{Nascimento2005}
J.~Nascimento and J.~Bioucas-Dias, ``Vertex component analysis: A fast
  algorithm to unmix hyperspectral data,'' \emph{IEEE Trans. Geosci. Remote
  Sens.}, vol.~43, no.~4, pp. 898--910, 2005.

\bibitem{Dobigeon2009}
N.~Dobigeon, S.~Moussaoui, M.~Coulon, J.-Y. Tourneret, and A.~O. Hero, ``Joint
  {B}ayesian endmember extraction and linear unmixing for hyperspectral
  imagery,'' \emph{IEEE Trans. Signal Process.}, vol.~57, no.~11, pp.
  4355--4368, 2009.

\bibitem{Li2008MVSA}
J.~Li and J.~Bioucas-Dias, ``Minimum {V}olume {S}implex {A}nalysis: A fast
  algorithm to unmix hyperspectral data,'' in \emph{Proc. IEEE Int. Conf.
  Geosci. Remote Sens. (IGARSS)}, vol.~3, Boston, MA, Jul. 2008, pp. III --
  250--III -- 253.

\bibitem{Wei2015whispers}
Q.~Wei, N.~Dobigeon, and J.-Y. Tourneret, ``Bayesian fusion of multispectral
  and hyperspectral images using a block coordinate descent method,'' in
  \emph{Proc. IEEE GRSS Workshop Hyperspectral Image SIgnal Process.: Evolution
  in Remote Sens. (WHISPERS)}, Tokyo, Japan, Jun. 2015.

\bibitem{Robert2007}
C.~P. Robert, \emph{The {B}ayesian Choice: from Decision-Theoretic Motivations
  to Computational Implementation}, 2nd~ed., ser. Springer Texts in
  Statistics.\hskip 1em plus 0.5em minus 0.4em\relax New York, NY, USA:
  Springer-Verlag, 2007.

\bibitem{Gelman2013}
A.~Gelman, J.~B. Carlin, H.~S. Stern, D.~B. Dunson, A.~Vehtari, and D.~B.
  Rubin, \emph{Bayesian data analysis}, 3rd~ed.\hskip 1em plus 0.5em minus
  0.4em\relax Boca Raton, FL: CRC press, 2013.

\bibitem{Bertsekas1999}
D.~P. Bertsekas, \emph{Nonlinear programming}.\hskip 1em plus 0.5em minus
  0.4em\relax Athena Scientific, 1999.

\bibitem{Wei2016RFUSE}
Q.~Wei, N.~Dobigeon, J.-Y. Tourneret, J.~M. Bioucas-Dias, and S.~Godsill,
  ``{R-FUSE}: Robust fast fusion of multi-band images based on solving a
  {S}ylvester equation,'' \emph{IEEE Signal Process. Lett.}, submitted.

\bibitem{Keshava2002}
N.~Keshava and J.~F. Mustard, ``Spectral unmixing,'' \emph{IEEE Signal Process.
  Mag.}, vol.~19, no.~1, pp. 44--57, Jan. 2002.

\bibitem{Bioucas2008}
J.~M. Bioucas-Dias and J.~M. Nascimento, ``Hyperspectral subspace
  identification,'' \emph{IEEE Trans. Geosci. Remote Sens.}, vol.~46, no.~8,
  pp. 2435--2445, 2008.

\bibitem{Boyd2011}
S.~Boyd, N.~Parikh, E.~Chu, B.~Peleato, and J.~Eckstein, ``Distributed
  optimization and statistical learning via the alternating direction method of
  multipliers,'' \emph{Foundations and Trends{\textregistered} in Machine
  Learning}, vol.~3, no.~1, pp. 1--122, 2011.

\bibitem{Eckstein1992}
J.~Eckstein and D.~P. Bertsekas, ``On the {D}ouglas-{R}achford splitting method
  and the proximal point algorithm for maximal monotone operators,''
  \emph{Mathematical Programming}, vol.~55, no. 1-3, pp. 293--318, 1992.

\bibitem{Held1974}
M.~Held, P.~Wolfe, and H.~P. Crowder, ``Validation of subgradient
  optimization,'' \emph{Mathematical programming}, vol.~6, no.~1, pp. 62--88,
  1974.

\bibitem{Michelot1986}
C.~Michelot, ``A finite algorithm for finding the projection of a point onto
  the canonical simplex of $\mathbb{R}^n$,'' \emph{J. Optimization Theory
  Applications}, vol.~50, no.~1, pp. 195--200, 1986.

\bibitem{Duchi2008}
J.~Duchi, S.~{Shalev-Shwartz}, Y.~Singer, and T.~Chandra, ``Efficient
  projections onto the $\ell_1$ for learning in high dimensions,'' in
  \emph{Proc. Int. Conf. Machine Learning (ICML)}, Helsinki, Finland, 2008, pp.
  272--279.

\bibitem{Condat2014Fast}
L.~Condat, ``Fast projection onto the simplex and the l1 ball,'' \emph{Hal
  preprint: hal-01056171}, 2014.

\bibitem{Horn2012}
R.~A. Horn and C.~R. Johnson, \emph{Matrix analysis}.\hskip 1em plus 0.5em
  minus 0.4em\relax Cambridge, UK: Cambridge university press, 2012.

\bibitem{Wei2015FastFusion}
Q.~Wei, N.~Dobigeon, and J.-Y. Tourneret, ``Fast fusion of multi-band images
  based on solving a {S}ylvester equation,'' \emph{IEEE Trans. Image Process.},
  vol.~24, no.~11, pp. 4109--4121, Nov. 2015.

\bibitem{Bioucas2009SISAL}
J.~M. Bioucas-Dias, ``A variable splitting augmented {L}agrangian approach to
  linear spectral unmixing,'' in \emph{Proc. IEEE GRSS Workshop Hyperspectral
  Image SIgnal Process.: Evolution in Remote Sens. (WHISPERS)}.\hskip 1em plus
  0.5em minus 0.4em\relax Grenoble, France: IEEE, Aug. 2009, pp. 1--4.

\bibitem{Green1998imaging}
R.~O. Green, M.~L. Eastwood, C.~M. Sarture, T.~G. Chrien, M.~Aronsson, B.~J.
  Chippendale, J.~A. Faust, B.~E. Pavri, C.~J. Chovit, M.~Solis \emph{et~al.},
  ``Imaging spectroscopy and the airborne visible/infrared imaging spectrometer
  ({AVIRIS}),'' \emph{Remote Sens. of Environment}, vol.~65, no.~3, pp.
  227--248, 1998.

\end{thebibliography}
\end{document}